%% file: main.tex
\documentclass{article}
\pdfoutput=1
\usepackage{arxiv}

\usepackage[utf8]{inputenc}  
\usepackage[T1]{fontenc}     
\usepackage{hyperref}        
\usepackage{url}             
\usepackage{booktabs}        
\usepackage{amsfonts}        
\usepackage{nicefrac}        
\usepackage{microtype}       
\usepackage{graphicx}
\usepackage{amsmath,amssymb}
\usepackage{bbm}             
\usepackage{subfig}
\usepackage{float}

\title{Hyperplane Arrangements of Trained ConvNets Are Biased}

\author{
  Matteo Gamba\\
  \texttt{mgamba@kth.se}
  \And
  Stefan Carlsson\\
  \texttt{stefanc@kth.se}
  \And
  Hossein Azizpour\\
  \texttt{azizpour@kth.se}
  \And
  M\r{a}rten Bj\"{o}rkman\\
  \texttt{celle@kth.se}
  \AND
  KTH Royal Institute of Technology \\
  Stockholm, Sweden
}

\begin{document}
\maketitle

\input{./sections/abstract}

\input{./sections/introduction}

\input{./sections/background}

\input{./sections/experiments}

\input{./sections/related_works}

\input{./sections/conclusion}

\input{./sections/acknowledgement}

\bibliographystyle{unsrt}  


\clearpage

\appendix
\input{./sections/appendix}

\end{document}

%% file: sections/abstract.tex
\begin{abstract}

 We investigate the geometric properties of the functions learned by trained ConvNets in the preactivation space of their convolutional layers, by performing an empirical study of hyperplane arrangements induced by a convolutional layer. We introduce statistics over the weights of a trained network to study local arrangements and relate them to the training dynamics. 
 We observe that trained ConvNets show a significant statistical bias towards regular hyperplane configurations. Furthermore, we find that layers showing biased configurations are critical to validation performance for the architectures considered, trained on CIFAR10, CIFAR100 and ImageNet.


\end{abstract}

%% file: sections/introduction.tex
\section{Introduction}
\label{sec:introduction}

In recent years, understanding and interpreting the inner workings of deep networks has drawn considerable attention from the community~\cite{dziugaite2017computing,kawaguchi2017generalization,keskar2016large,jastrzebski2018three}. One long-standing question is the problem of identifying the inductive bias of state-of-the-art networks and the form of implicit regularization that is performed by the optimizer~\cite{neyshabur2014search,zhang2018understanding,arora2018stronger} and possibly by natural data itself~\cite{arpit2017closer}.

While earlier studies focused on the theoretical expressivity of deep networks and the advantage of deeper representations~\cite{montufar2014number,pascanu2013number,raghu2017expressive}, a recent trend in the literature is the study of the \textit{effective capacity} of \textit{trained} networks~\cite{zhang2018understanding,zhang2019all,hanin2019complexity,hanin2019deep}. In fact, while state-of-the-art deep networks are largely overparametrized, it is hypothesized that the full theoretical capacity of a model might not be realized in practice, due to some form of self-regulation at play during learning.

Some recent works have, thus, tried to find statistical bias consistently present in trained state-of-the-art models that is interpretable and correlates well with generalization~\cite{jiang2018predicting,novak2018sensitivity}. 

\begin{figure}[htbp]
    \centering
    \subfloat{\includegraphics[width=0.18\textwidth]{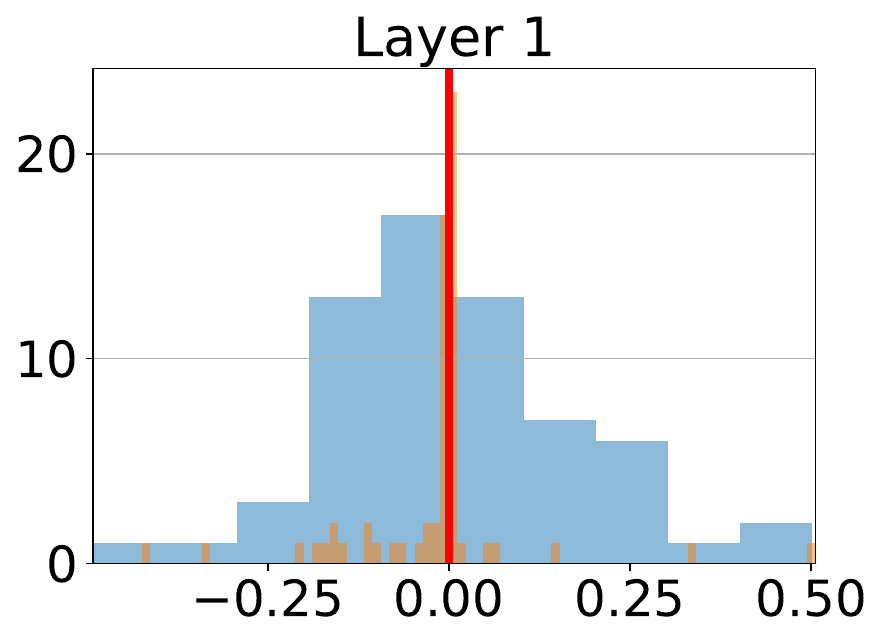}} ~
    \subfloat{\includegraphics[width=0.18\textwidth]{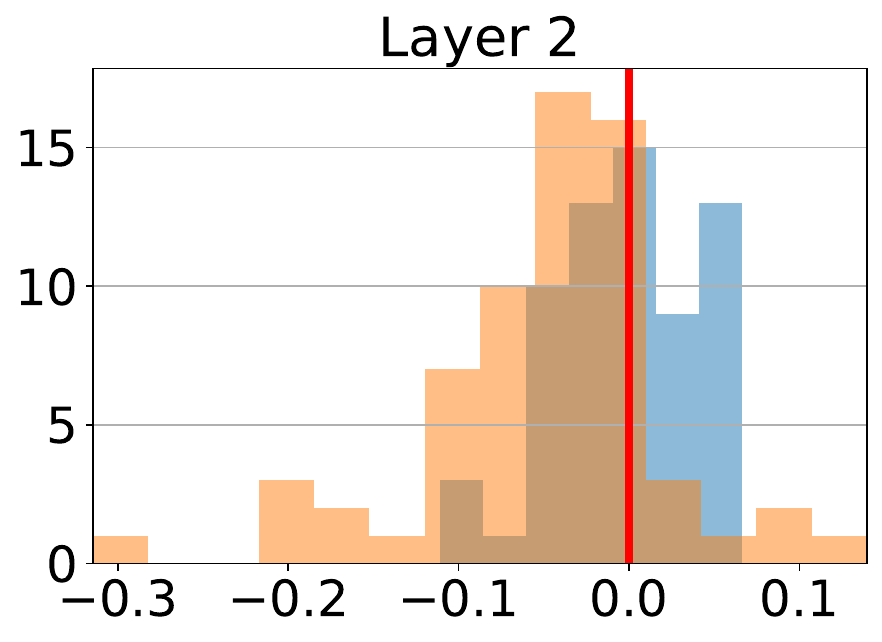}} ~
    \subfloat{\includegraphics[width=0.18\textwidth]{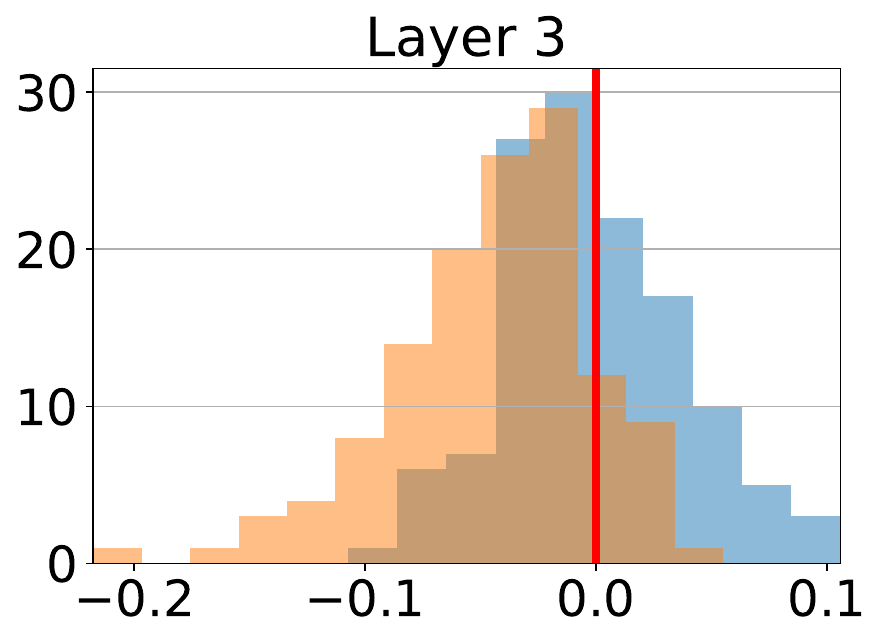}} ~
    \subfloat{\includegraphics[width=0.18\textwidth]{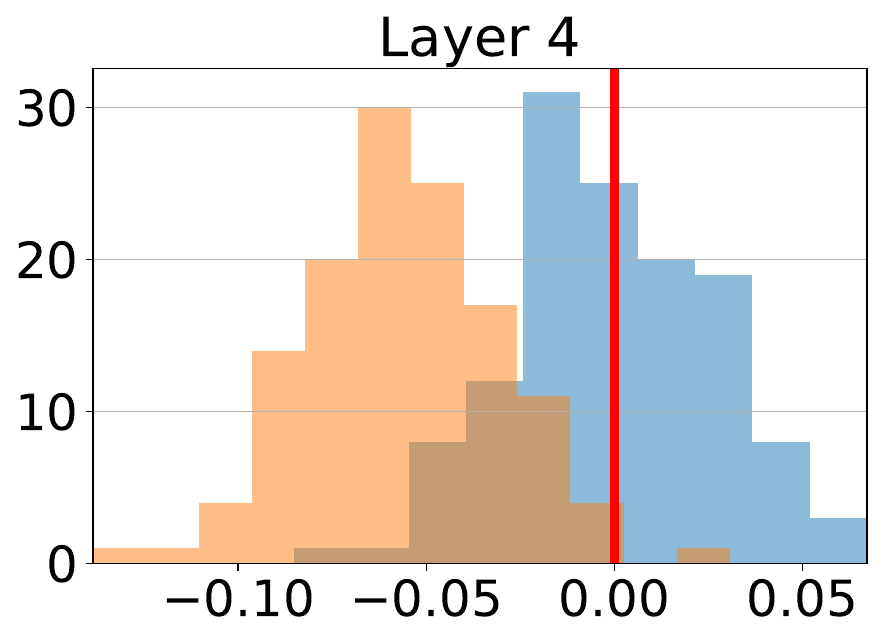}} ~
    \subfloat{\includegraphics[width=0.18\textwidth]{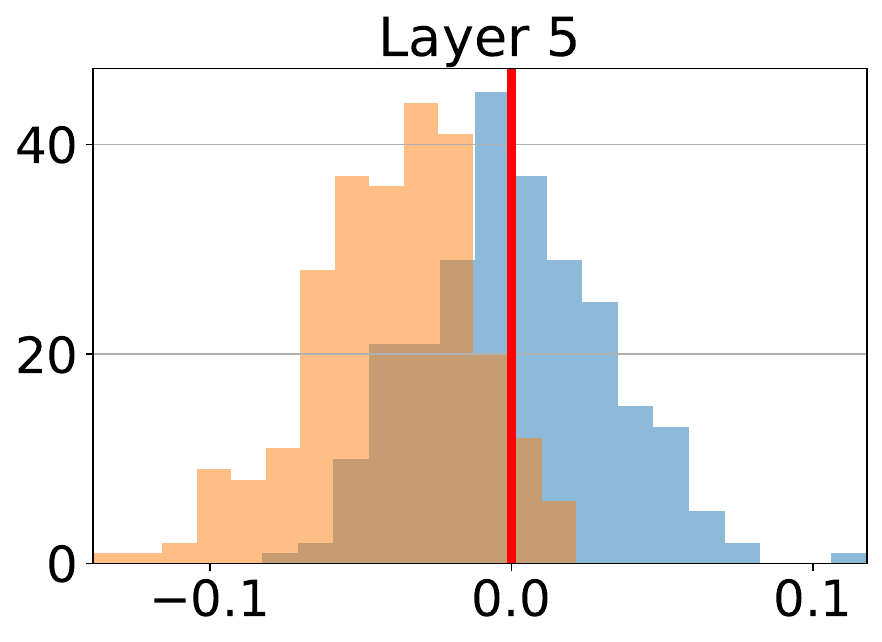}} \\
    \subfloat{\includegraphics[width=0.18\textwidth]{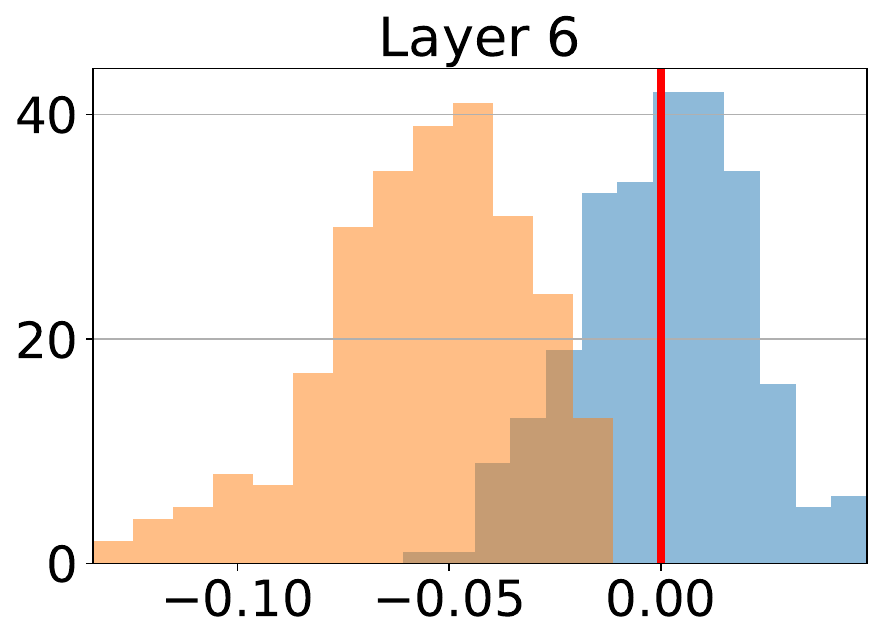}} ~
    \subfloat{\includegraphics[width=0.18\textwidth]{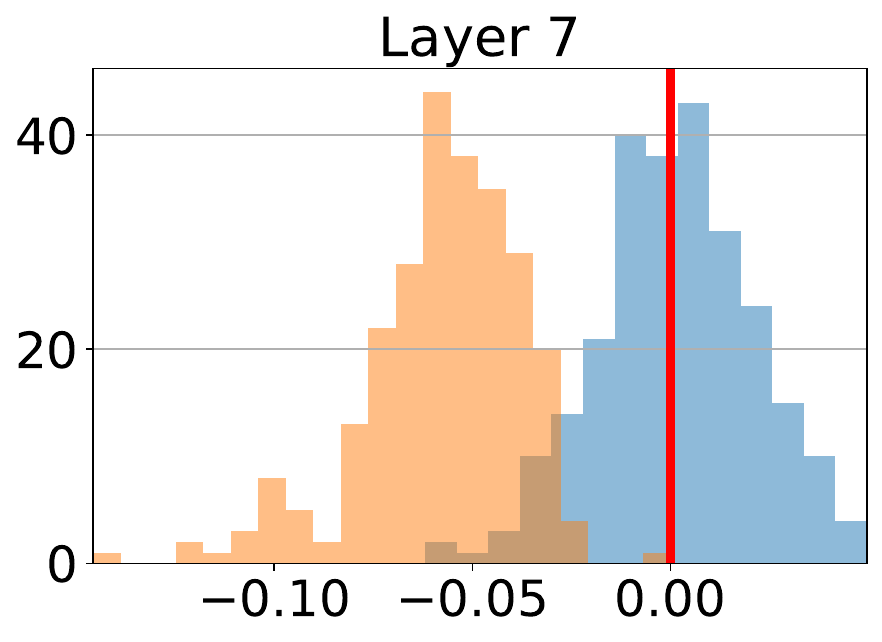}} ~
    \subfloat{\includegraphics[width=0.18\textwidth]{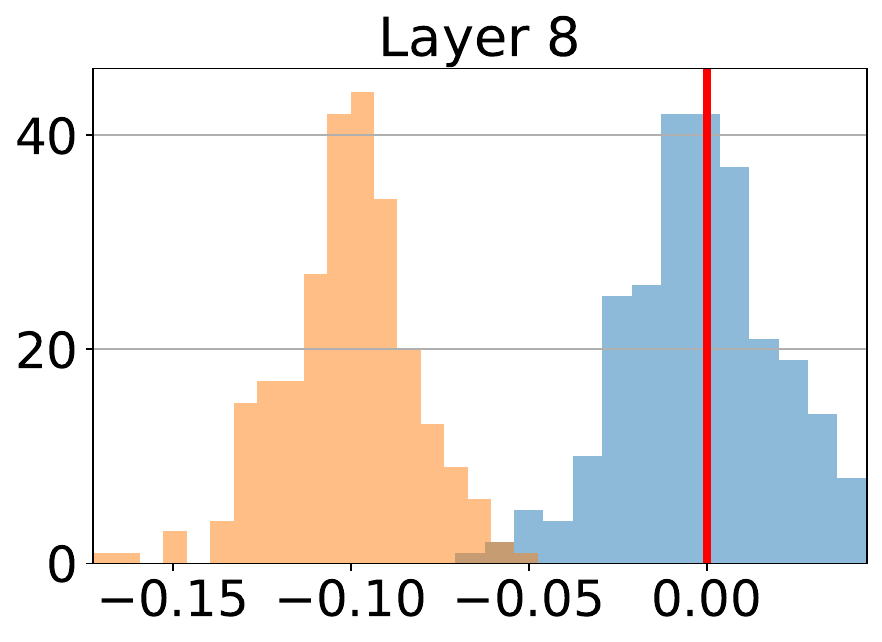}} ~
    \subfloat{\includegraphics[width=0.18\textwidth]{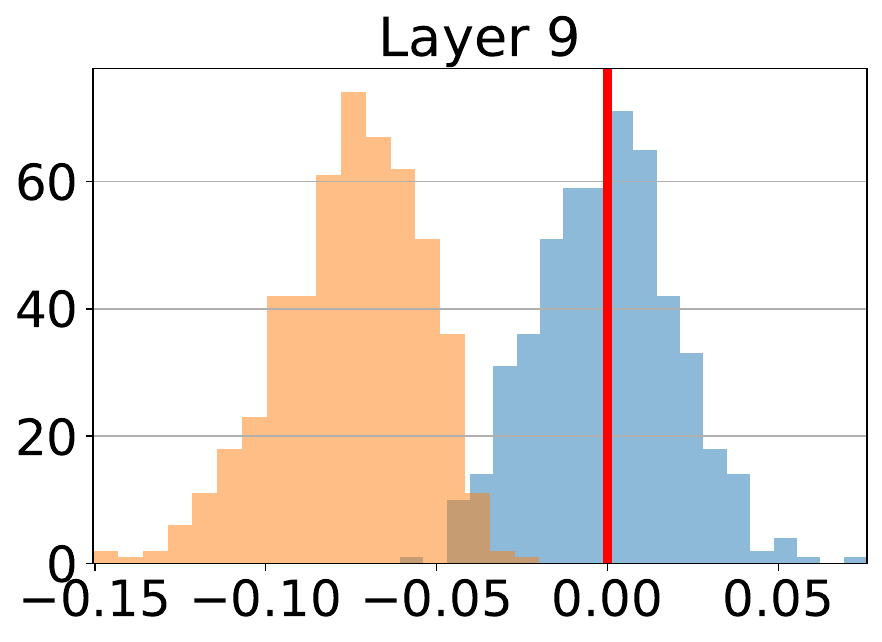}} ~
    \subfloat{\includegraphics[width=0.18\textwidth]{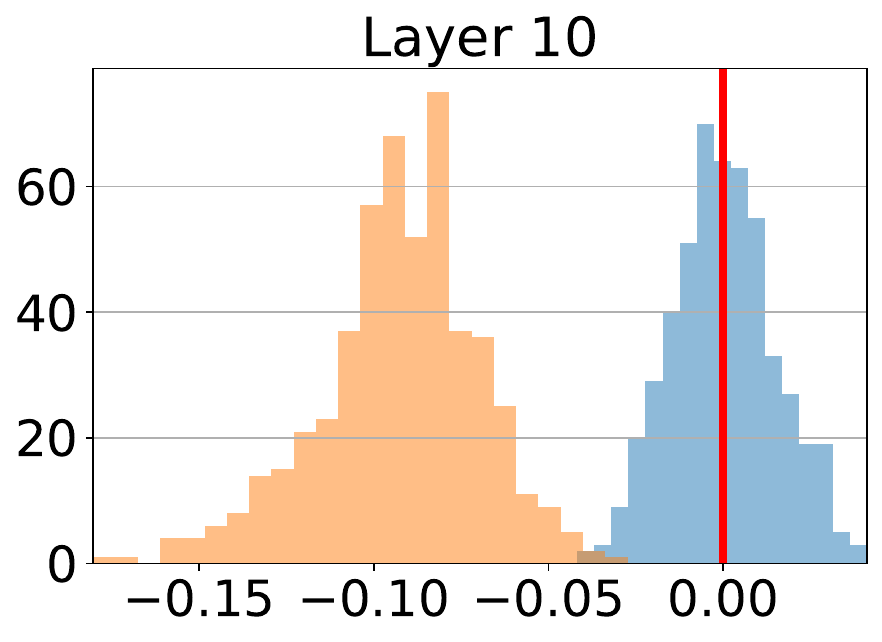}} \\
    \subfloat{\includegraphics[width=0.18\textwidth]{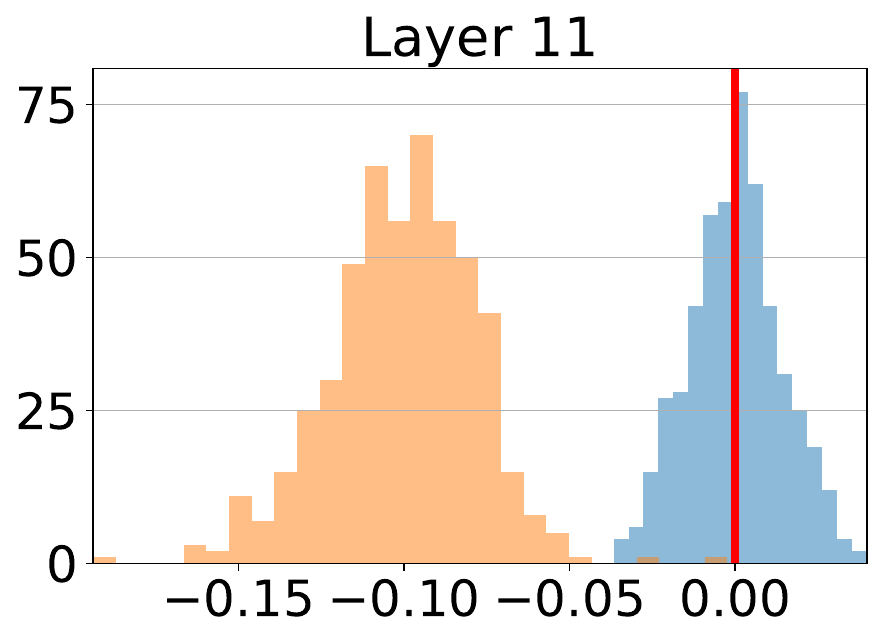}} ~
    \subfloat{\includegraphics[width=0.18\textwidth]{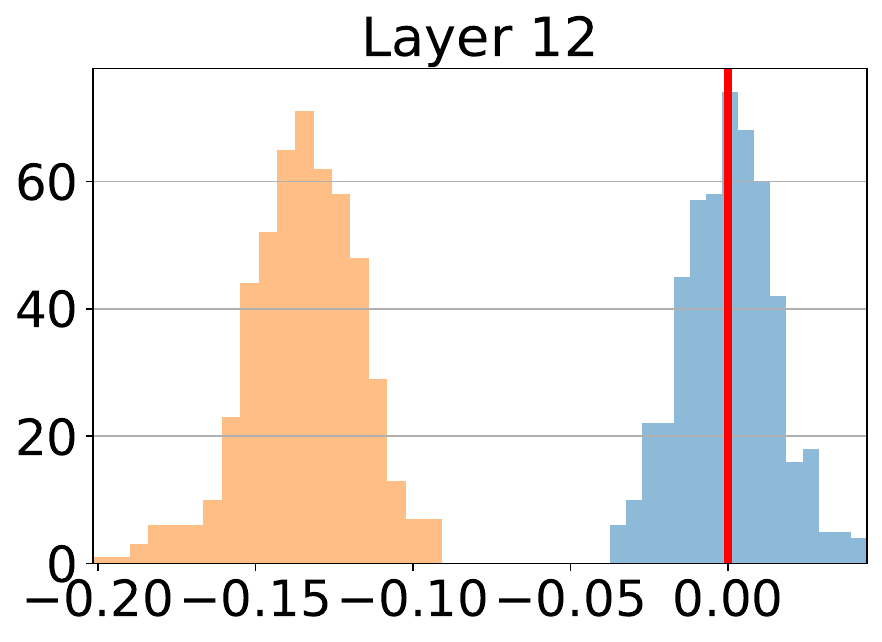}} ~
    \subfloat{\includegraphics[width=0.18\textwidth]{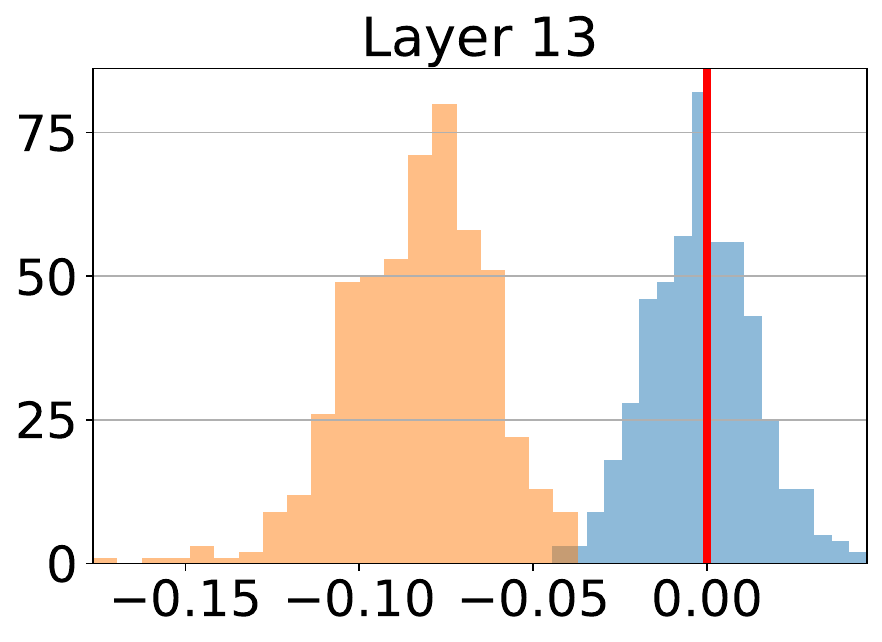}} ~
    \subfloat{\includegraphics[width=0.18\textwidth]{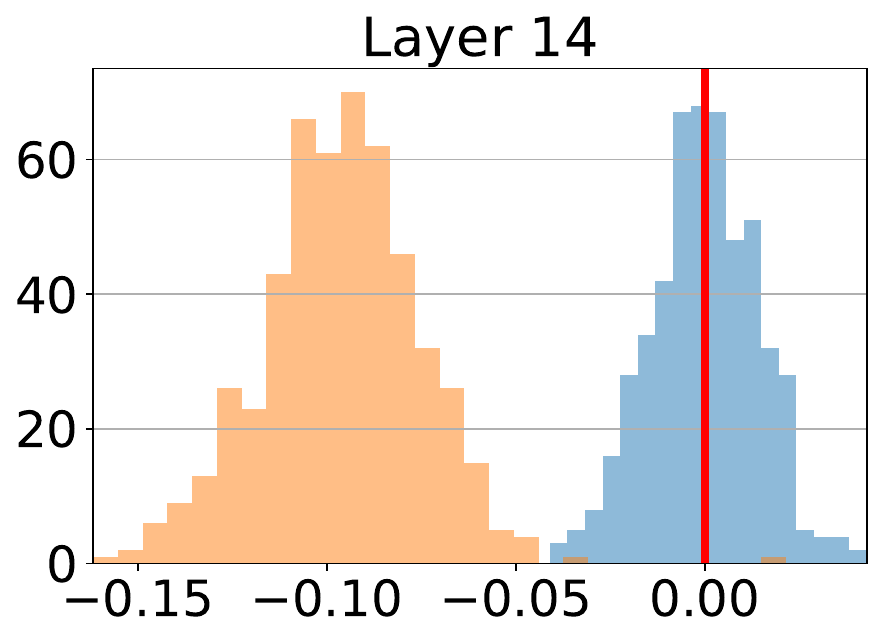}} ~
    \subfloat{\includegraphics[width=0.18\textwidth]{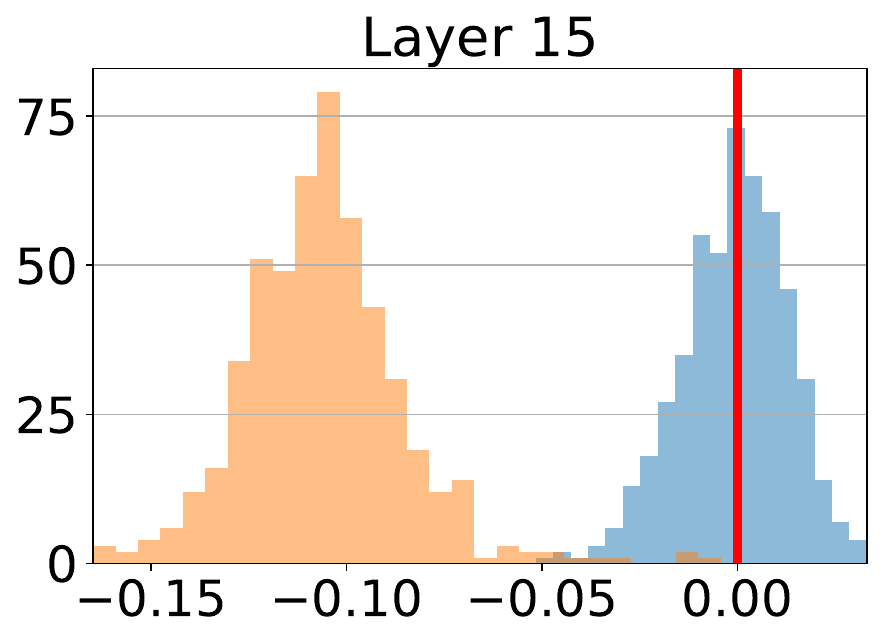}} \\
    
    \caption{\textbf{Distribution of our projection statistic} for a random initialization (blue) and for a \textit{pretrained} VGG19 on ImageNet (orange). We found corresponding statistical bias when training convolutional networks on various image classification tasks and present our empirical findings in the present work.}
    \label{fig:pretrained}
  \end{figure}

In this work, we take a geometrical perspective and look for statistical bias in the weights of trained convolutional networks, in terms of hyperplane arrangements induced by convolutional layers with ReLU activations. Notably, for networks combining linear (affine) layers with piece-wise linear activations, hyperplane arrangements define the function computed by the network and characterize how data is transformed non-linearly by the model~\cite{arora2018understanding,carlsson2016preimage,hanin2019complexity}.

Specifically, we exploit symmetries arising from parameter sharing in convolutional layers 
to describe the local arrangements of hyperplanes associated with convolutional filters at each layer, for various networks trained on image classification tasks.

More precisely, we first show that each convolutional filter at a given layer is symmetric w.r.t.\ the identity line. We then look for regularity of layer-wise hyperplane arrangements by measuring degree of agreement in orientation of individual filters along the identify line in the layer's preactivation space. We correlate this regularity measure to the learning dynamics of convolutional layers. Our main findings are summarized as follows.

\begin{figure}[htbp]
    \centering
    \subfloat{\includegraphics[width=0.33\textwidth]{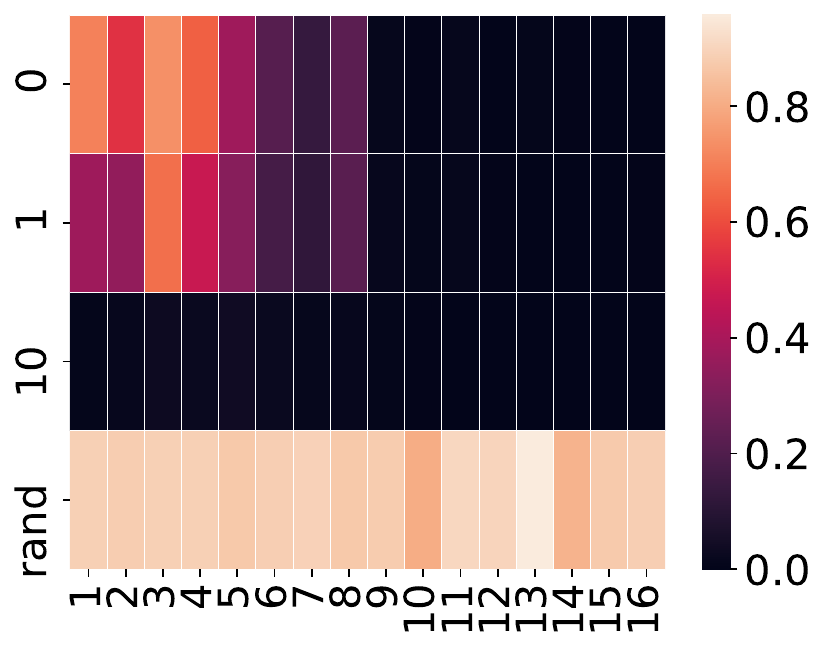}}
    \subfloat{\includegraphics[width=0.33\textwidth]{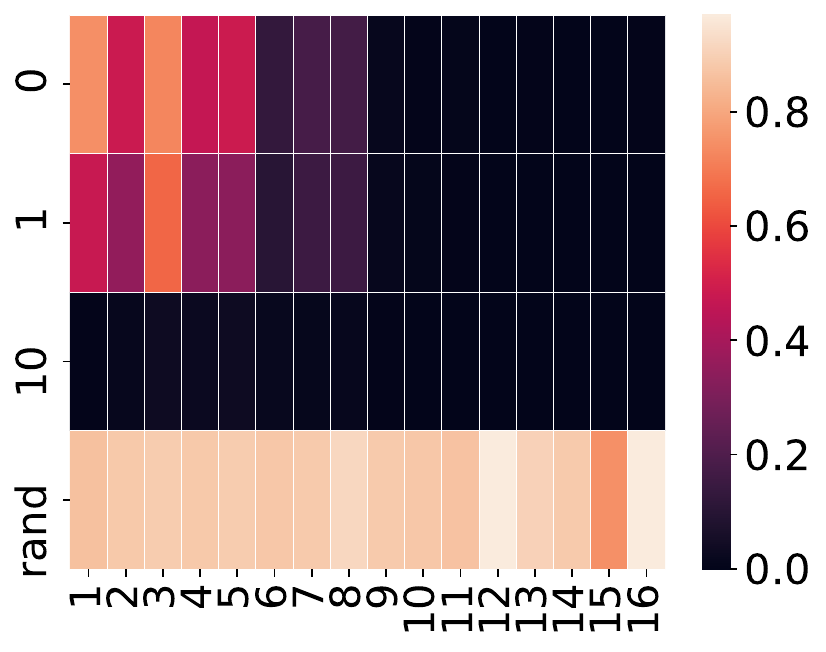}} ~
    \subfloat{\includegraphics[width=0.33\textwidth]{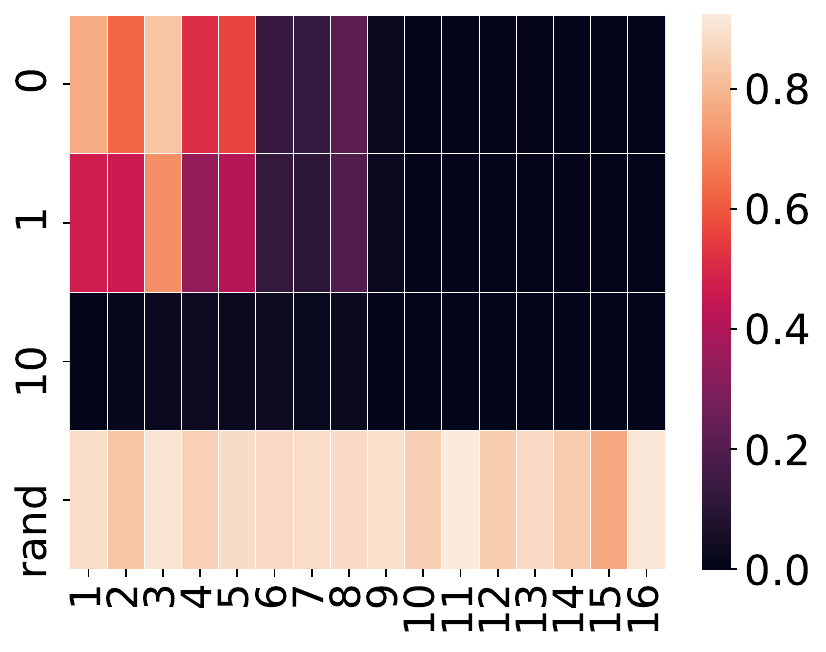}} ~
    \caption{\textbf{Weight re-initialization experiments on VGG19 trained on CIFAR10}, in which we reproduce the results of Zhang et al.\  \cite{zhang2019all}. The horizontal axis denotes the convolutional layer number, while the vertical axis represents the epoch that was used to reinitialize the weights. The colour intensity in each entry of the heatmap represents the drop in test accuracy when the weights of the corresponding layers are reset to their value at the specified epoch. Each plot corresponds to a different independent training run. On average the first 8 layers seem to be critical. Interestingly, our measure shows significantly-stronger bias for the first 8 layers (see figure~\ref{fig:cifar10}, except layer 1 which is discussed later).}
    \label{fig:reinit}
  \end{figure}

\vspace{1em}
\textit{Hyperplane arrangements of many layers of trained convolutional networks exhibit strong regularities, that emerge from training and correlate with learning (Figure~\ref{fig:pretrained}). Furthermore, for low-complexity datasets, layers presenting biased hyperplane arrangements are critical to the performance of the network -- our measure correlates with the notion of critical layers introduced in the recent intriguing work of Zhang et al.~\cite{zhang2019all} -- (Figure~\ref{fig:reinit}). That means, when the bias is not observed, the corresponding layers' weights can be reset to their value at initialization without considerable loss in performance.}
\vspace{1em}

We believe these regularities of local arrangements encode important information for understanding the implicit bias of trained convolutional networks and present a novel angle to study the problem.

The paper is organized as follows. Section~\ref{sec:method:polyhedra} and \ref{sec:method:data} present the background theory for describing local hyperplane arrangements for convolutional layers. Section~\ref{sec:method:statistics} introduces novel statistics to study the arrangements for convolutional layers. In section~\ref{sec:experiments}, we perform experiments to investigate how the measure correlates with learning. Related work are summarized in section~\ref{sec:related_works}. Finally, section~\ref{sec:conclusion} discusses future work.


%% file: sections/background.tex
\section{Method}
\label{sec:method}


We consider convolutional networks with rectifier activations (ReLU) and focus our study on convolutional layers. For each layer $l$, we analyse the arrangement of hyperplanes induced by $l$ on its preactivation space. Let $\mathcal{W}^l \in \mathbb{R}^{F \times C \times k \times k}$ denote the weight tensor of $l$, consisting of $F$ output maps, each corresponding to a kernel of size $C \times k \times k$. Then, the matrix $\mathbf{W}^l$ obtained by a sparse row-major vectorization of $\mathcal{W}^l$ has dimensionality $Fr \times CHW$, where $C, H, W$ denote respectively the number of channels, height and width of the input to the layer and $r$ is the number of receptive fields of each output map. Each row $m$ of $\mathbf{W}^l$, hereafter denoted $\mathbf{W}_m^l$, is a sparse vector with non-zero elements corresponding to the $C \times k \times k$ entries of the respective filter $F^l_i := \mathcal{W}^l_{{[}i,:,:,:{]}}$, for \mbox{$m = ir, \ldots, (i+1)r -1$}, as illustrated in figure~\ref{fig:vectorization}. Convolution with an input vector $\mathbf{X}$ is then performed as general matrix-vector multiplication, and the operation is typically augmented to an affine transformation by introducing a bias term $b_i^l$ for each filter $F_i^l$.

Crucially, parameter sharing implemented by convolutional layers imposes a circulant structure over the rows of $\mathbf{W}^l$. Particularly, if each kernel is convolved with stride $s=1$, the $m$-th row is obtained by cyclically shifting the components of the $(m-1)$-th row by one step to the right. Thus, each sub-matrix $\mathbf{W}^l_{{[}ir:(i+1)r,:{]}} \in \mathbb{R}^{r \times CHW}$ corresponding to a filter $F_i^l$, approximates a circulant matrix\footnote{The approximation is exact when $r=CHW$ and each row of $\mathbf{W}^l$ is highly sparse. For instance, this holds for single-channel convolutions ($C=1$) with stride $s=1$, zero-padding $p=2$, spatial kernel size $k=3$ and $k \ll HW$.}.

In the next section, we describe how the circulant symmetry induces polyhedra in the preactivation space of a convolutional layer. We begin by briefly considering the special case of single-channel convolutions, and then treat the more general multi-channel problem.

\subsection{Local hyperplane arrangements are described by polyhedra}
\label{sec:method:polyhedra}

  \begin{figure}[t!]
    \centering
    \subfloat{\includegraphics[width=0.5\textwidth]{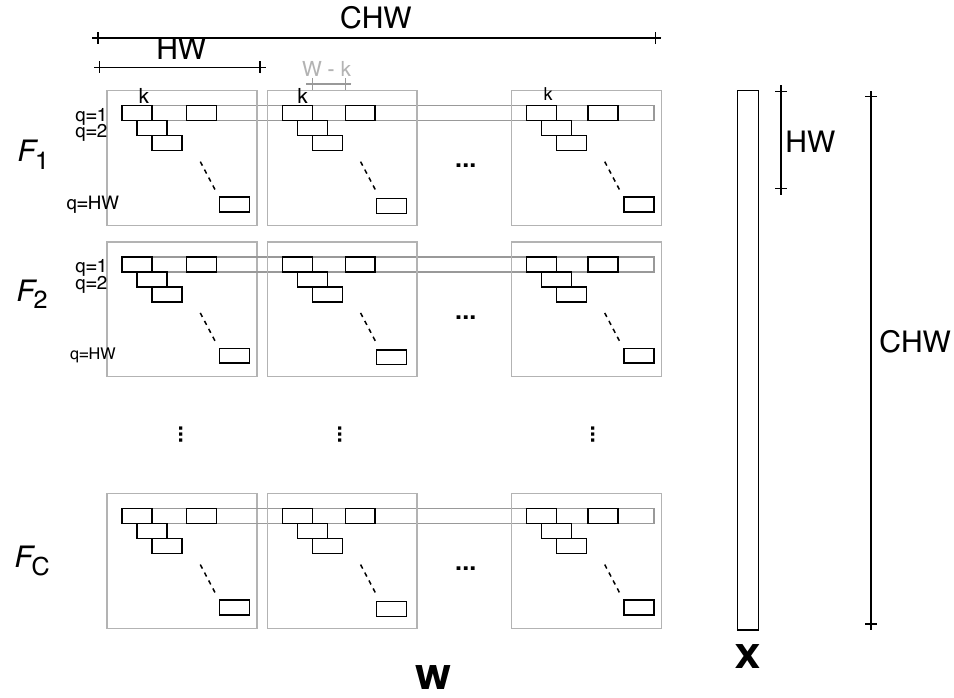}} \qquad
    \subfloat{\includegraphics[width=0.38\textwidth,trim={0 0 0 4em},clip]{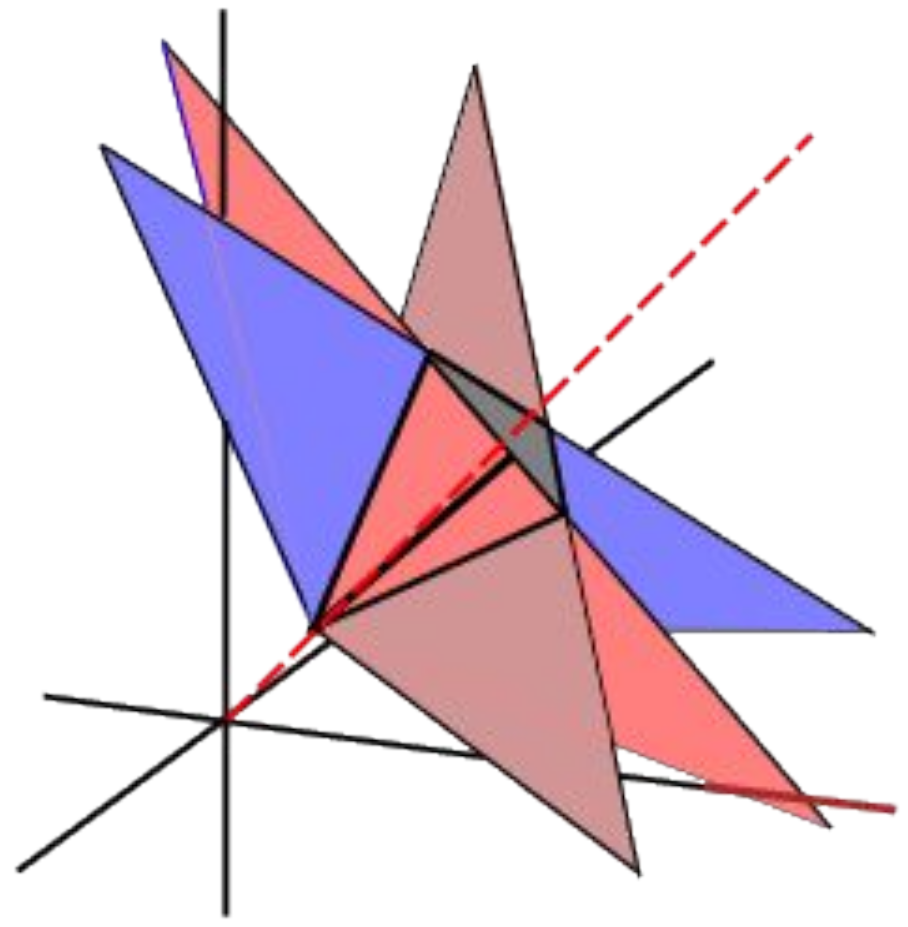}}
    \caption{(Left) \textbf{vectorization of a convolutional tensor} $\mathcal{W}$ and input $\mathcal{X}$ into a weight matrix $\mathbf{W}$ and column input vector $\mathbf{X}$, defining $r=HW$ receptive fields for each output map. (Right) polyhedral cone induced by hyperplanes with circulant symmetry around the identify line, for the case of single-channel convolutions.}
    \label{fig:vectorization}
  \end{figure}

When ReLU is applied element-wise to the output of $l$, each row $\mathbf{W}_m^l$ identifies a hyperplane splitting the preactivation space into two halfspaces, $\{\mathbf{x} : \mathbf{W}^l_m \mathbf{x} + b_i \ge 0 \}$, and its complementary, corresponding respectively to the positive and negative side of the hyperplane with normal vector ${\mathbf{W}^l_m}^T$.

When all hyperplanes induced by filter $i$ are considered at the same time, the resulting system of inequalities $\{\mathbf{x} : \mathbf{W}^l_m \mathbf{x} + b_i \ge 0, \quad m=ir, \ldots, (i+1)r\}$ defines a polyhedron $\mathcal{H}_i^l$ in the preactivation space of $l$.

For the special case of single-channel convolutions (where $F=C=1$), the matrix $\mathbf{W}^l$ induces a polyhedral cone $\mathcal{H}_i^l$  whose hyperplanes all intersect at a common apex, with circulant symmetry around the identity line of the preactivation space of $l$~\cite{carlsson2019geometry,gamba2019geometry}. The number of faces of the cone corresponds to the number of receptive fields $r$ of $F_i^l$, and is determined by the stride, zero-padding and kernel size of the convolution. Interestingly, while parameter sharing imposes very strong priors on the weights, without further constraints, the variability of the cone is still considerable, corresponding to its opening angle, apex position along the identity line and rotations fixing identity line~\cite{gamba2019geometry}.


In the general multi-channel case ($F > 1, C > 1$), each filter $F_i^l$ is described by a matrix with doubly block-circulant structure (figure~\ref{fig:vectorization}, left). Parameter sharing is implemented over the spatial dimensions $H$ and $W$, while each neuron of $l$ is fully-connected over depth. Consequently, the sub-matrix corresponding to each filter represents a circulant matrix with missing rows, inducing a polyhedron $\mathcal{H}_i^l$ which resembles a polyhedral cone with missing faces. Due to parameter sharing, the hyperplanes corresponding to the rows of each circulant submatrix preserve their symmetry around the identify line, and consequently it is still possible to consider orientation and opening angles for $\mathcal{H}_i^l$. However, without a-priori constraints, the variability of the problem increases significantly, since the intersection of all hyperplanes of $\mathcal{H}_i^l$ is no longer a point and the dimensionality of the problem increases compared to the single-channel setting.

A central question to our study is whether there exist additional constraints that are induced on the orientation of the polyhedra by learning. For the present work, we focus on studying the orientation of polyhedra $\{\mathcal{H}_i^l \}_{i=1}^F$ for each layer along the identity line, as motivated in section~\ref{sec:method:data}. In section~\ref{sec:method:statistics}, we conclude our methodology and introduce statistics over the convolutional weights of a layer to express the orientation of polyhedra.





\subsection{Positive orthant in the preactivation space}
\label{sec:method:data}

The location of polyhedra $\{\mathcal{H}_i^l \}_{i=1}^F$ in the preactivation space is central to the understanding of how data is mapped non-linearly between layers.

In fact, when all hyperplanes of filter $i$ are considered together, the intersections of the corresponding halfspaces divide the input space of $l$ into distinct cells\footnote{the exact number of cells depends on the total number of hyperplanes, their mutual position, and the dimensionality of the preactivation space where they are defined~\cite{stanley2004introduction}.}, each characterized by a distinct sparsity pattern under ReLU, corresponding to whether the input $\mathbf{x}$ is located on the positive or negative side of each hyperplane $\mathbf{W}^l_m\mathbf{x} + b_i^l \ge 0$.

Interestingly, due to the circulant structure arising from parameter sharing~\cite{carlsson2019geometry}, each filter $i$ defines two cells that are readily interpretable: the intersection of all negative half-spaces, which ReLU contracts to a single point, and the intersection of all positive half-spaces, which is instead transformed linearly. In section~ \ref{sec:method:statistics}, we introduce \textit{projection statistics} to study the distribution of hyperplanes defining such cells with respect to the positive orthant of the input space of a layer.

In general, whether a specific cell in the input space of a layer, and ultimately whether any linear region in the input space of the network, is relevant to characterize the function computed by the network, depends on the distribution of image data in the input space. For the present study, the role of training data is considered indirectly, by analysing the weights of convolutional networks trained in practice.

Without making special assumptions on the actual position of data samples with respect to the hyperplanes induced by each filter $F_i^l$, we observe that:
\begin{itemize}
  \item in pixel space, according to the strategy used, normalized pixel values are either represented in the range ${[}-1,1{]}$ or as positive values.
  \item for all other layers $l = 2, \ldots, L$, ReLU maps the activation of layer $l-1$ to the positive orthant of the preactivation space of $l$, that is $\{ x_k \ge 0, \forall k=1, \ldots, CHW\}$.
\end{itemize}

Furthermore, as described in the previous section, each filter $F^l_i$ has circulant structure around the identity line and hence, the all-positive and all-negative cells have non-empty intersection with the identity line. Since such a line is partly contained by all polyhedra, we choose it to study the agreement in orientation of $\{F_i^l\}_{i=i}^F$, which we use as a proxy for finding strongly biased arrangements.

Therefore, we focus our study on the intersection of the identity line with the positive orthant of the preactivation space of $l$
$$\mathcal{I}^+ = \{x_k = x_{k+1}, \quad k = 1, \ldots, CHW -1\} \cap \{ x_k \ge 0, \forall k=1, \ldots, CHW\}.$$
More precisely, we consider all layers $1, \ldots, l-1$ as a fixed feature extraction step, and study layer $l$ using the standard basis of its preactivation space $\mathbb{R}^{CHW}$.

We note that, in principle, it would be possible to remove the preference for $\mathcal{I}^+$, but this would in turn require to study the location of input data with respect to the hyperplanes, in order to understand which dimensions (and directions) are meaningful. We leave this direction as interesting future work.

\subsection{Projection statistics}
\label{sec:method:statistics}

To track strongly biased arrangements, we introduce projection statistics over the weights of a convolutional layer $l$ and study for each filter whether the all-negative or the all-positive cell is aligned with the positive orthant of the preactivation space of $l$ along the positive direction spanned by the identity line. As presented in section~\ref{sec:method:polyhedra}, this relates directly to the contraction properties of the non-linear mapping learned by the layer.

To do so, $\mathcal{I}^+$ is parametrized by $\lambda(1, \ldots, 1) = \lambda\mathbf{1}$, with $\dim{(\mathbf{1})} = CHW$ and $\lambda > 0$. We recall that, due to the circulant symmetry, either the all-positive or all-negative cell of each filter must contain asymptotically  $\mathcal{I}^+$. Then, for each filter $F_i^l, i = 1,\ldots, F$, we compute the inner product $\langle \hat{\mathbf{W}_m^l}^T, \hat{\mathbf{1}} \rangle$ between the unit normal vector $\hat{\mathbf{W}_m^l}^T$ to any hyperplane induced by filter $F_i^l$ and the unit vector $\hat{\mathbf{1}}$.

We observe that, since each row $\mathbf{W}_m^l$ of the vectorized tensor $\mathbf{W}^l$ is a sparse vector with non-zero entries corresponding exactly to the $C \times k \times k$ parameters of the $i$-th output map, $m = ir, \ldots, (i+1)r$, the inner  product can be efficiently computed as:

\begin{equation}
\label{eq:sum}
  \begin{aligned}
  \langle \hat{\mathbf{W}_m^l}^T, \hat{\mathbf{1}} \rangle &= \frac{1}{A}\sum\limits_{c=0}^{C-1}\sum\limits_{p=0}^{k-1}\sum\limits_{q=0}^{k-1} \mathcal{W}_{{[}i,c,p,q{]}}
  \end{aligned}
\end{equation}
with $A = ||{\mathbf{W}_m^l}^T||_2 \cdot ||\mathbf{1}||_2$ is the product of the norms of the two vectors.

\subsubsection{Interpretation}


The inner product between the unit normal vector $\hat{\mathbf{W}_m^l}^T$ and unit vector $\hat{\mathbf{1}}$, measures the cosine opening angle of the polyhedron $\mathcal{H}_i^l$ w.r.t.\ the identity line $\mathcal{I}$. For the purpose of studying orientation, we can assume all bias terms $b_i^l$ to be zero, since they do not affect the cosine angles computed.

Importantly, the sign of the cosine angle expresses whether any point $\lambda\mathbf{1},~\lambda >0$, is on the positive or negative side of all hyperplanes of $\mathcal{H}_i^l$. Due to the circulant symmetry, every hyperplane induced by a filter $F_i^l$ shares the same angle with $\hat{\mathbf{1}}$, and hence the same cosine angle.

Additionally, the sign of the cosine angle is a coarse indicator of the orientation of $\mathcal{H}_i^l$ with respect to the direction spanned by $\mathbf{1}$, and expresses whether the all-positive (resp.\ all-negative) cell contains $\mathcal{I}^+$.

Notably, if the cosine angles for all filters $F^l_i, i = 1, \ldots, F$ of layer $l$ agree in sign, then the corresponding polyhedra $\mathcal{H}_i^l$ all agree in orientation along the direction of the identity line, indicating that the hyperplane arrangement for $l$ exhibits regularity. To measure the level of agreement, we compute the cumulative density of negative cosine angles, as described in the next section.

Finally, we note that the angle does not depend on the bias term $b_i$ of the filter and its value is the same for every parallel vector $\lambda\mathbf{1}$.


\subsubsection{Cumulative density of negative projections}
Let $\mathbf{c}^l = (c_1^l, \ldots, c_F^l)$ denote the observed cosine angles for each filter $F_i^l$, for $i=1, \ldots, F$. Abusing notation, we define 
\begin{equation}
\label{eq:statistic}
n_l := \mathbb{P}(\mathbf{c} \le 0) = \frac{1}{F} \sum\limits_{i=1}^F \mathbbm{1}_{c_i^l \le 0}
\end{equation}
the empirical cumulative distribution of negative projections, where $\mathbbm{1}_e$ denotes the indicator function for event $e$.

In section~\ref{sec:experiments}, we study $n_l$ for convolutional networks trained on various image classification datasets, and show
a statistical bias towards all-negative projections, for hyperplane arrangements induced by all filters of some of the layers. For simplicity, in the following sections we will refer to $n_l$ simply as our \textit{measure}.

%% file: sections/experiments.tex
\section{Experiments}
\label{sec:experiments}


We study the cumulative density of negative projections for VGG19~\cite{simonyan15very} and AlexNet~\cite{krizhevsky2012imagenet} trained on image classification tasks of different complexity, namely CIFAR10, CIFAR100~\cite{krizhevsky2009learning} and ImageNet~\cite{deng2009imagenet}\footnote{The source code to reproduce our experiments is available at \url{https://github.com/magamba/hp_arrangements}.}.


The rest of the paper is organized as follows. Section~\ref{sec:experiments:setup} summarizes our experimental setup. In section~\ref{sec:experiments:learning}, we present our main finding and provide an intuitive explanation for the different trend observed for the first layer of all networks. Section~\ref{sec:experiments:critical} highlights a correlation between \textit{critical layers} and biased arrangements, while section~\ref{sec:experiments:dynamics} discusses the relationship between the observed bias and the learning dynamics.

\subsection{Experimental setup}
\label{sec:experiments:setup}


All networks are trained using SGD with momentum $0.9$, weight decay coefficient of $0.0005$, and step learning rate schedule with base learning rate of $0.01$. No batch normalization layers are used and all layers are initialized using Kaiming normal initialization~\cite{he2015delving}. All bias terms are initialized to zero. Unless otherwise noted, we use the hyper-parameters reported in~\cite{simonyan15very}. To report the training performance in a single plot, train and validation losses are rescaled by their highest respective value during training. Validation loss and accuracy are computed on a split of 10\% of the training set. Accuracy on the test set is not reported since it differs by at most 1\% from the validation accuracy for all experiments. Unless stated otherwise, we run $3$ independent seeds (randomizing the initialization and training samples' order) for each experiment and report the mean observed measure with shaded areas corresponding to one standard deviation.

  \begin{figure}[b!]
    \centering
    \subfloat{\includegraphics[width=0.18\textwidth]{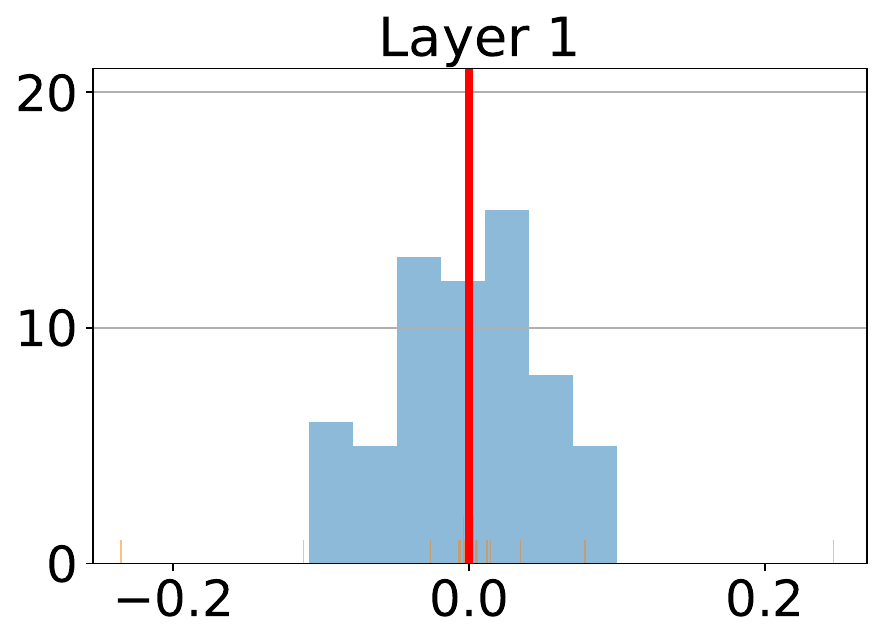}} ~
    \subfloat{\includegraphics[width=0.18\textwidth]{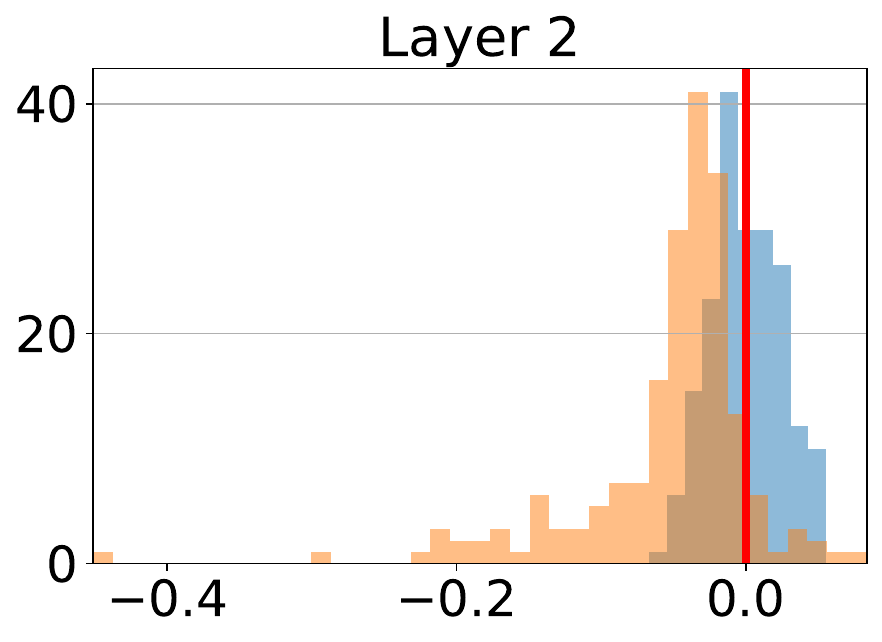}} ~
    \subfloat{\includegraphics[width=0.18\textwidth]{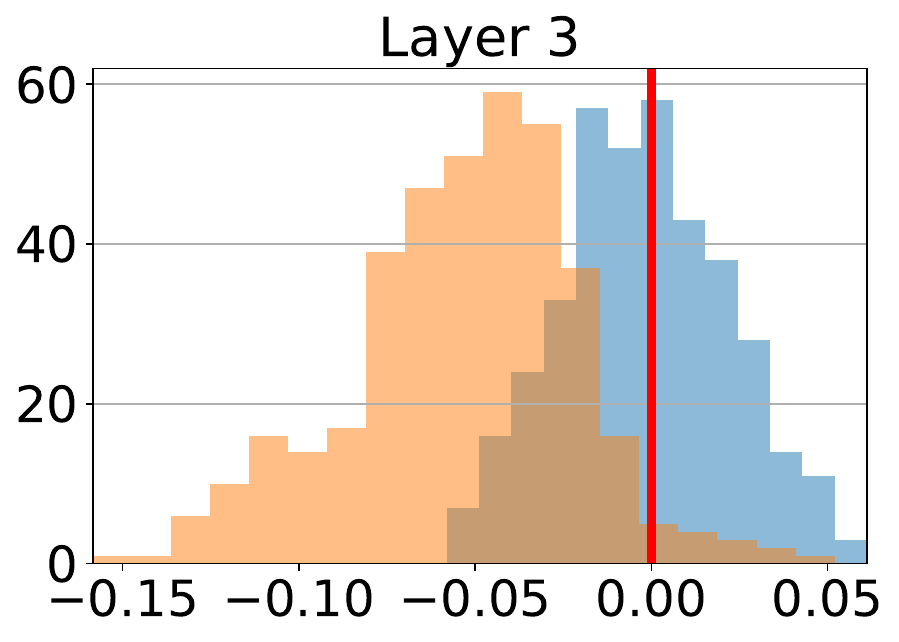}} ~
    \subfloat{\includegraphics[width=0.18\textwidth]{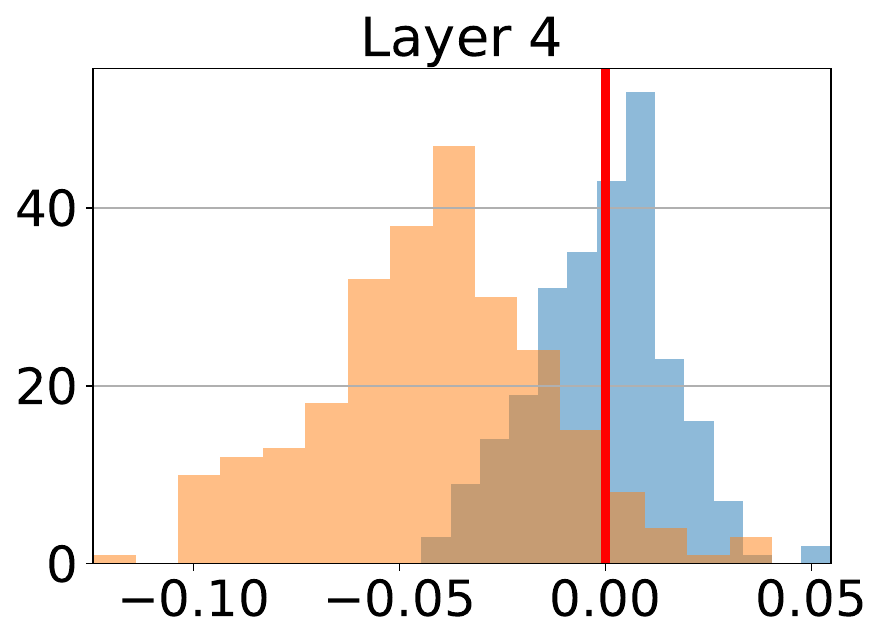}} ~
    \subfloat{\includegraphics[width=0.18\textwidth]{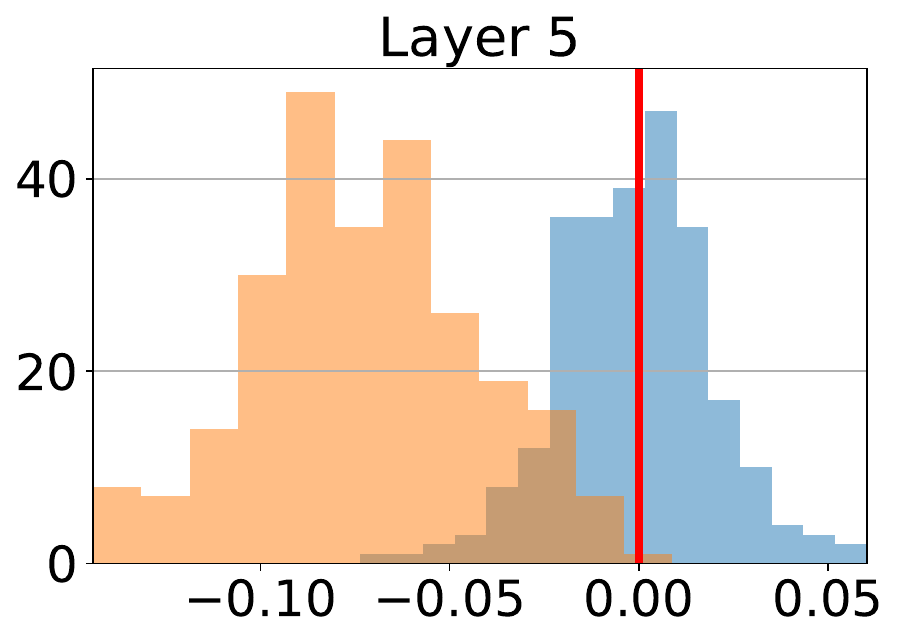}} 
    \caption{\textbf{Distribution of cosine angles for AlexNet trained on ImageNet} at initialization (blue) and after convergence (organge). Aside from layer 1 (top left plot), the distribution of cosine angles is shifted towards negative values, indicating a statistical bias in each layer's hyperplane arrangements.}
    \label{fig:histograms}
  \end{figure}

\subsection{Bias emerges from learning}
\label{sec:experiments:learning}



In figure~\ref{fig:histograms}, for each layer of AlexNet trained on ImageNet, we plot the distribution of cosine angles for all filters, computed according to equation~\ref{eq:sum}, and compare the observed values at initialization and after convergence.

At initialization, each weight in the network is drawn independently from a Gaussian distribution $\mathcal{N}(0, \sigma_l^2)$, with $\sigma_l^2$ proportional to the fan-out of each neuron in the corresponding layer $l$. From the central limit theorem, we have that the inner product defined in equation~\ref{eq:sum} is normally distributed with zero mean, as reflected in the histograms of each layer at initialization (blue). After convergence (orange), the distribution of cosine angles is shifted towards negative values, for all layers except the first. As highlighted in section~\ref{sec:method:statistics}, this indicates that the negative cell induced by each filter is aligned with the positive direction defined by $\mathcal{I}^+$. Such a level of agreement for all filters suggests that additional constraints on the hyperplane arrangements are induced during training.

\begin{figure}[t!]
    \centering
    \includegraphics[width=0.7\textwidth]{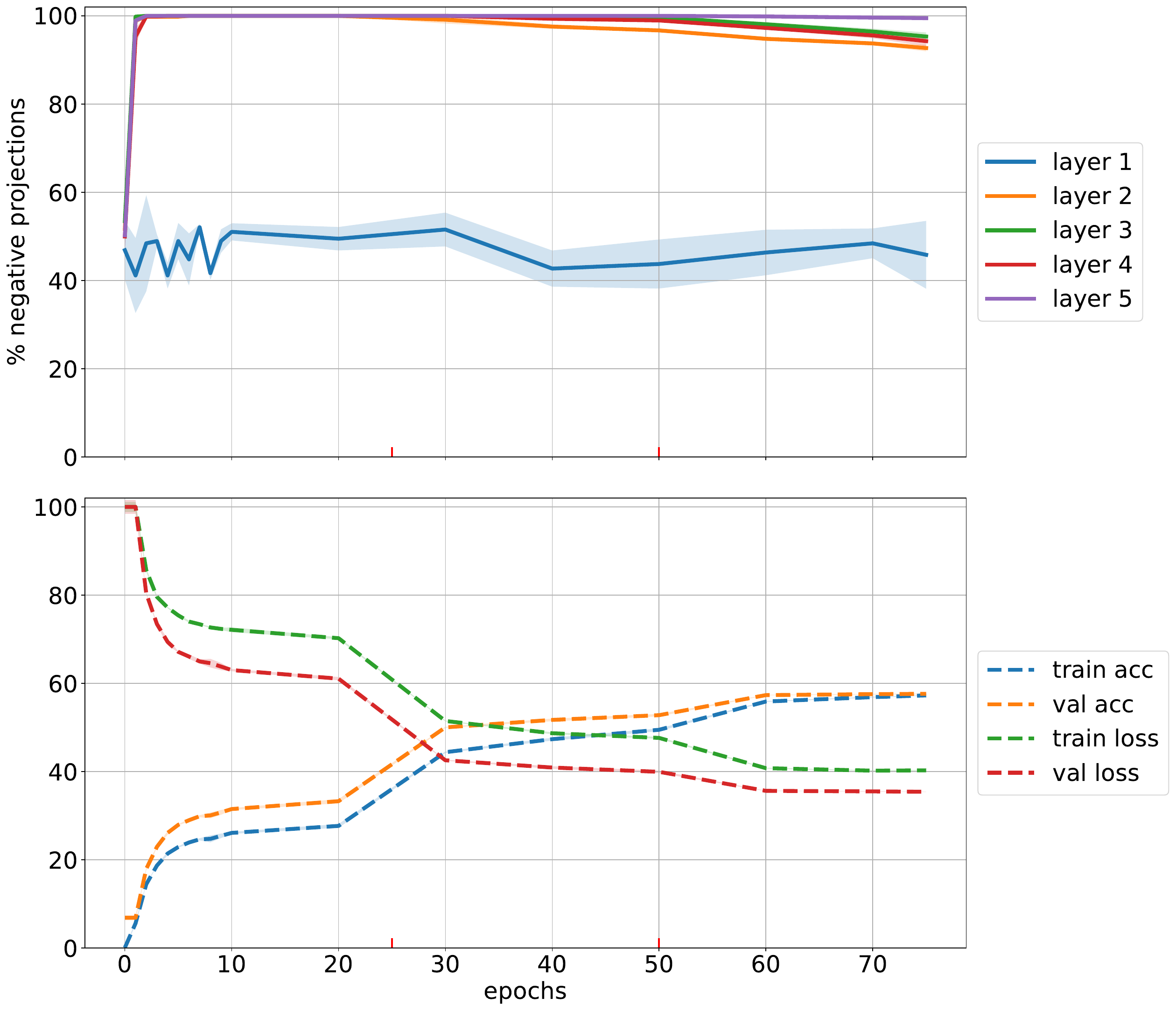}
    \caption{(Top) \textbf{cumulative density of negative projections for AlexNet on ImageNet} throughout training, with epoch 0 denoting initialization. During training, all layers beyond the first quickly become heavily biased towards negative projections. (Bottom) train and validation loss and accuracy.}
    \label{fig:alexnet_epochs}
    \vspace{-1.5em}
  \end{figure}

To study this hypothesis in more detail, in a second experiment (figure~\ref{fig:alexnet_epochs}) we track the development of the cumulative density of negative projections across epochs for each layer of AlexNet. During training, all layers past the first start from a mixed distribution ($\sim$50\% negative projections) and rapidly shift towards all-negative projections. Due to the hardness of ImageNet classification, we hypothesize that the network is making use of its full capacity in order to fit the dataset and this is reflected in the change in distribution of all layers past the first. This conjecture is further explored in section~\ref{sec:experiments:critical}, where we study the behaviour of a high-capacity VGG network on a lower-complexity dataset.

Throughout training, the first layer oscillates around 50\% negative projections, indicating that the hyperplane arrangements induced on the input space by each filter of the first layer have disagreeing orientations. The same phenomenon is consistently observed for all trained networks considered in this work.

\noindent \textbf{Low level features} An intuitive argument for explaining the observed behaviour of the first layer is inspired by visualizing the learned filters of AlexNet on ImageNet, as illustrated in~\cite{krizhevsky2012imagenet}, figure~3. For low level features, which typically correspond to edge detectors~\cite{matthew2014visualizing}, the network is intuitively required to recognize a pattern and its negative, since both are likely to appear in the data and would be informative. Hence, it is likely that the network would learn in its first layer a filter as well as its negative.

Conversely, deeper-layer filters, which respond to more-abstract patterns (e.g.\ a face), have no use in detecting the nonexistence of an abstract pattern, since it is unlikely to be present in natural data. In summary, the presence of low-level features and their inverse at the first layer might explain the mixed weight distribution (resulting in mixed projection statistics) observed in our experiments, and potentially why such distribution is preserved throughout training.

  
\noindent \textbf{Bias correlates with learning} A natural question is whether the observed bias is an artifact of optimization, without implication for learning, or if it instead pertains to successful training. To assess this, we consider a network that fails to converge and study the cumulative distribution of projections during training. More precisely, we ask whether the observed shift towards negative projections could occur even when training fails to pick a meaningful signal from data.
  \begin{figure}[t!]
    \centering
    \includegraphics[width=0.7\textwidth]{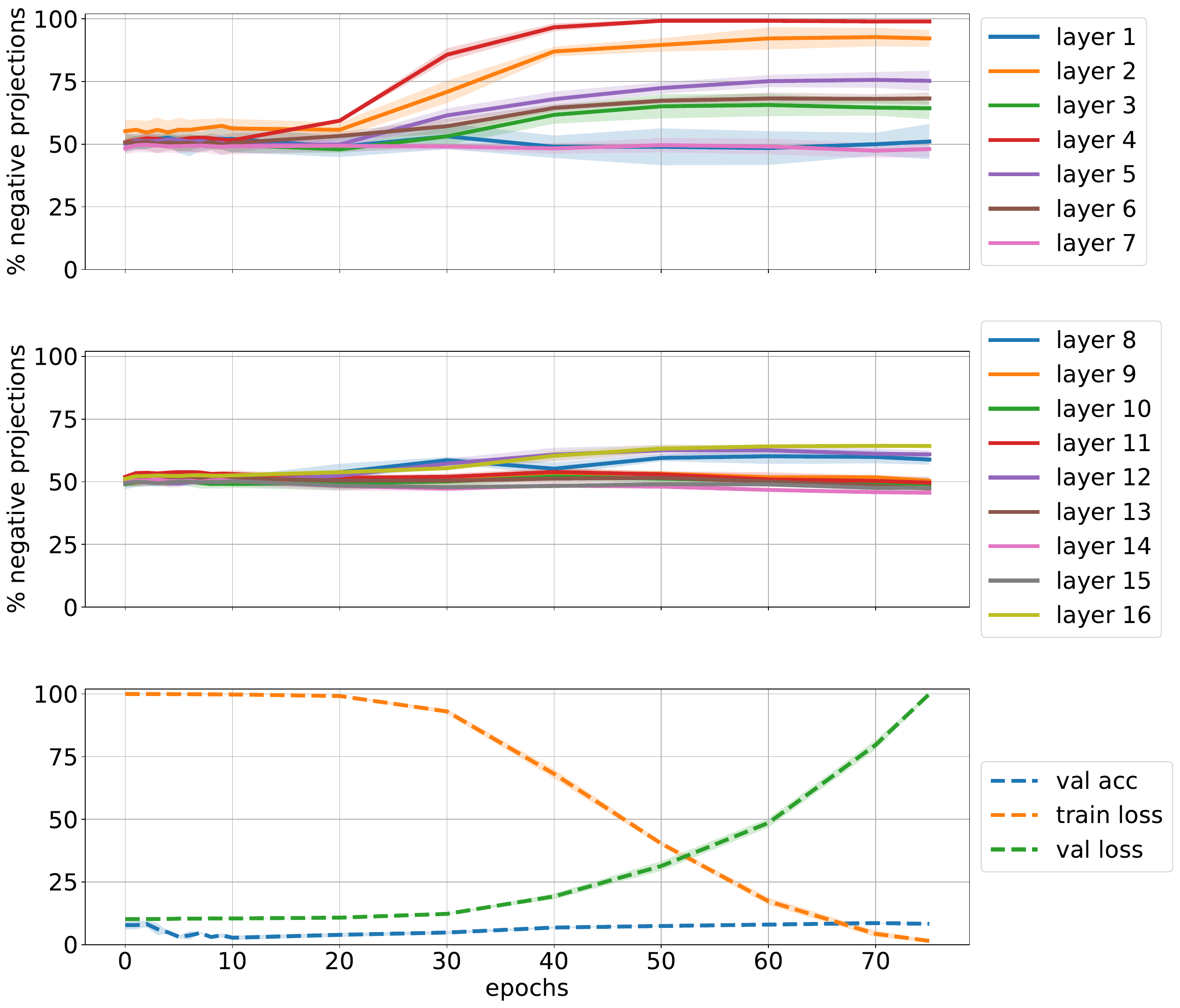}
    \caption{VGG19 on CIFAR10, where \textit{all} training labels have been replaced by noisy ones. Following~\cite{zhang2018understanding}, learning rate is $0.01$ with a smooth exponential decay of $0.95$ used to ensure complete fitting of the data. All explicit regularizations (data augmentation, weight decay) are disabled for this experiment.}
    \label{fig:noise}
  \end{figure}

\noindent In the first version of~\cite{simonyan15very}, the authors report that VGG19 fails to converge on ImageNet if weights are sampled from $\mathcal{N}(0, 10^{-2})$. Using the same initialization scheme, we train the network for $10$ epochs on ImageNet, and compute the cumulative density of negative projections throughout training.  We focus on the last $8$ convolutional layers of the network, whose weights have been changed from initialization by gradient updates, and observe that cumulative density for all layers is constant at $50\%$; its value at initialization. A similar behaviour is observed for the first $8$ layers as well, which are not reported, since their inability to learn is likely due to the vanishing gradient problem. Since the observed projections are unbiased, we argue that the statistical bias observed for networks that converge is more likely correlated with \textit{learning} than with optimization.


\noindent \textbf{Off-the-shelf pretrained models}
We were able to observe a statistical bias towards either all-negative or all-positive projections also for off-the-shelf pretrained models on ImageNet published by the Pytorch and MatConvNet communities. We include an example in figure~\ref{fig:pretrained}, showing how the cosine angles shift towards negative values at the end of training. Since snapshots at initialization were not available for pretrained models, we plot for reference the statistics for one random initialization.

In our experiments (figure~\ref{fig:alexnet_epochs}), we were able to reproduce the behaviour observed for pretrained AlexNet (released by Pytorch) and pretrained VGG19 (MatConvNet), whose trend matches the last epoch of our implementation (figure~\ref{fig:positive_bias}). Additional histograms are included in the supplemental material.

\noindent \textbf{Bias correlates with learning under noisy labels}
Lastly, we study the learning dynamics under 100\% noisy labels on CIFAR10 and report our findings in figure~\ref{fig:noise}. We observe that, for the early epochs of training, as long as the training loss is approximately constant, the projection statistics are largely unbiased. Around epoch $20$, when the training loss starts decreasing, layers $2$ to $5$ start assuming more biased hyperplane configurations, which also in this case seems to correlate with fitting to the data.

Additional experiments for noise in the data (rather than in the labels) are presented in the supplemental material.

\subsection{Biased layers are critical to performance}
\label{sec:experiments:critical}
We now consider lower-complexity classification tasks and study the role of biased layers for high-capacity VGG19 on CIFAR10. In a recent paper, Zhang et al.~\cite{zhang2019all} introduce the notion of critical layers, whose learned parameters are crucial to generalization, and contrast it to ambient layers, whose parameters can be reinitialized to their value at the start of training, without a significant drop in performance. For low-complexity datasets and high-capacity models, the critical role of a limited number of layers is posited to indicate some form of implicit capacity control at play during training~\cite{zhang2019all}. 

To study the potential relationship between the regularity of hyperplane arrangements for certain layers and implicit regularization, we begin by reproducing the main result of~\cite{zhang2019all}. Our results are reported in figure~\ref{fig:reinit}, and agree with what was shown in the original paper, i.e.\ that for CIFAR10, only the first $8$ layers of VGG19 are critical, while the remaining convolutional layers can be reset to their value at initialization without considerable performance loss.

\begin{figure}[t!]
    \centering
    \includegraphics[width=0.7\textwidth]{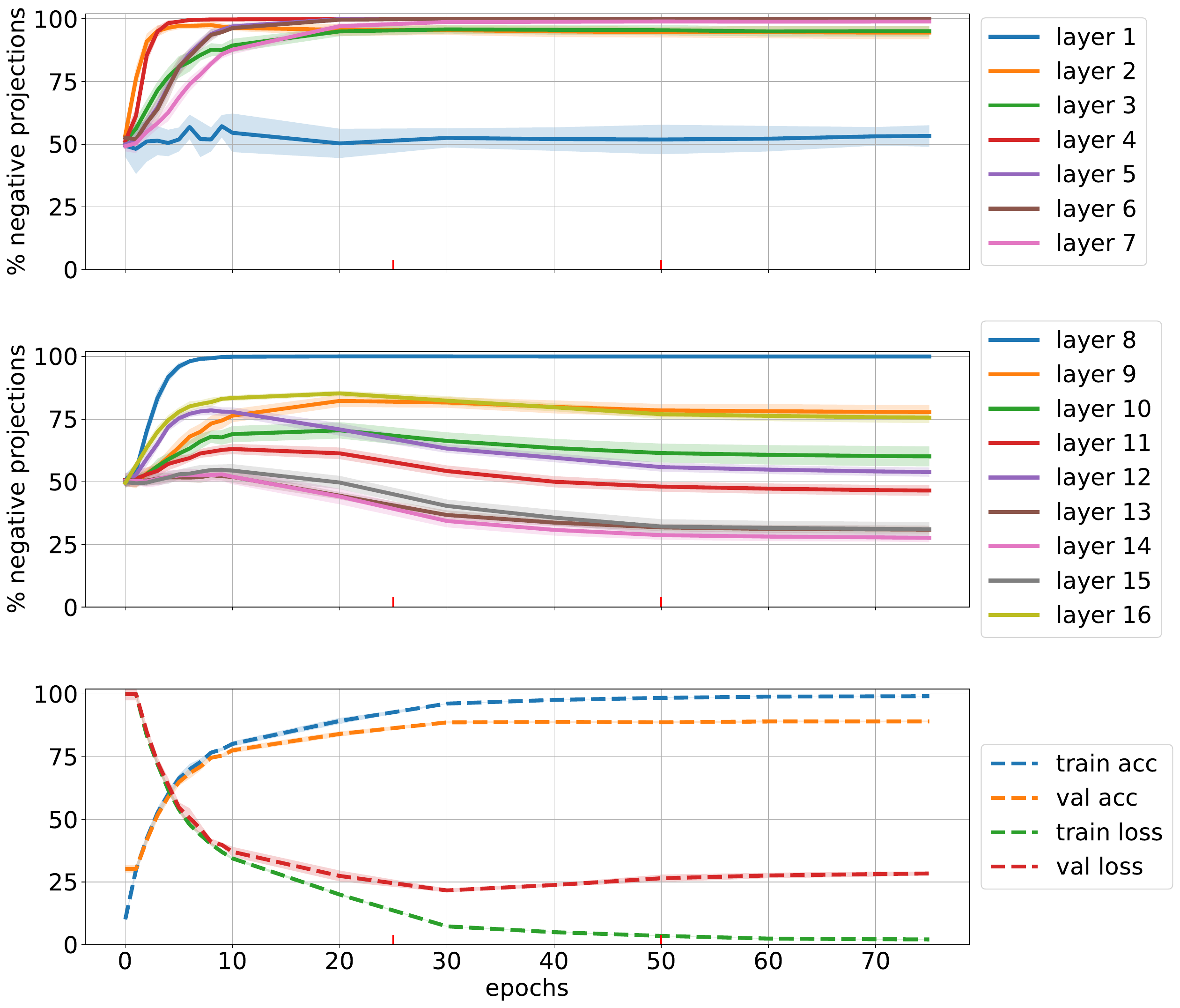}
    \caption{Cumulative distribution of \textbf{negative projections for VGG19 on CIFAR10}, averaged over 10 independent runs. The top plot shows the development of the first $7$ layers over epochs, while the middle plots shows convolutional layers $8-16$. The bottom plot reports the train and validation loss and accuracy.}
    \label{fig:cifar10}
  \end{figure}
  
Next, we compute our projection statistic for each convolutional layer in the network. In figure~\ref{fig:cifar10}, we show that layers $2$-$8$, which play a crucial role for generalization, show also heavily biased configurations. To further the statistical significance of the observation, we report averaged results over $10$ runs.



  
The experiment shows that layers with hyperplane arrangements that are biased towards regular configurations are critical to the performance of the network, and hence the observed regularity might encode important information to understand the implicit regularization and capacity control performed during learning. We think this interesting observation provides a potential angle to study the findings in \cite{zhang2019all}.

In the supplemental material, we further extend this experiment to VGG19 on CIFAR100, and show that with a harder learning task, more layers become biased and, in turn, critical to performance.



\subsection{Learning dynamics}
\label{sec:experiments:dynamics}

  \begin{figure}[t!]
    \centering
    \includegraphics[width=0.7\textwidth]{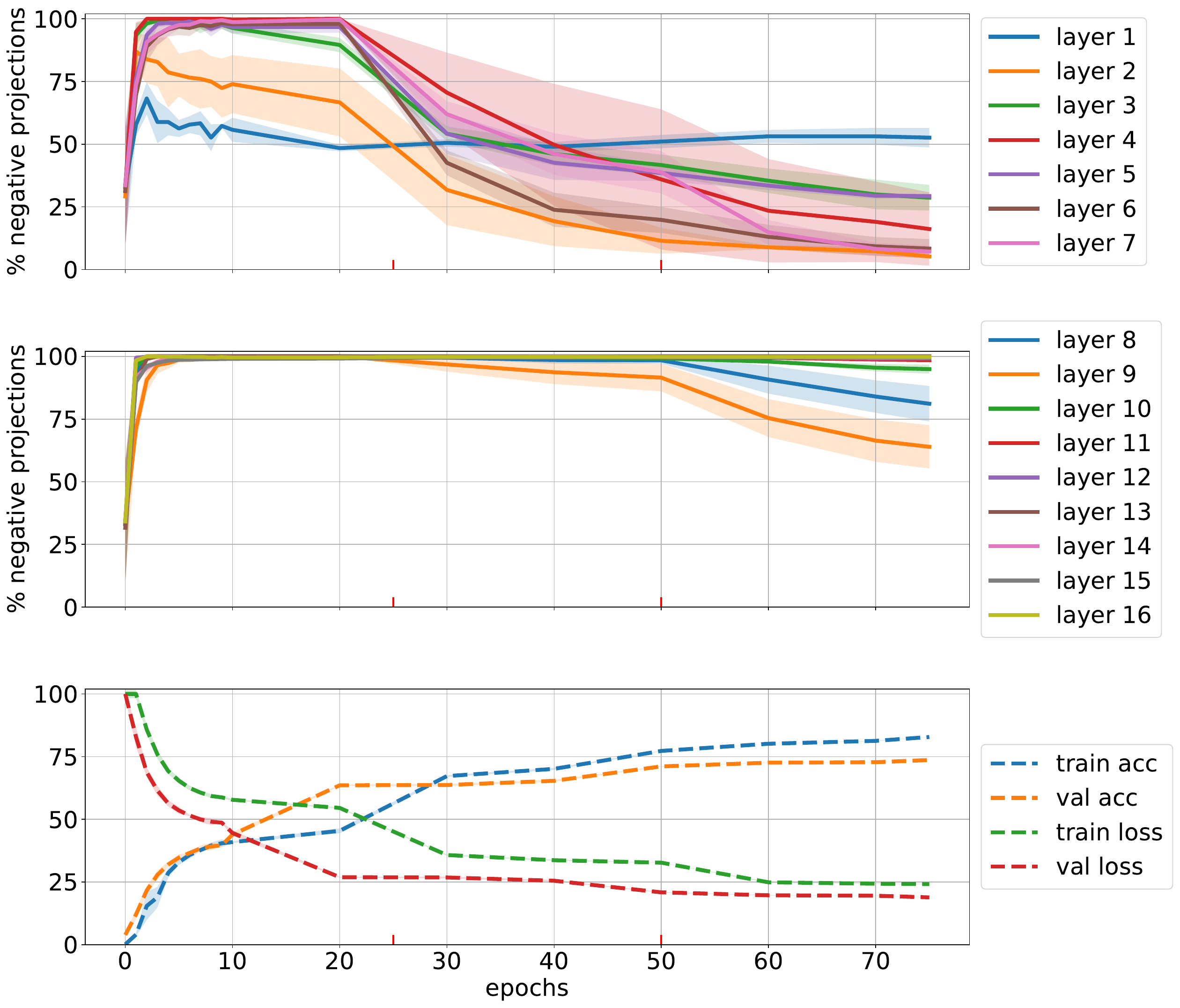}
    \caption{Cumulative distribution of negative projections for VGG19 on ImageNet for layers $1$-$7$ (top) and $8$-$16$ (middle). (bottom) train/val dynamics}
    \label{fig:positive_bias}
    \vspace{-1em}
  \end{figure}

Throughout our experiments, it is interesting to note how the low standard deviation across multiple runs closely follows the mean measure for most of the layers, amplifying the statistical significance of our result, and suggesting that our measure reflects biased learning dynamics. The same observation does not hold in general for all layers for large datasets, as shown for example in figure~\ref{fig:positive_bias}.

We now turn our attention to VGG19 trained on ImageNet and report the projection statistics in figure~\ref{fig:positive_bias}. We observe that, during the early epochs of training, almost all layers show a strong bias towards negative projections. Then, an intriguing phenomenon is observed, -- the cumulative distribution for layers $2$ to $7$ changes, and towards the end of training layers $2, 6$ and $7$ show a strong bias towards positive projections, indicating that the all-positive cells of all filters in these layers are now aligned along $\mathcal{I}^+$.

To provide a possible intuitive argument for the difference in behaviour compared to AlexNet, we focus on the training loss and accuracy for the two models, keeping in mind that both networks share the same hyperparameter settings and seeds for our experiments.
During the early epochs of training, the training loss of both models sharply decreases, corresponding to a sharp increase in the cumulative density of negative projections. This might correspond to the learning of ``easy'' examples, which typically happens in the early stages of training~\cite{arpit2017closer,toneva2018an}, and would indicate that our layer-wise statistic captures properties of the local hyperplane arrangements of each layer relating to learning those examples.

While networks like AlexNet, which slowly achieve training accuracy of $57\%$ without further sharp decrease in training loss, might fail to learn most hard examples, and hence preserve the local structure acquired in early epochs, VGG19, manages to reach $82\%$ training accuracy, by drastically changing the local hyperplane arrangements of each layer. Interestingly, between epochs $30$-$50$, when VGG19 is shifting from negative to positive projections, the training loss plateaus.

This observation suggests that the observed bias might arise from soft constraints imposed by learning, which can however be escaped by optimization in order to accommodate for harder patterns, if the network has enough capacity. 



More details on the learning dynamics on other datasets and their relation to the training setup can be found in the supplemental material.

%% file: sections/related_works.tex
\section{Related Works}
\label{sec:related_works}
This work concerns understanding the properties of convolutional networks with ReLU activations that produce piecewise-linear functions. The study assumes a geometrical perspective focusing on the layer-wise hyperplane arrangement induced by \textit{trained} convolutional filters. Statistics of the per-filter arrangements are collected showing a strong bias correlated with learning and generalization.  \\
Thus, different aspects of this work make it related to different subsets of the recent and yet-vast literature on understanding deep networks. Here, we try to draw such connections and at the same time delineate the specifics of this work.


\textbf{Network capacity based on affine regions.} Several works are concerned with the learning capacity of deep networks, especially as a result of sequential layers~\cite{Hastad1986}. A subset of those investigates the expressivity of feed-forward ReLU networks by analyzing the affine regions induced by hyperplane arrangement in the input space \cite{pascanu2013number,montufar2014number,raghu2017expressive,serra2018bounding,arora2018understanding,hanin2019deep,hanin2019complexity}. Among those, \cite{pascanu2013number,montufar2014number,serra2018bounding,arora2018understanding,hanin2019deep,hanin2019complexity} offer upper and lower bounds on the number of affine regions with a specific structure (number of layers and neurons). The study of hyperplane arrangement is the common theme between our work and those. However, this work is fundamentally different from them in that it is not concerned with the expressivity and instead studies the geometrical form of the function represented by a trained network and demonstrates a qualitative arrangement bias emerging as a result of training.
%






\textbf{Complexity Measures.} Another set of works try to propose measures of complexity/regularity for trained networks \cite{neyshabur2014search,neyshabur2015norm,neyshabur2017exploring,raghu2017expressive,serra2018bounding,hanin2019deep}. \cite{raghu2017expressive,serra2018bounding,hanin2019deep} define complexity measures based on the affine regions and activation patterns of a trained network, while \cite{neyshabur2014search,neyshabur2015norm,neyshabur2017exploring} use norm-based measures of the trained weights to obtain regularization measures. All of them, similar to us, look for the inductive bias of a trained and generalizing network. Different from these works we seek the inductive bias in the geometrical properties of the learned functions.


  

\textbf{Study of decision boundaries.} Finally, \cite{jiang2018predicting,novak2018sensitivity} are related to us in that they also study the properties of the learned functions and the correlation of those properties to generalization. However, they both focus on decision boundaries while this study concerns all the boundaries between affine regions. Furthermore, our work is only focused on trained weights while \cite{jiang2018predicting,novak2018sensitivity} uses data as well as trained weights.

  
  
  

%% file: sections/conclusion.tex
\section{Conclusion}
\label{sec:conclusion}

We performed an empirical exploration of the local structure of hyperplane arrangements induced by the filters of a convolutional layer in its preactivation space, for large networks trained in practice on image classification tasks of various complexity. We introduced a measure over the weights of a trained network and highlighted interesting correlations with the dynamics of learning. For large networks trained on lower-complexity datasets, we have shown that our measure correlates with layers that are critical for \textit{validation} performance, suggesting that local hyperplane configurations might encode properties relevant to generalization, that could potentially help shed more light on the puzzle of understanding the implicit bias of deep networks.

In particular, by exploiting the symmetric structure of hyperplane arrangements for convolutional layers, we provided a novel angle to study the problem of learning and generalization for convolutional networks.

Finally, we note that the general problem of studying hyperplane arrangements for deep networks is much more complex when the hierarchical dependency between layers is taken into account. We leave for future work the study of the relationship between hyperplane arrangements and image data, and the problem of characterizing non-empty cells in the hyperplane configurations of a layer.

%% file: sections/acknowledgement.tex
\subsubsection*{Acknowledgments}
The work was partially funded by Swedish Research Council project 2017-04609.

%% file: sections/appendix.tex
\section{Supplemental Material}
\label{appendix}

This document collects complementary experiments to the main paper. More details on the experimental setup and learning rate schedules used are given in section~\ref{appendix:setup}. Sections~\ref{appendix:pytorch} and~\ref{appendix:matconvnet}, respectively present histograms of the cosine angles (projection statistics) for off-the-shelf pretrained models released by the Pytorch and MatConvNet communities. Section~\ref{appendix:convergence} presents the cumulative distribution of negative projection statistics for a network that fails to converge. Section~\ref{appendix:reinit} extends our main experiments to VGG19 trained on CIFAR100, while section~\ref{appendix:pixel_shuffle} studies the cumulative distribution of negative projections for noisy data. Finally, section~\ref{appendix:first_layer} presents additional experiments on the distribution of the first layer for various datasets and different pixel representations, and section~\ref{appendix:schedules} discusses how our experiments, and other related work, are partly affected by the training setup.

\subsection{Experimental Setup}
\label{appendix:setup}

This section describes in more detail the training schedules and data augmentation strategies used in all experiments. With the exception of figure~\ref{fig:noise}, where training data with random labels is fitted, all experiments in the main paper are performed with step learning rate schedule:
\begin{itemize}
  \item Starting from base learning rate $0.01$, the learning rate is decreased by a factor of $10$ every $25$ epochs. Each epoch in which the learning rate is decreased is denoted by red ticks on the epochs axis in each relevant figure. 
  \item All networks are trained with data augmentation consisting of vertical and horizontal shifts of $4$ pixels in each direction and random horizontal flipping.
\end{itemize}

\noindent In section~\ref{appendix:schedules}, we repeat most of the experiments using a smooth (exponential) learning rate schedule. 
\begin{itemize}
  \item Starting from base learning rate $0.01$, at every epoch the learning rate is decayed by a factor of $10^{-\frac{1}{25}}$. 
  \item The smooth  learning rate schedule is employed to interpolate between the learning rates obtained with the step schedule and is meant to remove artifacts arising from drastic changes in the learning rate. 
\end{itemize}
  
\noindent Finally, for the pixel shuffle experiment (figure~\ref{fig:pixel_shuffle}), a base learning rate of $0.01$ is decayed by $0.95$ at each epoch, following the training setup of~\cite{zhang2018understanding}.

\subsection{Pytorch Pretrained Models}
\label{appendix:pytorch}

In addition to figure~\ref{fig:pretrained}, this section shows histograms of cosine angles for various off-the-shelf models released by the Pytorch community. For all networks considered, most histograms show how the cosine angles shift towards negative values at the end of training, as opposed to random initialization. Since snapshots at initialization were not available for pretrained models, we plot for reference the statistics for one random initialization.

We believe the histograms for pretrained models are interesting, because they show how the statistical bias that we observe is also present for training setups and implementations independent from ours.

Importantly, for deeper layers, we consistently observe a strong bias towards either all-negative or all-positive projections. We stress that such bias is statistically significant since initialization is zero-centered with small variance, and for a convolutional layer learning $F$ output maps, the probability of all polyhedra $\{\mathcal{H}^l_i\}_{i=1}^F$ to be aligned along $\mathcal{I}^+$, is $\frac{F}{2^F}$, which is fairly small, highlighting constraints consistently arising from training.

  \begin{figure}[t!]
    \centering
    \subfloat{\includegraphics[width=0.18\textwidth]{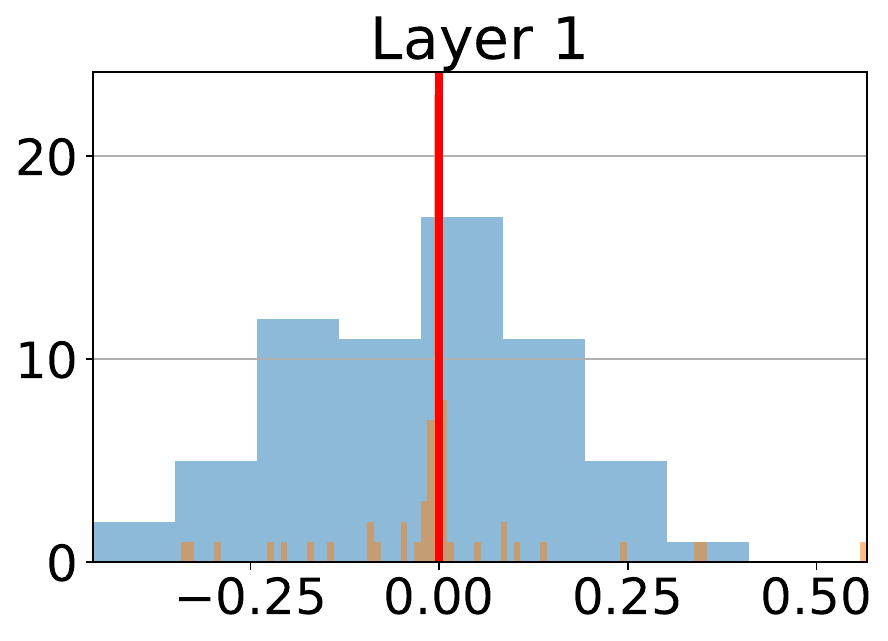}} ~
    \subfloat{\includegraphics[width=0.18\textwidth]{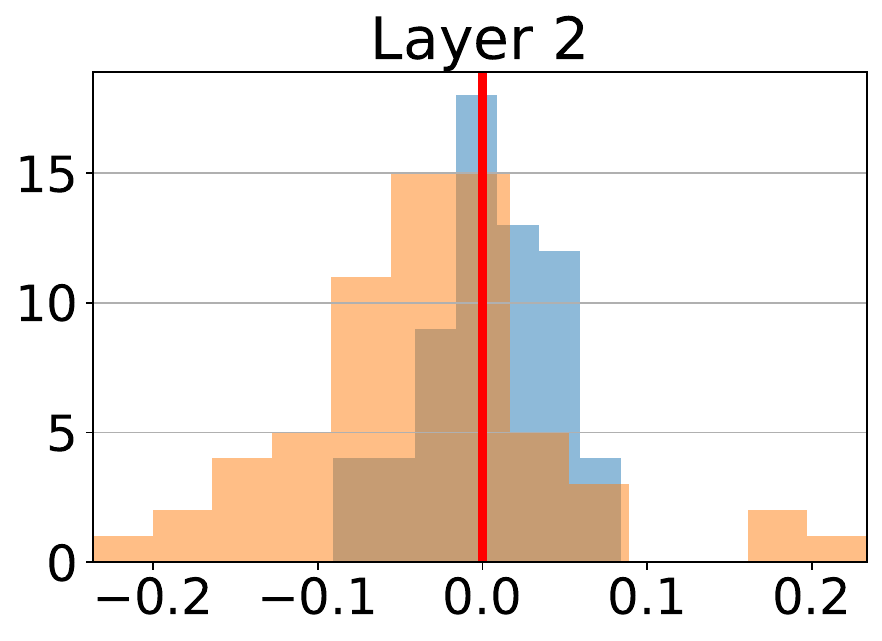}} ~
    \subfloat{\includegraphics[width=0.18\textwidth]{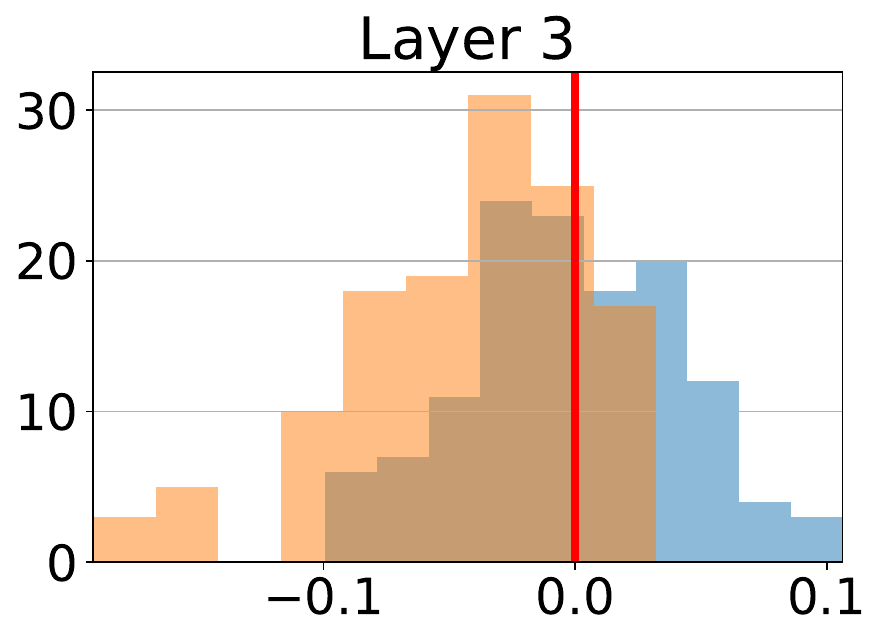}} ~
    \subfloat{\includegraphics[width=0.18\textwidth]{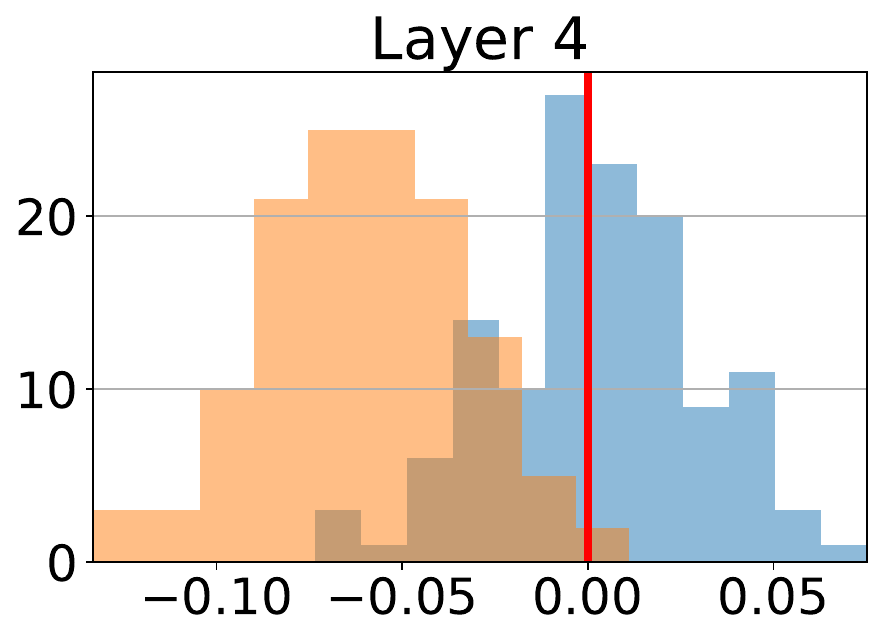}} ~
    \subfloat{\includegraphics[width=0.18\textwidth]{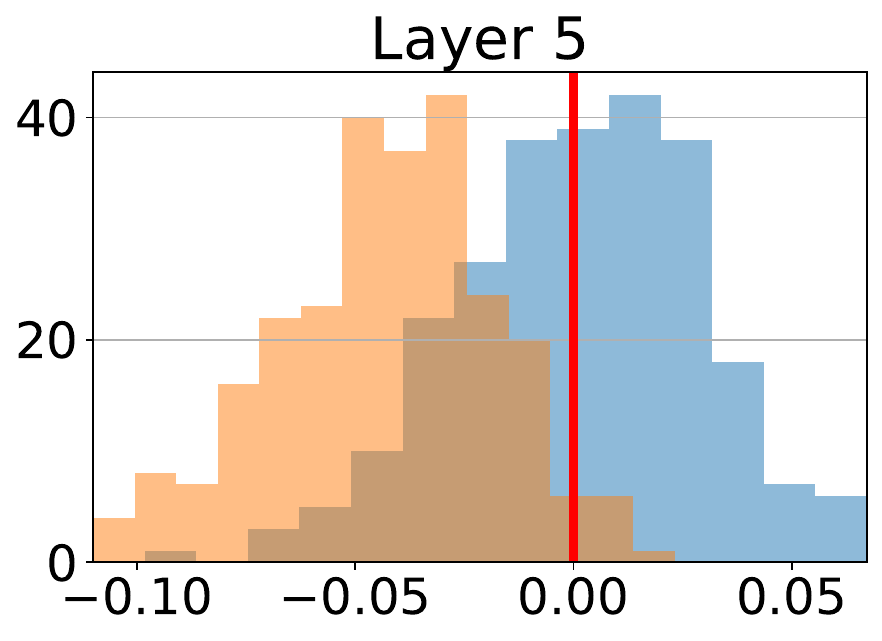}} \\
    \subfloat{\includegraphics[width=0.18\textwidth]{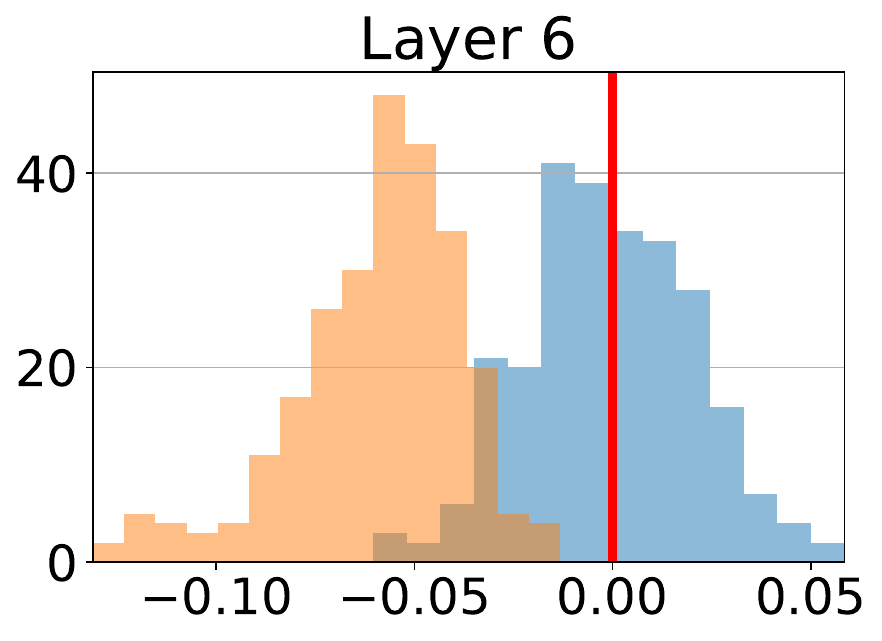}} ~
    \subfloat{\includegraphics[width=0.18\textwidth]{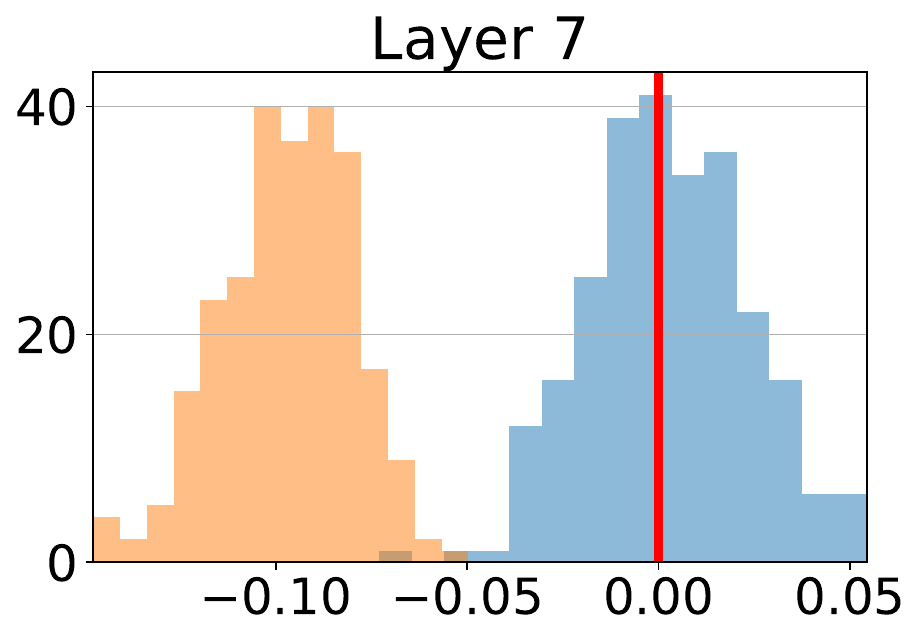}} ~
    \subfloat{\includegraphics[width=0.18\textwidth]{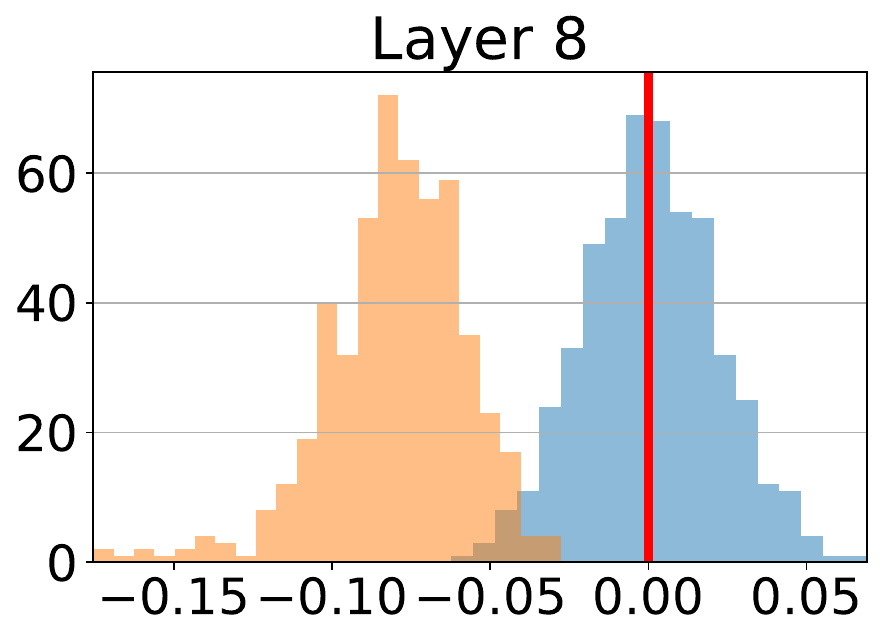}} ~
    \subfloat{\includegraphics[width=0.18\textwidth]{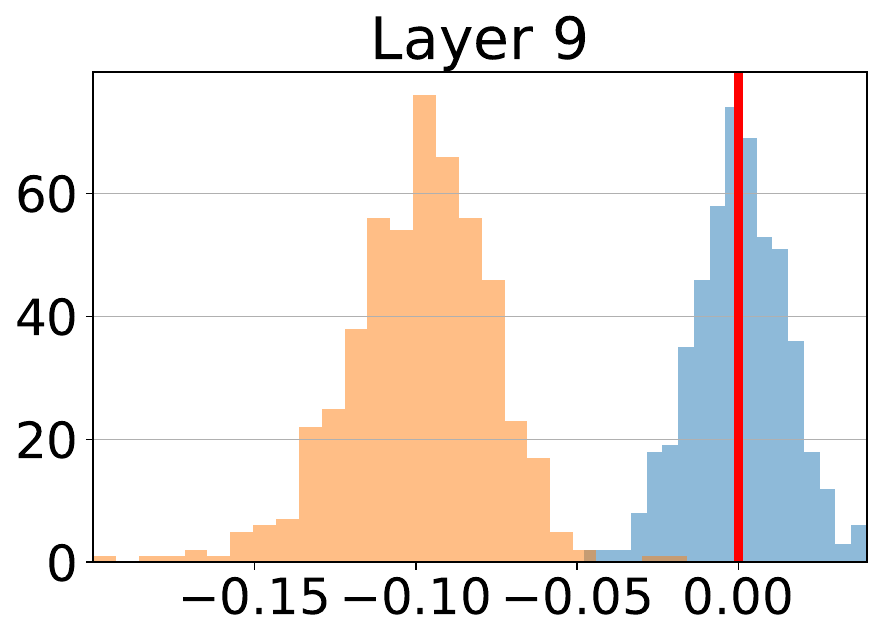}} \\
    \subfloat{\includegraphics[width=0.18\textwidth]{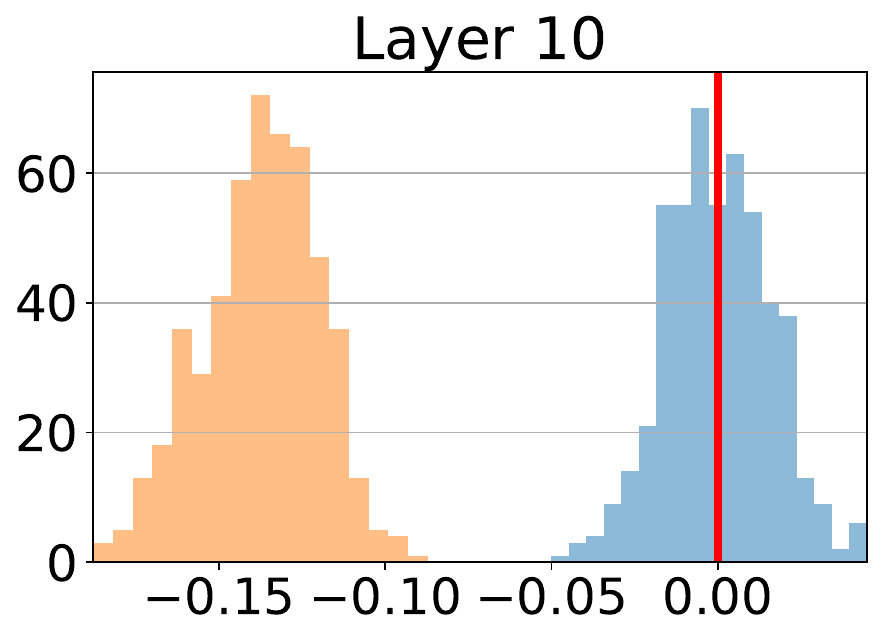}} ~
    \subfloat{\includegraphics[width=0.18\textwidth]{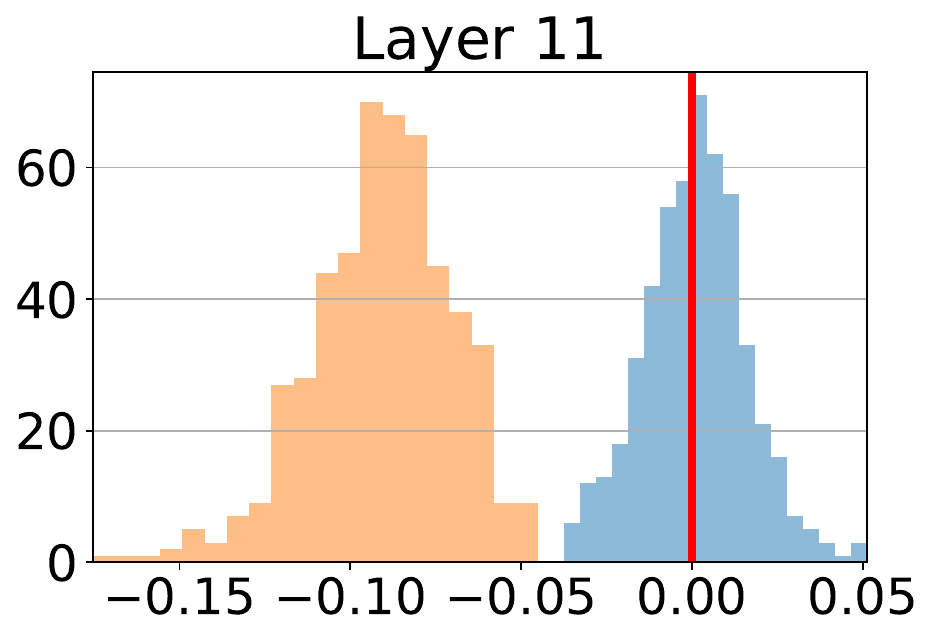}} ~
    \subfloat{\includegraphics[width=0.18\textwidth]{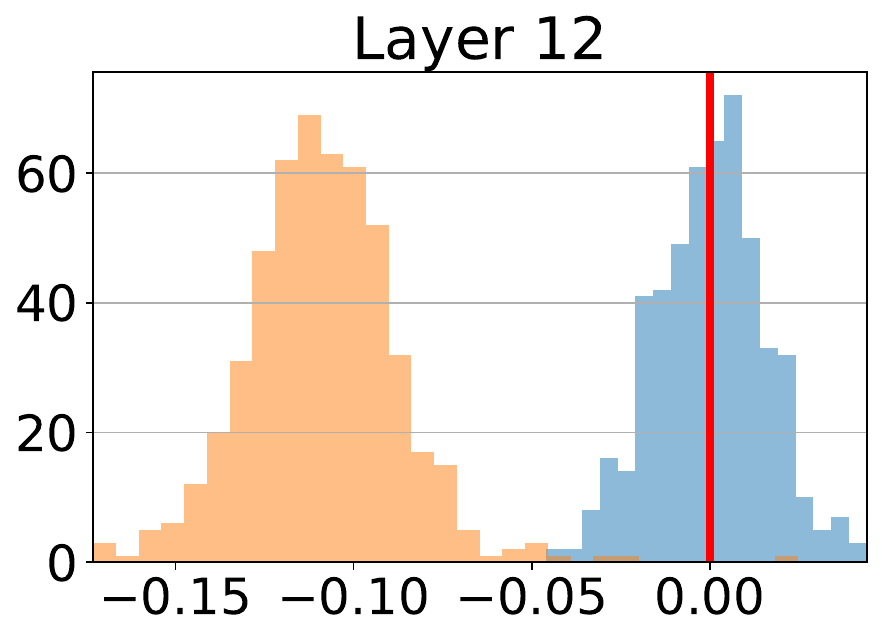}} ~
    \subfloat{\includegraphics[width=0.18\textwidth]{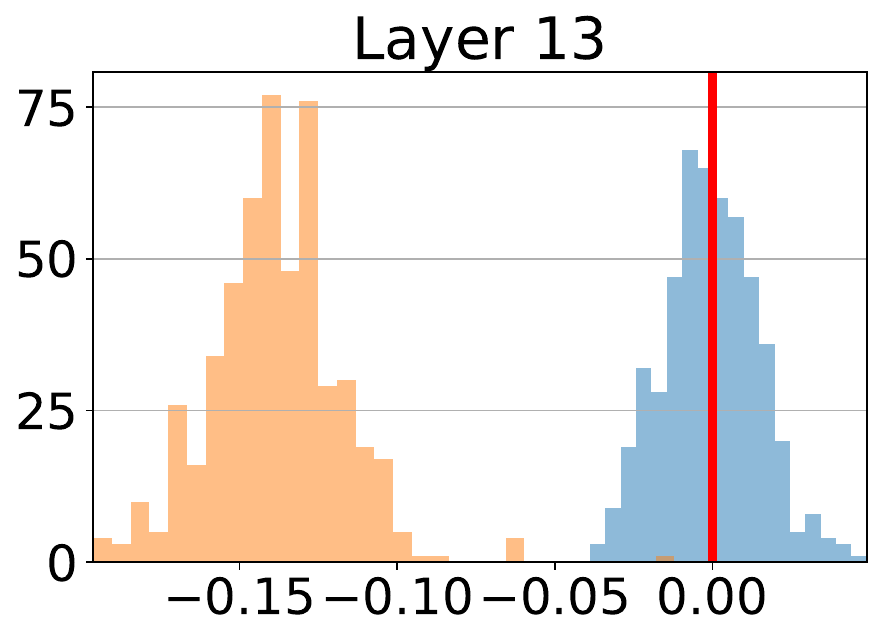}} ~
    
    \caption{Distribution of our projection statistic for a random initialization (blue) and for a \textit{pretrained} \textbf{VGG16 on ImageNet} by the \texttt{PyTorch} community (orange).}
    \label{fig:pytorch_vgg16}
  \end{figure}
  
  \begin{figure}[h!]
    \centering
    \subfloat{\includegraphics[width=0.18\textwidth]{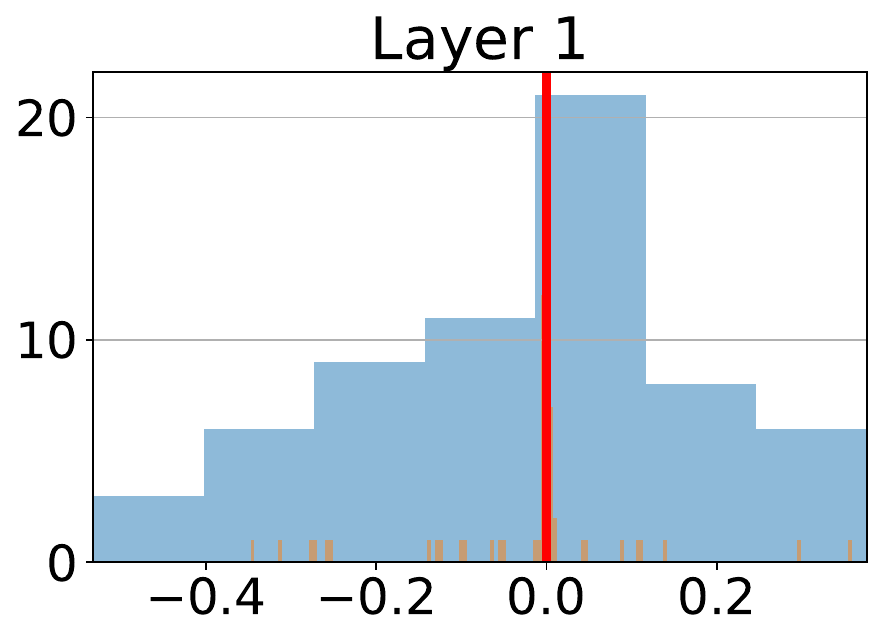}} ~
    \subfloat{\includegraphics[width=0.18\textwidth]{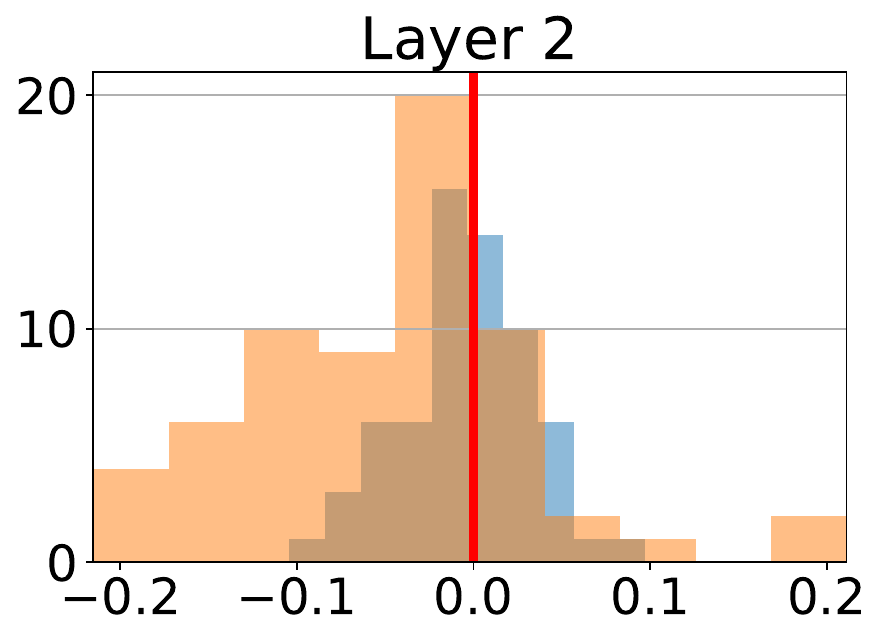}} ~
    \subfloat{\includegraphics[width=0.18\textwidth]{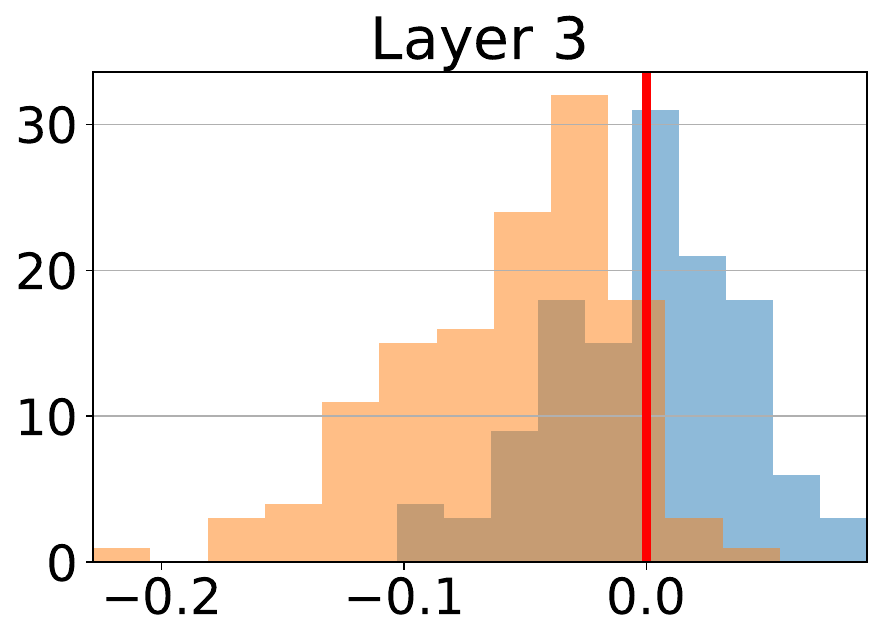}} ~
    \subfloat{\includegraphics[width=0.18\textwidth]{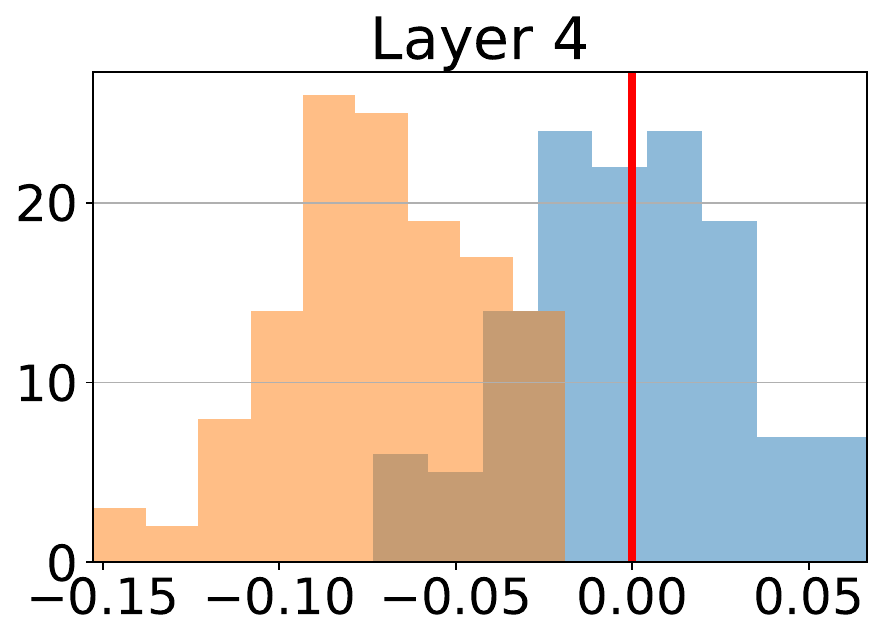}} ~
    \subfloat{\includegraphics[width=0.18\textwidth]{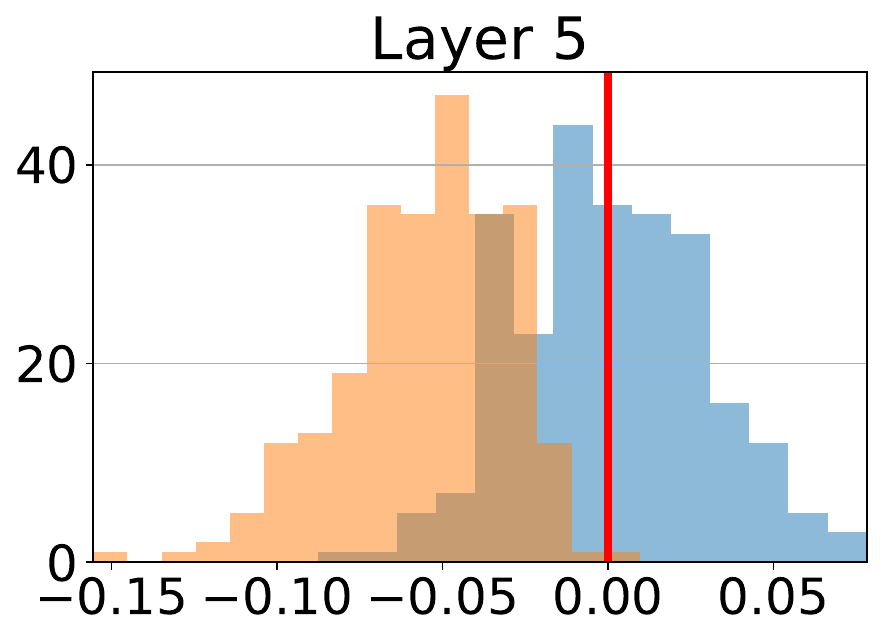}} \\
    \subfloat{\includegraphics[width=0.18\textwidth]{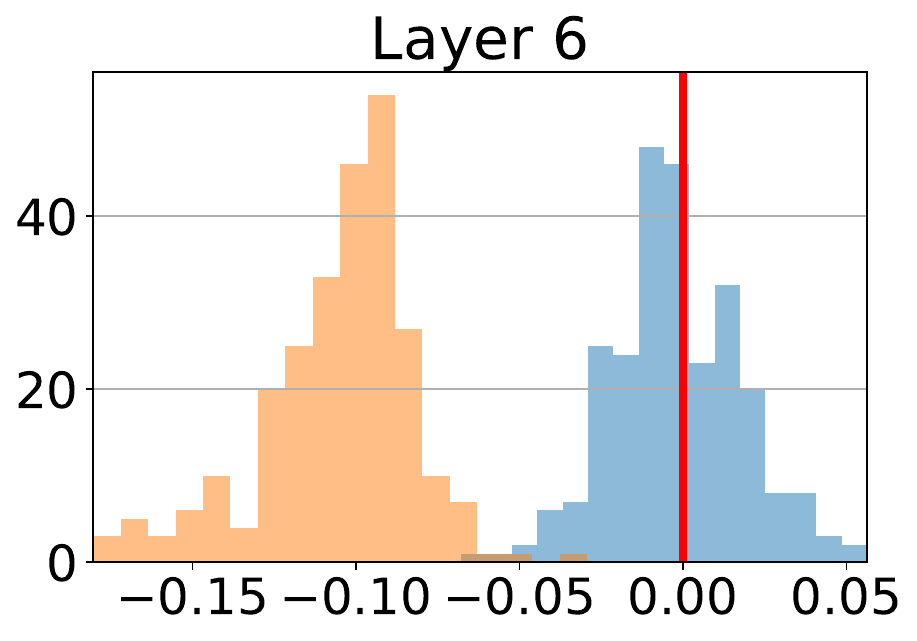}} ~
    \subfloat{\includegraphics[width=0.18\textwidth]{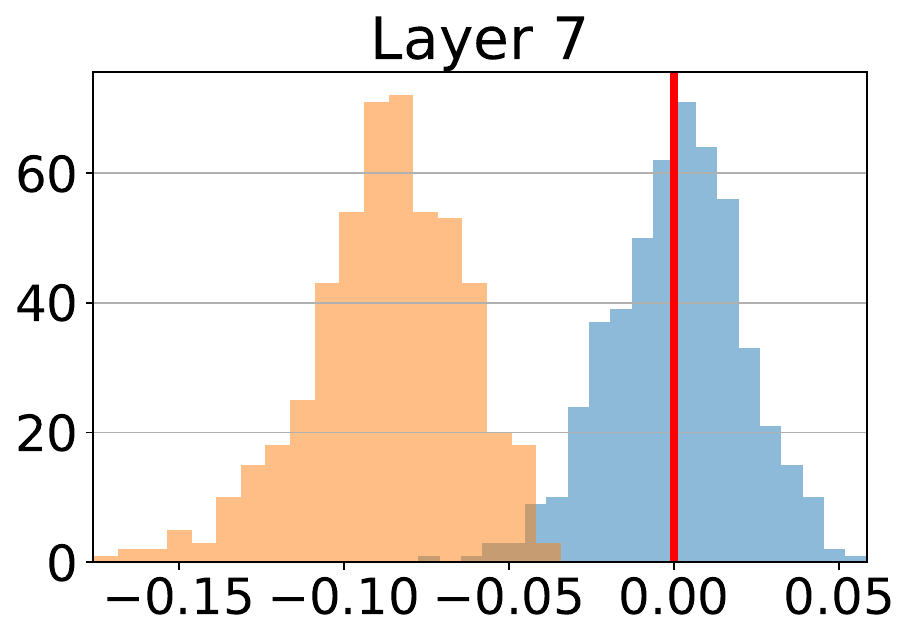}} ~
    \subfloat{\includegraphics[width=0.18\textwidth]{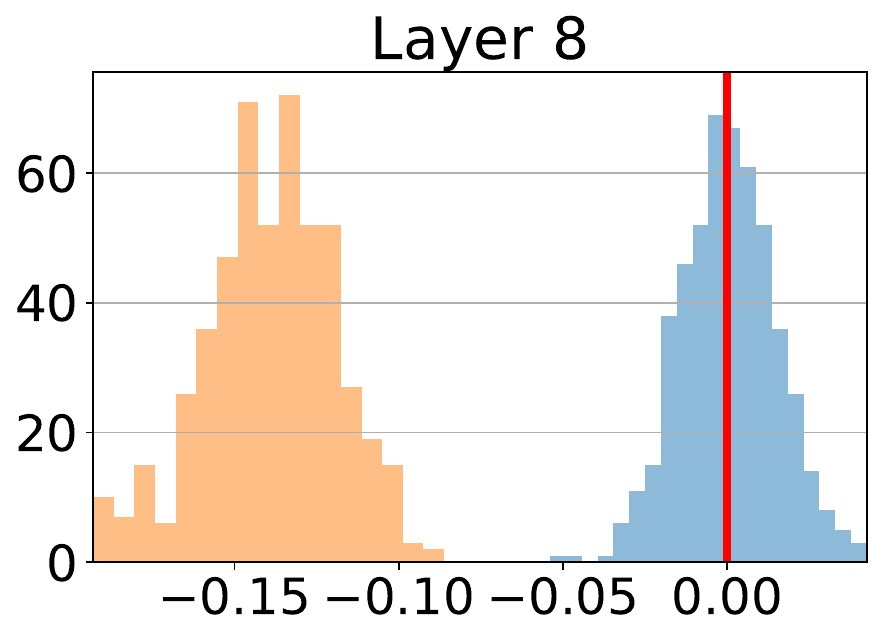}} ~
    \subfloat{\includegraphics[width=0.18\textwidth]{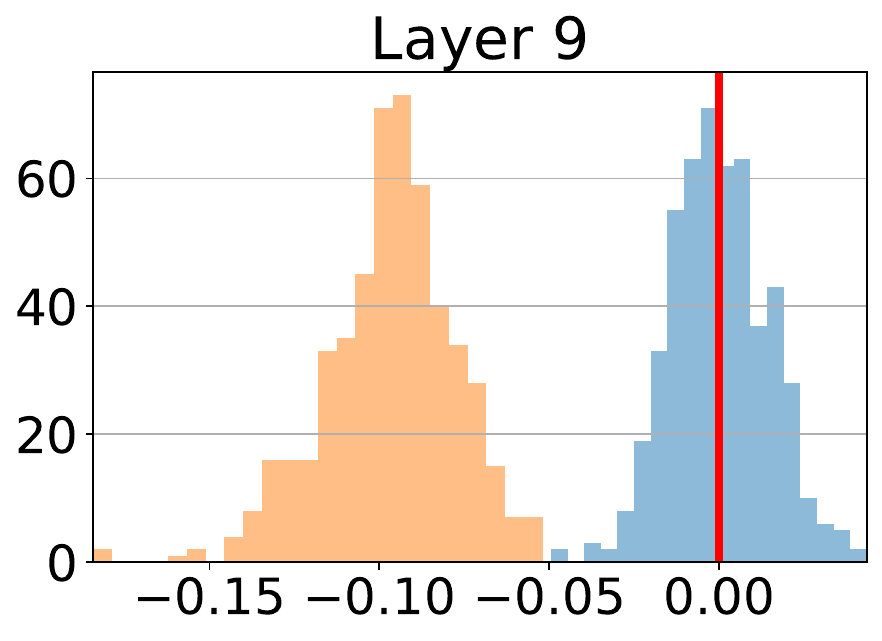}} ~
    \subfloat{\includegraphics[width=0.18\textwidth]{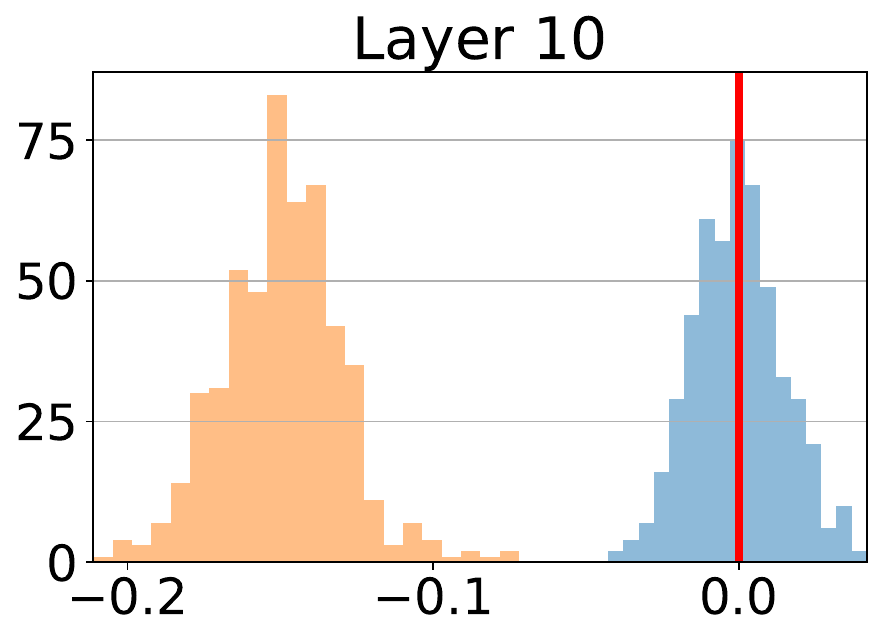}} \\
    
    \caption{Distribution of our projection statistic for a random initialization (blue) and for a \textit{pretrained} \textbf{VGG13 on ImageNet} by the \texttt{PyTorch} community  (orange).}
    \label{fig:pytorch_vgg13}
  \end{figure}

  \begin{figure}[h!]
    \centering
    \subfloat{\includegraphics[width=0.18\textwidth]{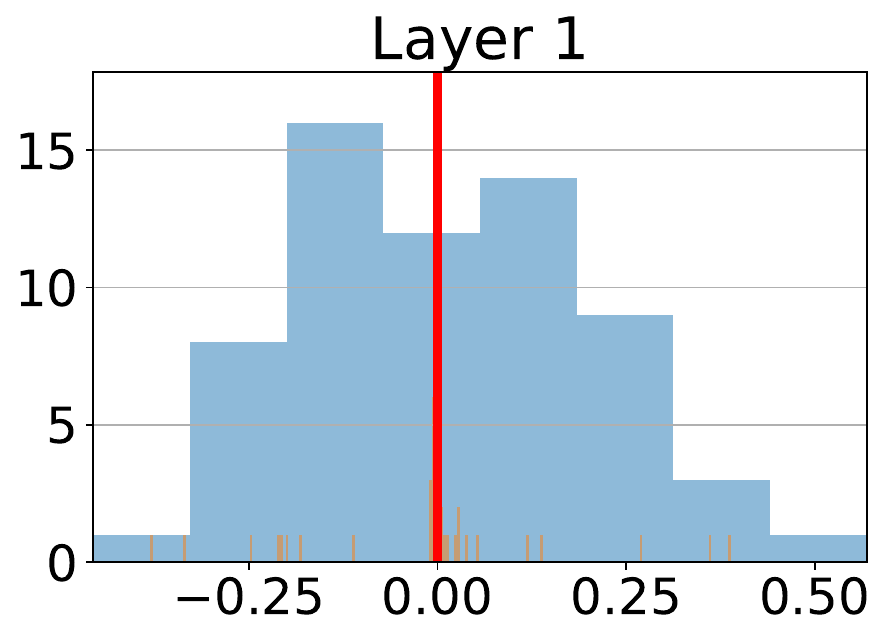}} ~
    \subfloat{\includegraphics[width=0.18\textwidth]{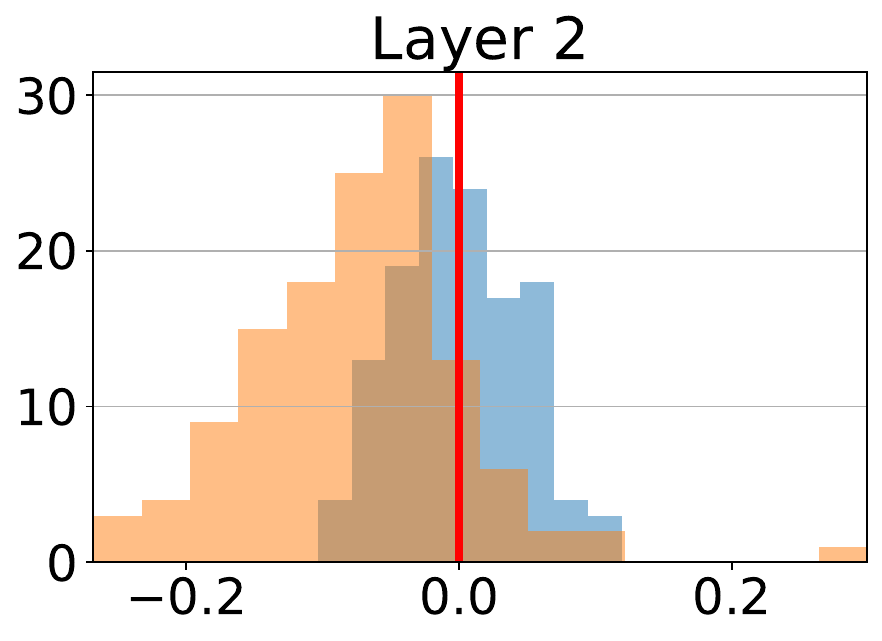}} ~
    \subfloat{\includegraphics[width=0.18\textwidth]{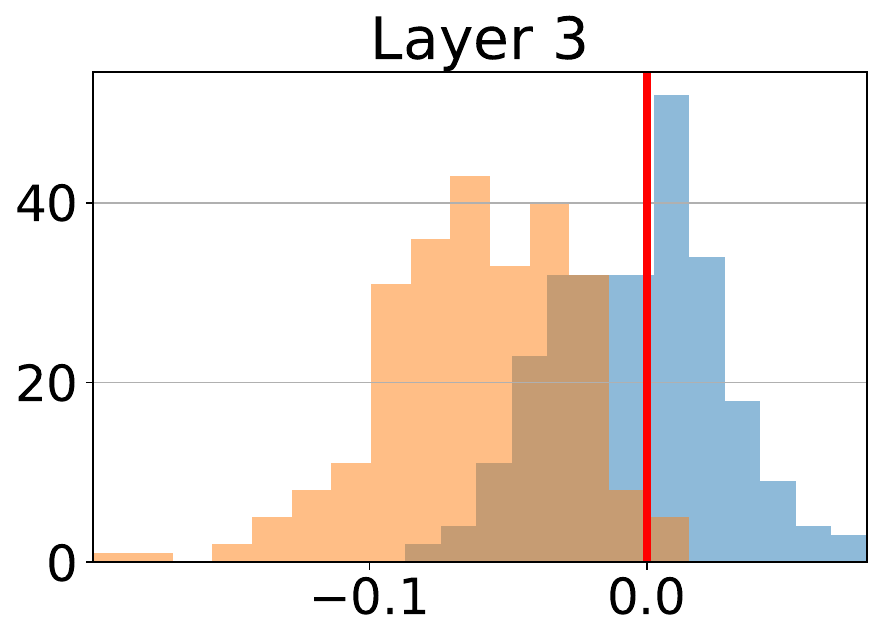}} ~
    \subfloat{\includegraphics[width=0.18\textwidth]{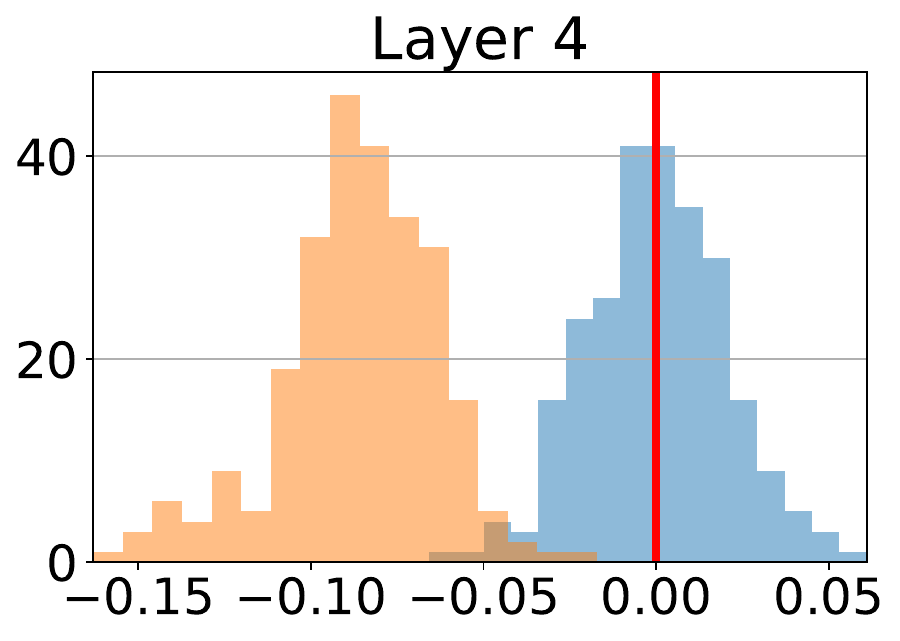}} \\
    \subfloat{\includegraphics[width=0.18\textwidth]{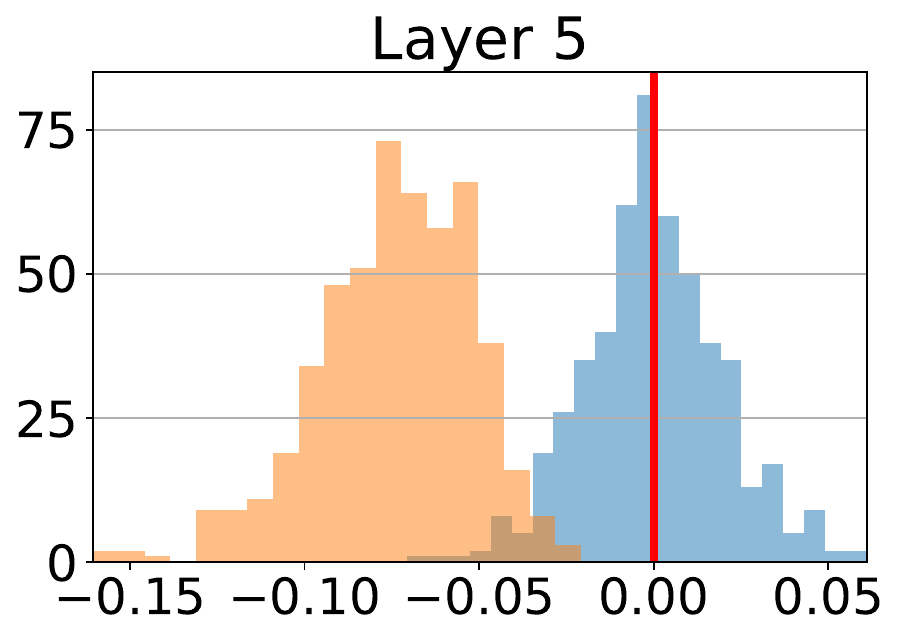}} ~
    \subfloat{\includegraphics[width=0.18\textwidth]{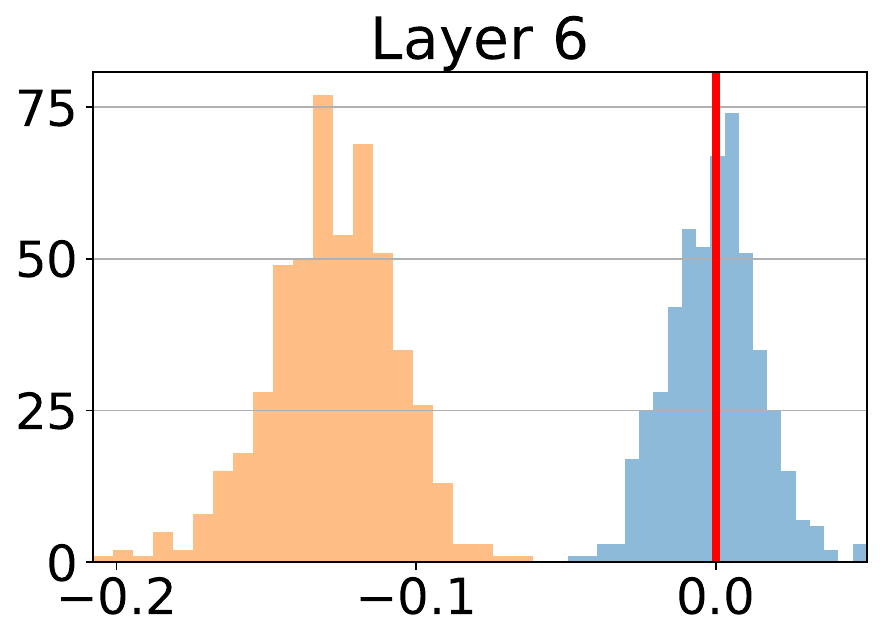}} ~
    \subfloat{\includegraphics[width=0.18\textwidth]{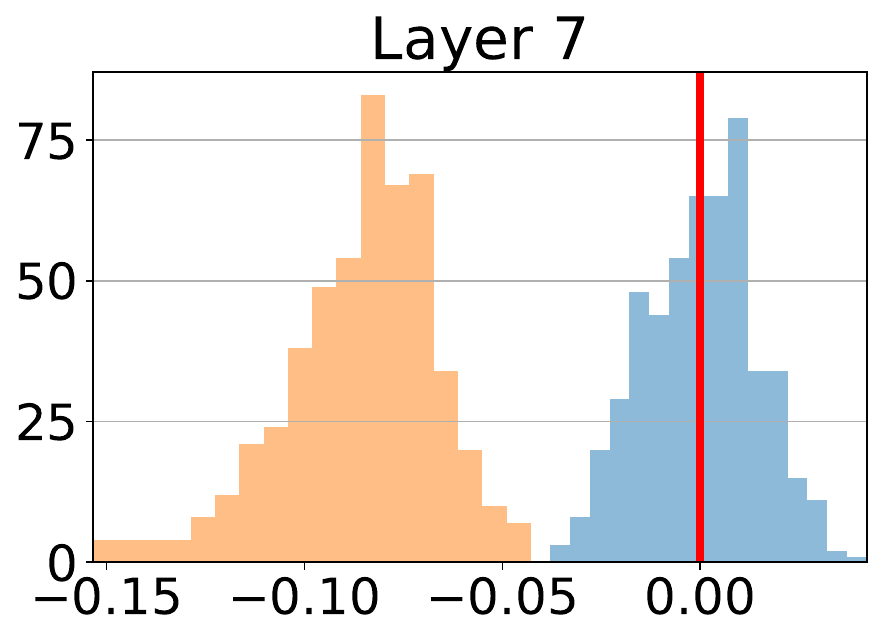}} ~
    \subfloat{\includegraphics[width=0.18\textwidth]{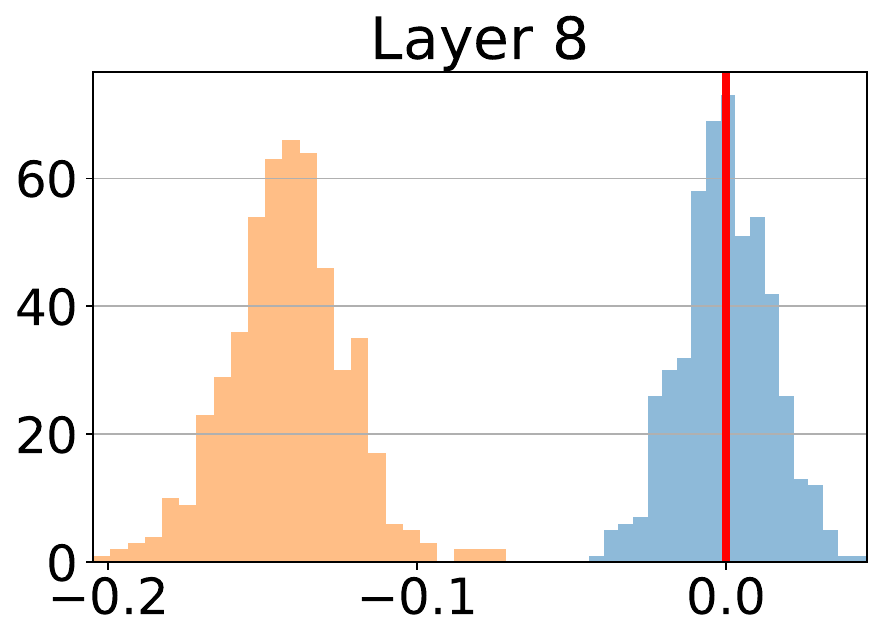}} \\
    
    \caption{Distribution of our projection statistic for a random initialization (blue) and for a \textit{pretrained} \textbf{VGG11 on ImageNet} by the \texttt{PyTorch} community  (orange).}
    \label{fig:pytorch_vgg11}
  \end{figure}
  
  \begin{figure}[h!]
    \centering
    \subfloat{\includegraphics[width=0.18\textwidth]{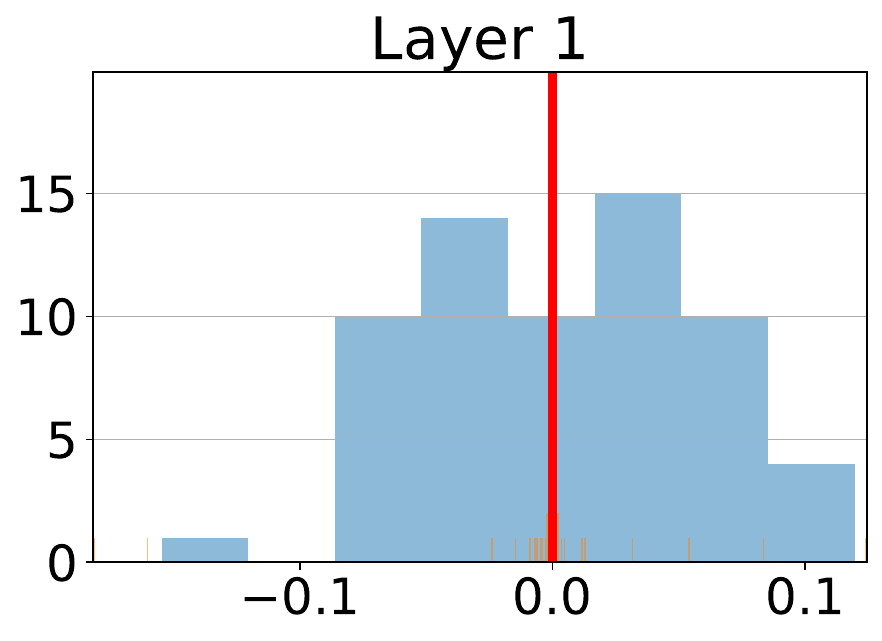}} ~
    \subfloat{\includegraphics[width=0.18\textwidth]{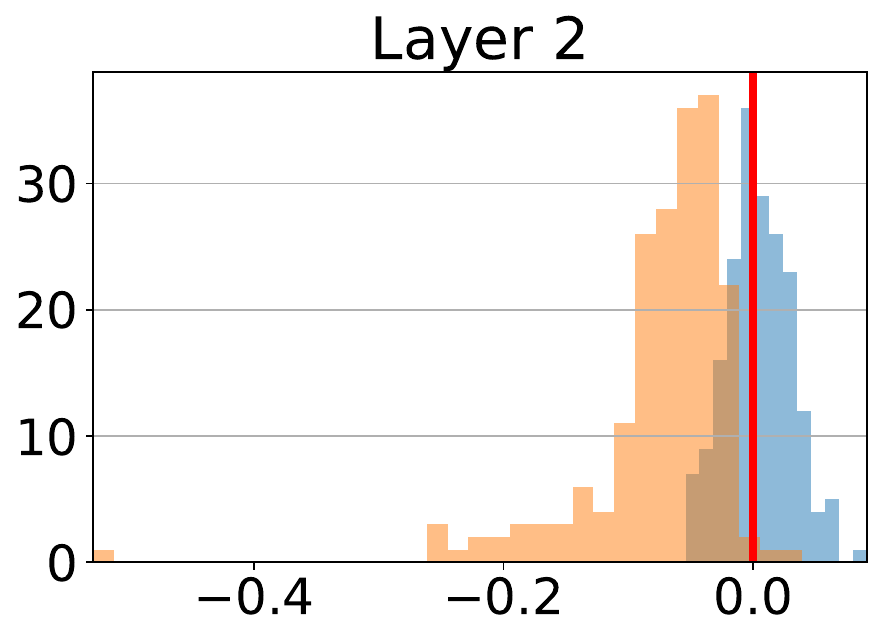}} ~
    \subfloat{\includegraphics[width=0.18\textwidth]{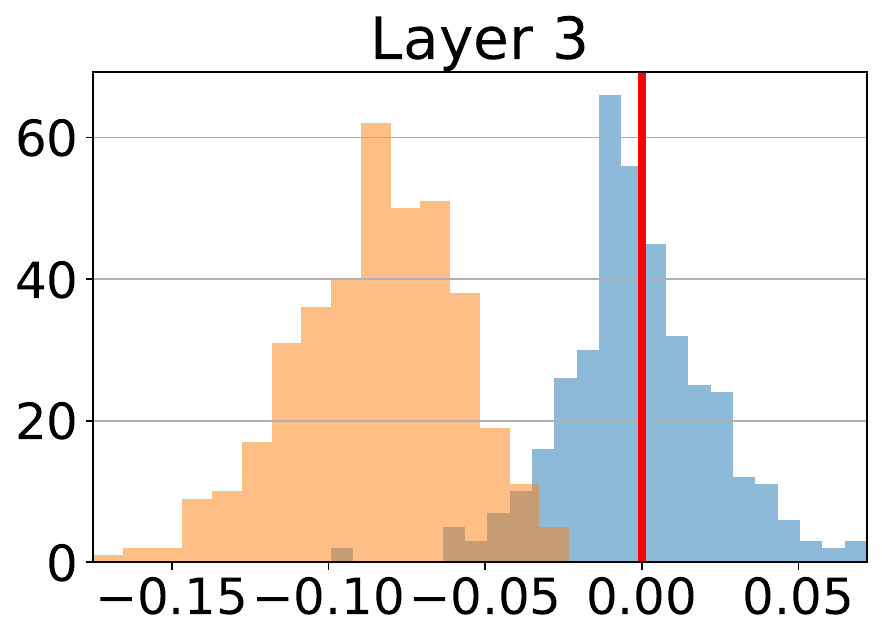}} ~
    \subfloat{\includegraphics[width=0.18\textwidth]{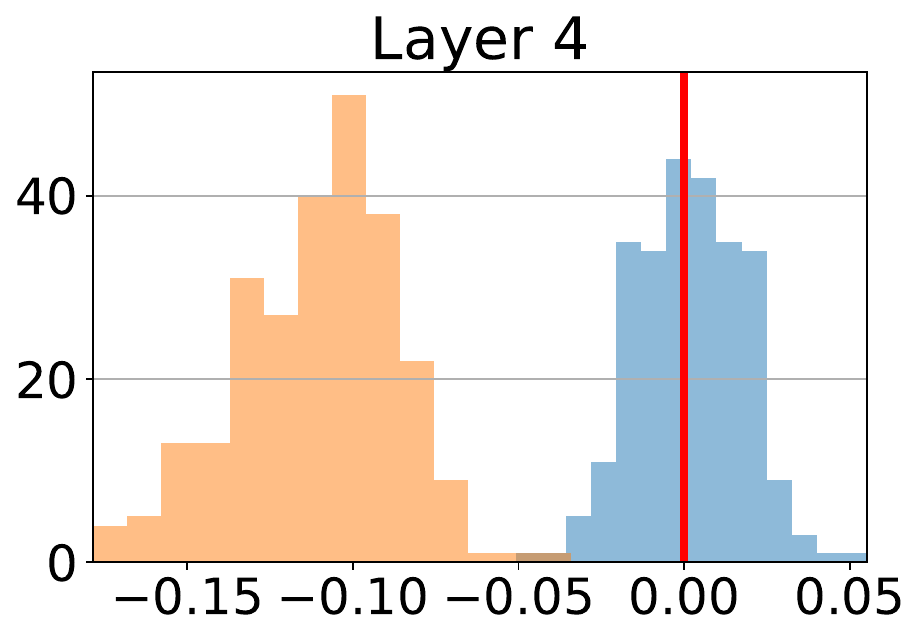}} ~
    \subfloat{\includegraphics[width=0.18\textwidth]{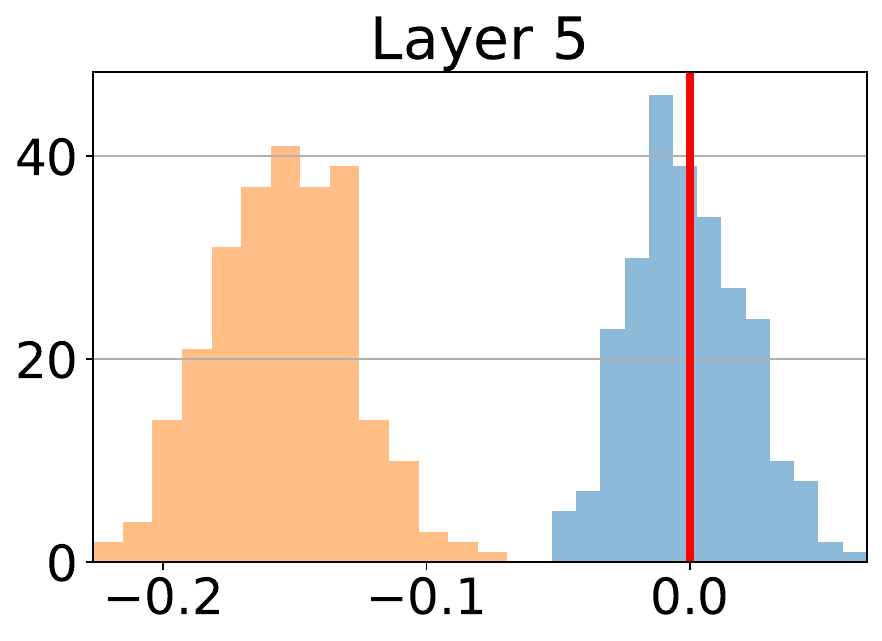}} \\
    
    \caption{Distribution of our projection statistic for a random initialization (blue) and for a \textit{pretrained} \textbf{AlexNet on ImageNet} by the \texttt{PyTorch} community  (orange).}
    \label{fig:pytorch_alexnet}
  \end{figure}

\subsection{MatConvNet Pretrained Models}
\label{appendix:matconvnet}

  \begin{figure}[b!]
    \centering
    \subfloat{\includegraphics[width=0.18\textwidth]{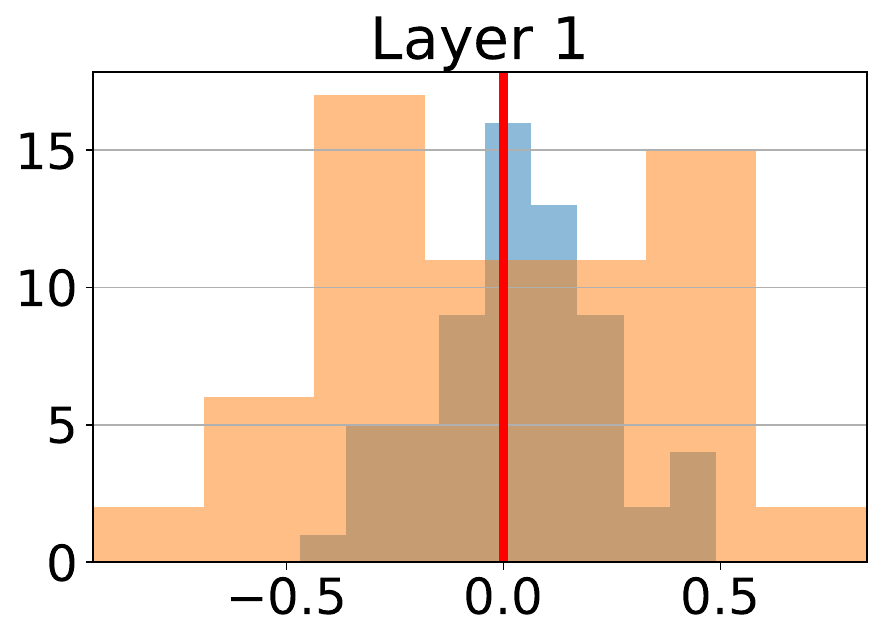}} ~
    \subfloat{\includegraphics[width=0.18\textwidth]{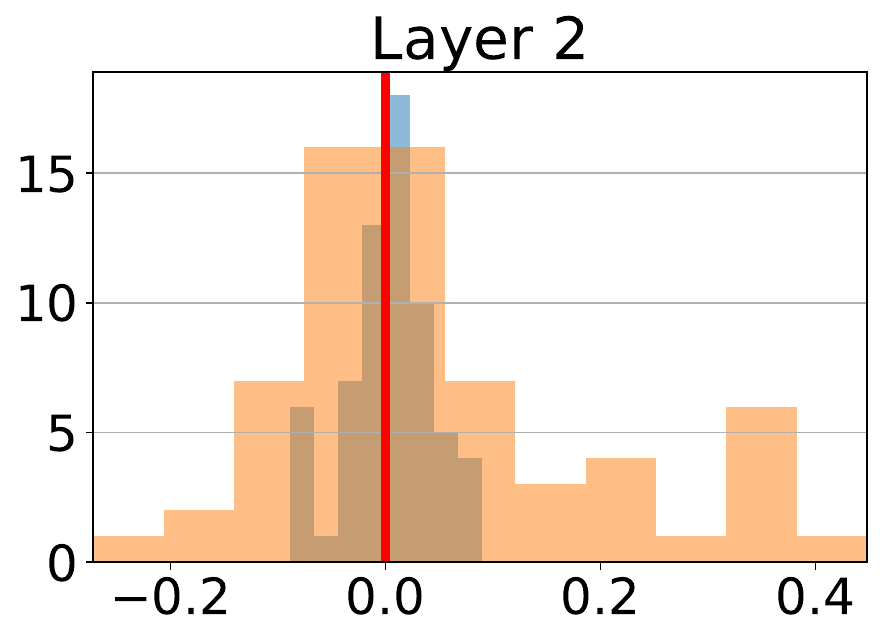}} ~
    \subfloat{\includegraphics[width=0.18\textwidth]{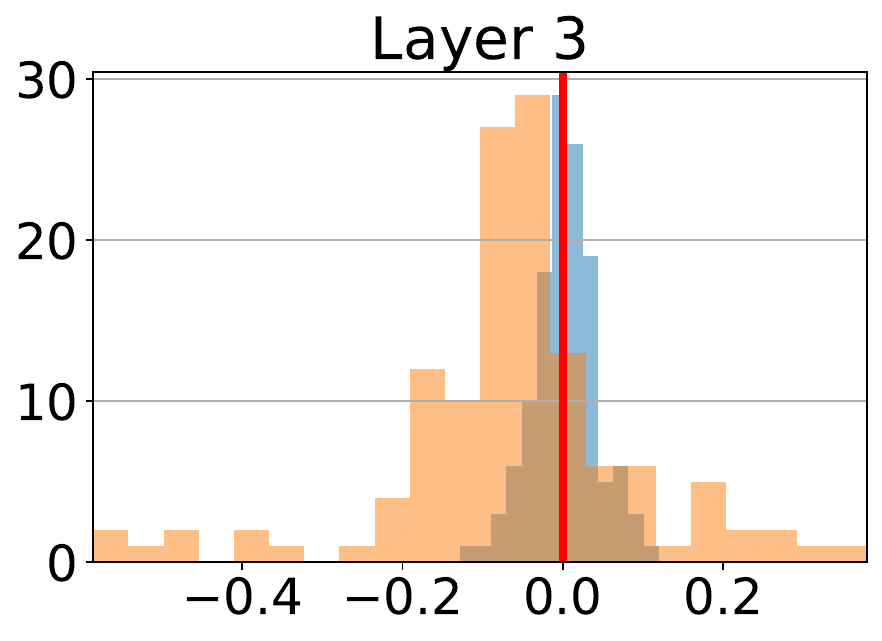}} ~
    \subfloat{\includegraphics[width=0.18\textwidth]{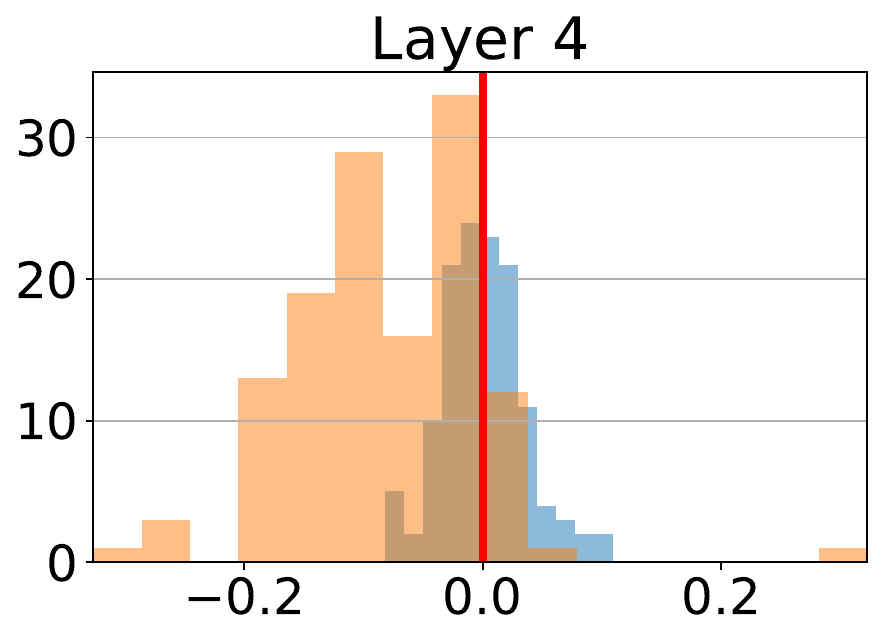}} ~
    \subfloat{\includegraphics[width=0.18\textwidth]{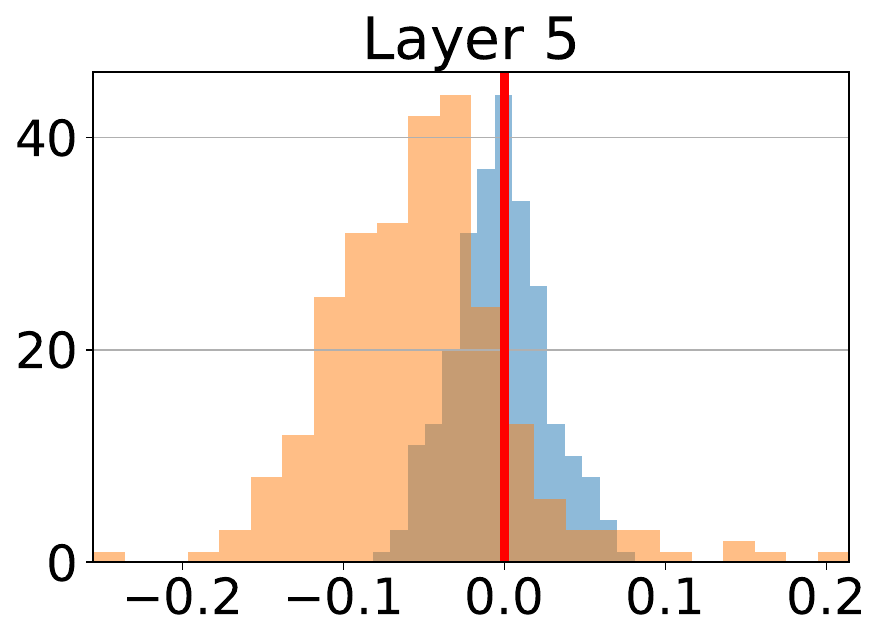}} \\
    \subfloat{\includegraphics[width=0.18\textwidth]{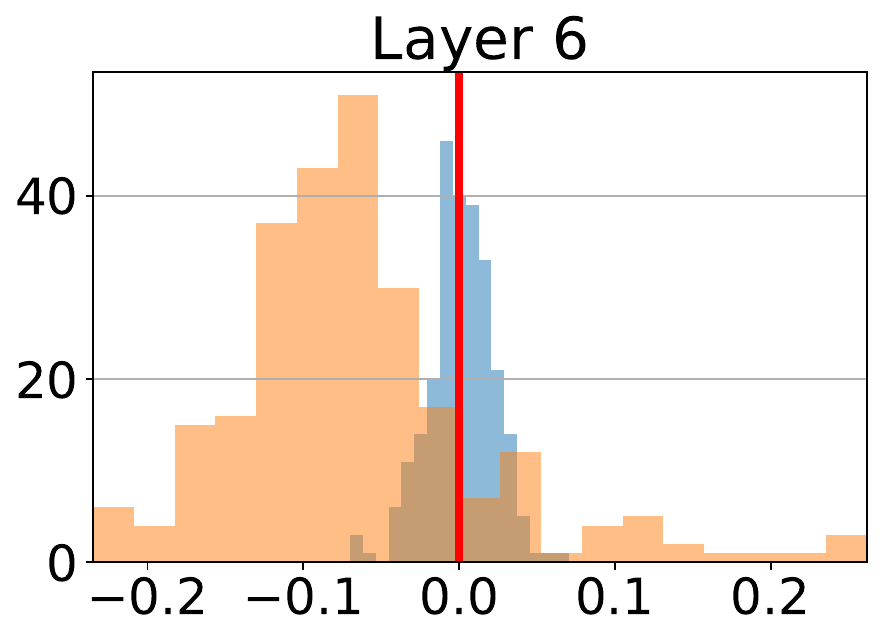}} ~
    \subfloat{\includegraphics[width=0.18\textwidth]{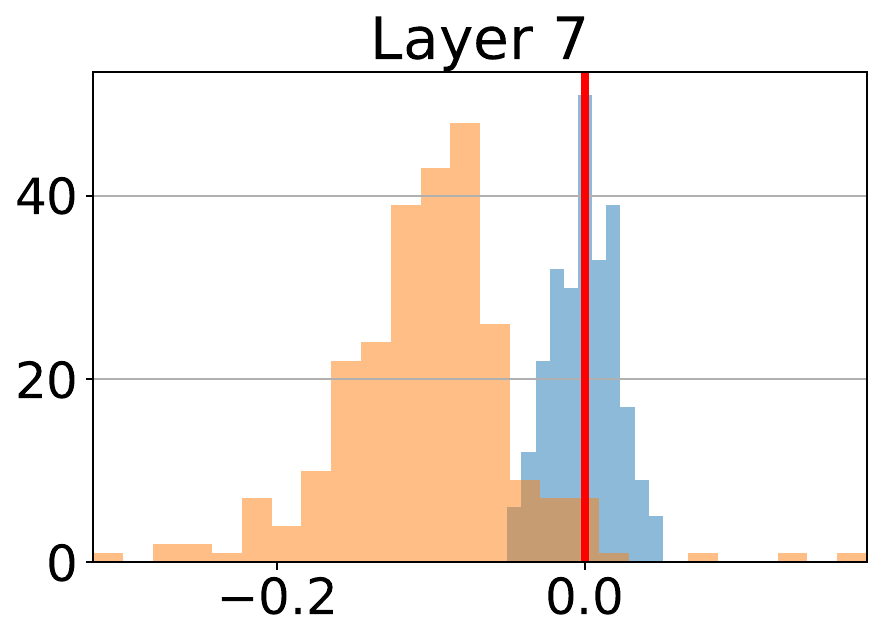}} ~
    \subfloat{\includegraphics[width=0.18\textwidth]{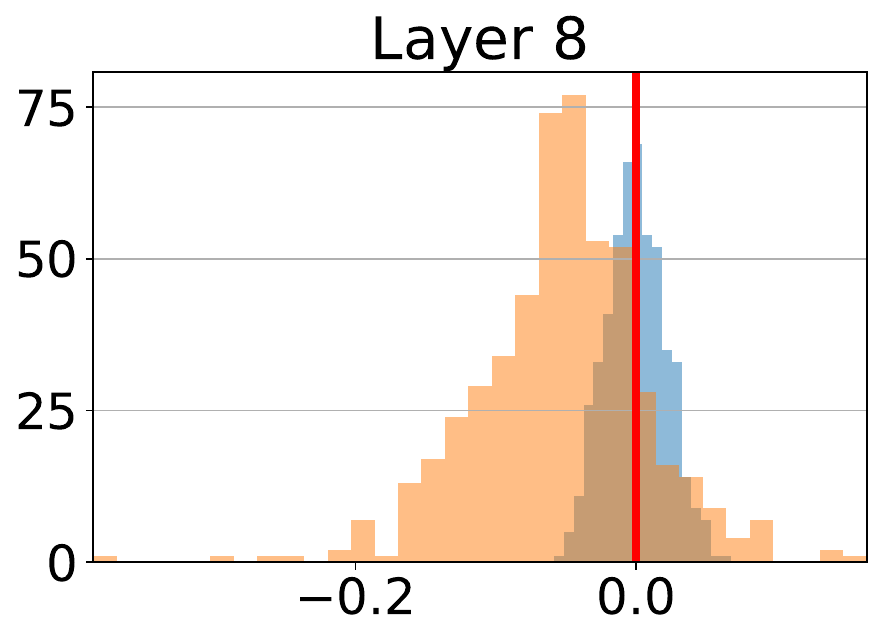}} ~
    \subfloat{\includegraphics[width=0.18\textwidth]{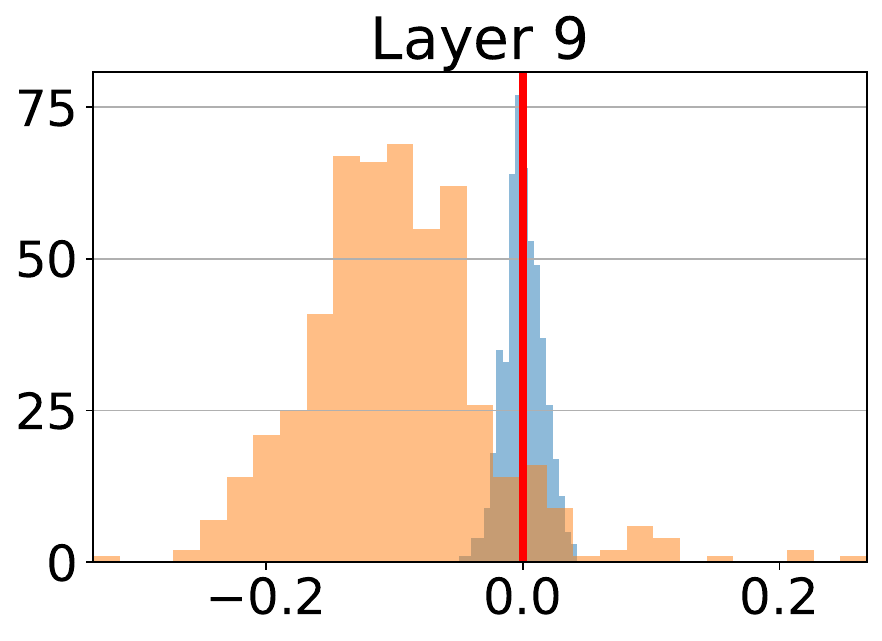}} \\
    \subfloat{\includegraphics[width=0.18\textwidth]{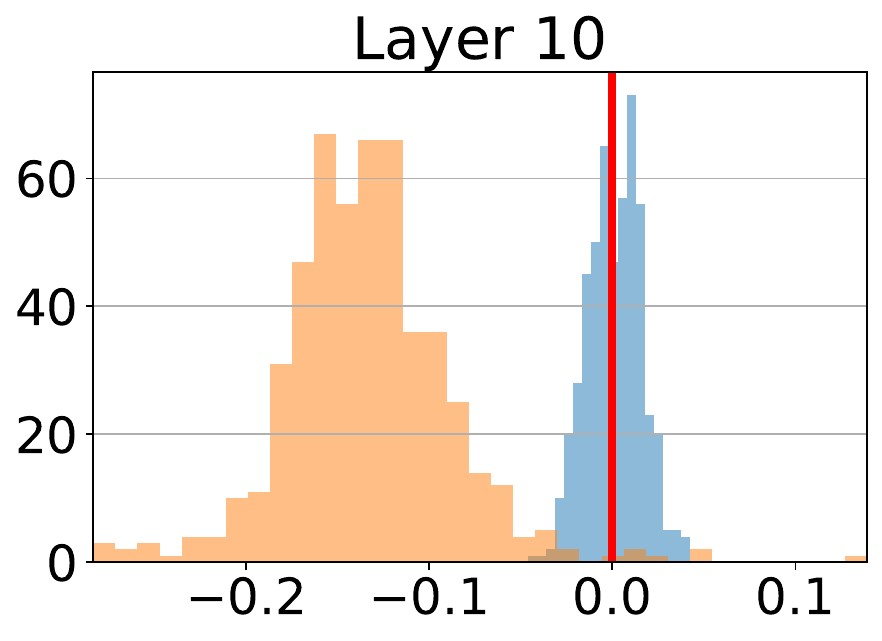}} ~
    \subfloat{\includegraphics[width=0.18\textwidth]{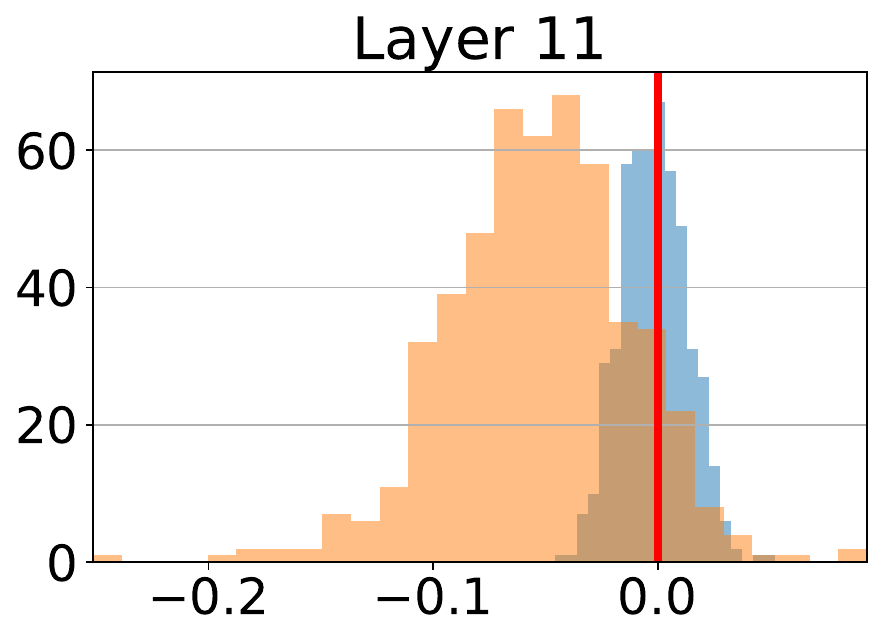}} ~
    \subfloat{\includegraphics[width=0.18\textwidth]{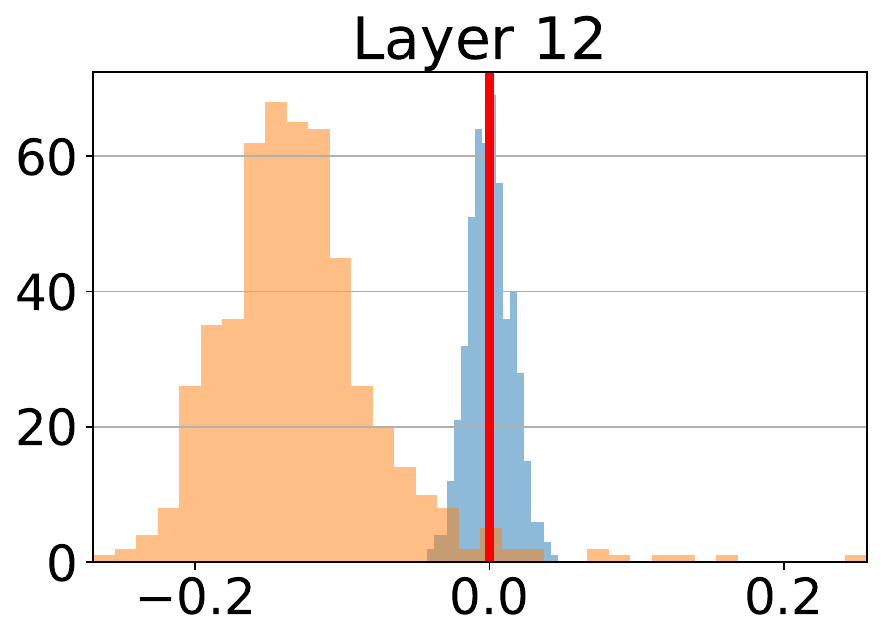}} ~
    \subfloat{\includegraphics[width=0.18\textwidth]{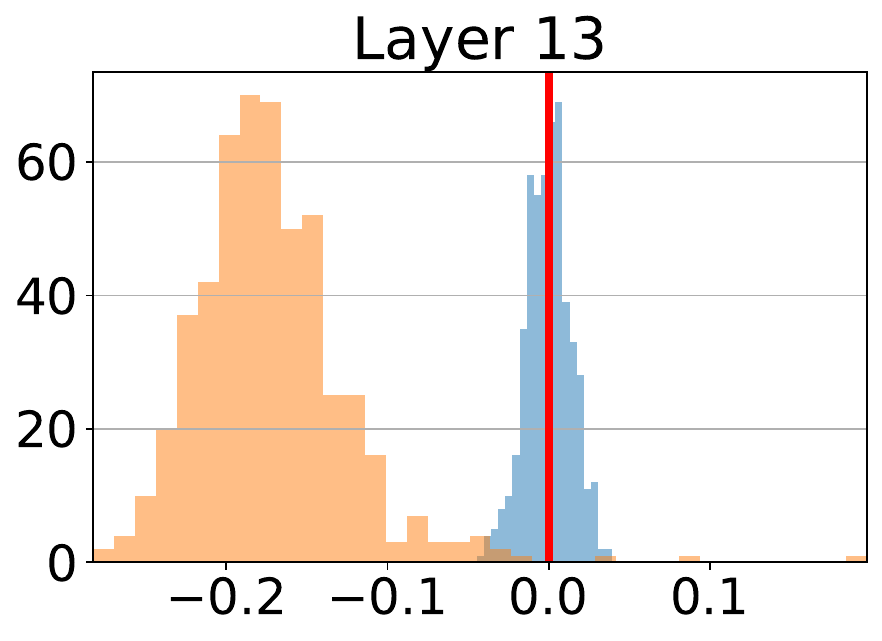}} ~
    
    \caption{Distribution of our projection statistic for a random initialization (blue) and for a \textit{pretrained} \textbf{VGG16 on ImageNet} by the \texttt{MatConvNet} community  (orange).}
    \label{fig:matconvnet_vgg16}
  \end{figure}

Similarly to section~\ref{appendix:pytorch}, we include histograms of cosine angles for off-the-shelf models released in the MatConvNet model zoo. Each model was converted to Pytorch for compatibility with our code base, using ~\texttt{pytorch-mcn}. We only used converted models whose performance was numerically verified to match that of the MatConvNet original model\footnote{\url{https://github.com/albanie/pytorch-mcn}}.

  \begin{figure}[t!]
    \centering
    \subfloat{\includegraphics[width=0.18\textwidth]{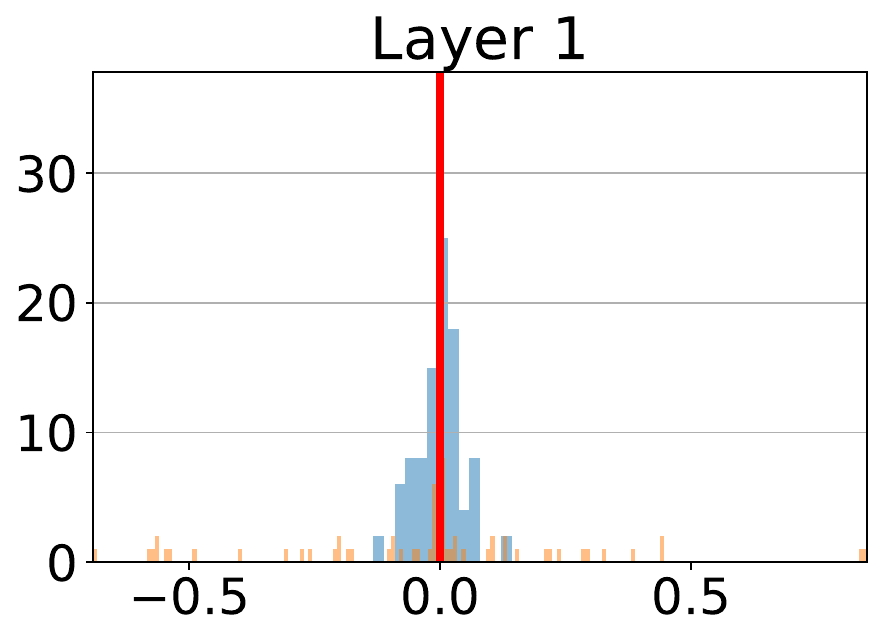}} ~
    \subfloat{\includegraphics[width=0.18\textwidth]{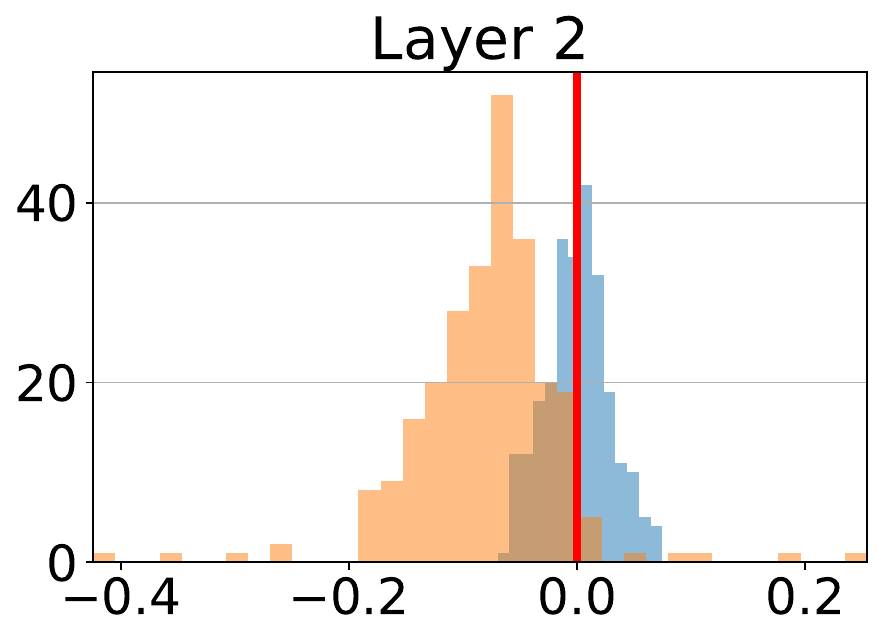}} ~
    \subfloat{\includegraphics[width=0.18\textwidth]{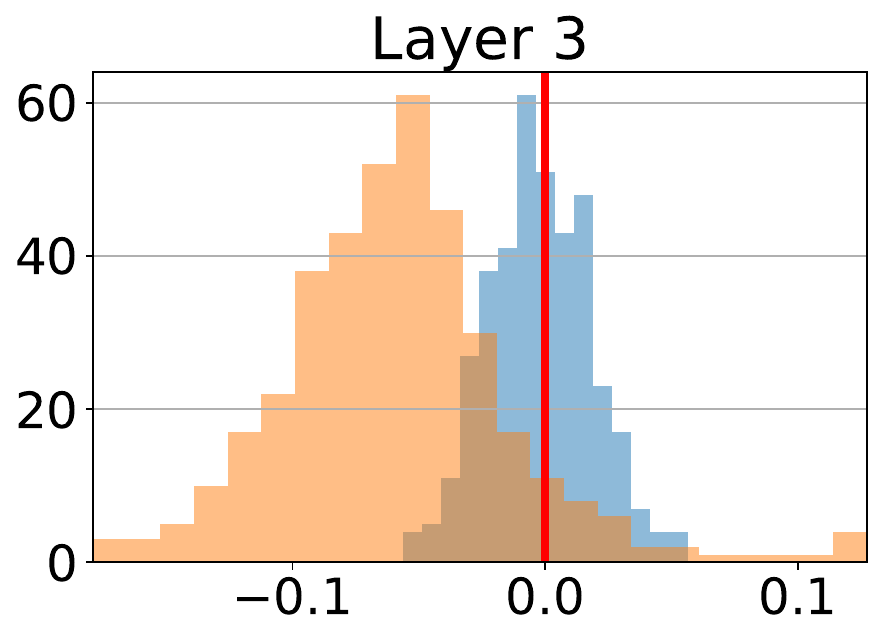}} ~
    \subfloat{\includegraphics[width=0.18\textwidth]{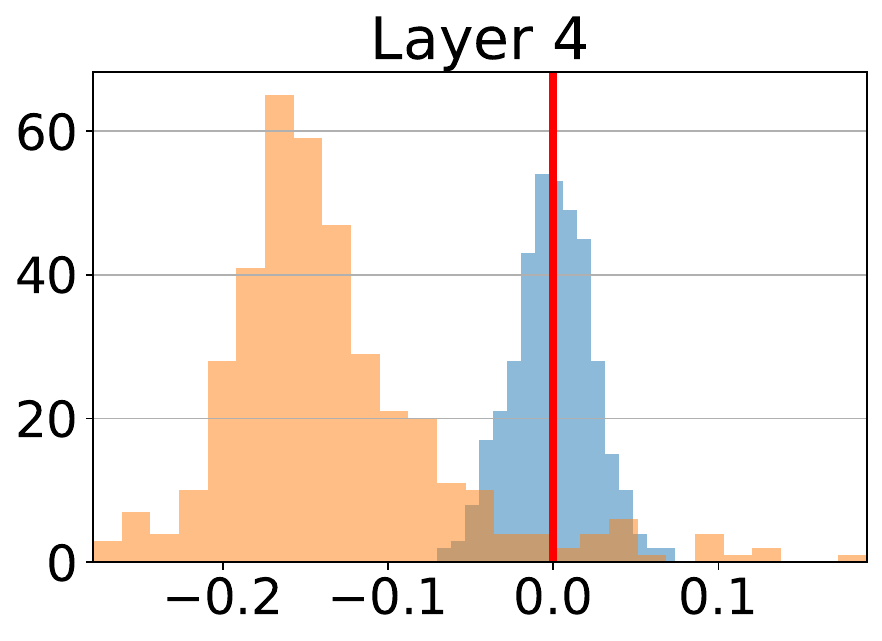}} ~
    \subfloat{\includegraphics[width=0.18\textwidth]{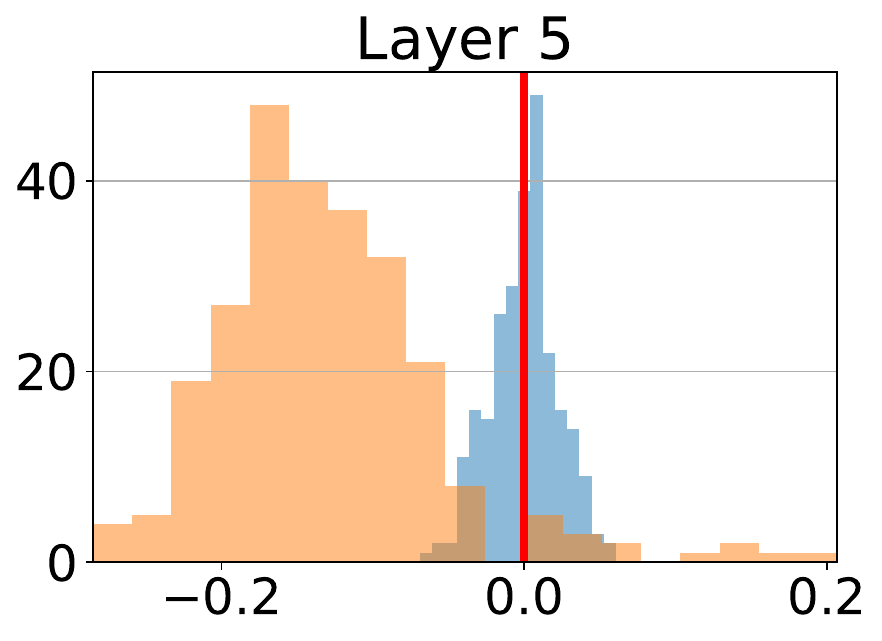}} \\
    
    \caption{Distribution of our projection statistic for a random initialization (blue) and for a \textit{pretrained} \textbf{AlexNet on ImageNet} by the \texttt{MatConvNet} community  (orange).}
    \label{fig:matconvnet_alexnet}
  \end{figure}

\subsection{Bias Correlates with Learning}
\label{appendix:convergence}

\begin{figure}[htbp!]
    \centering
    \includegraphics[width=0.7\textwidth]{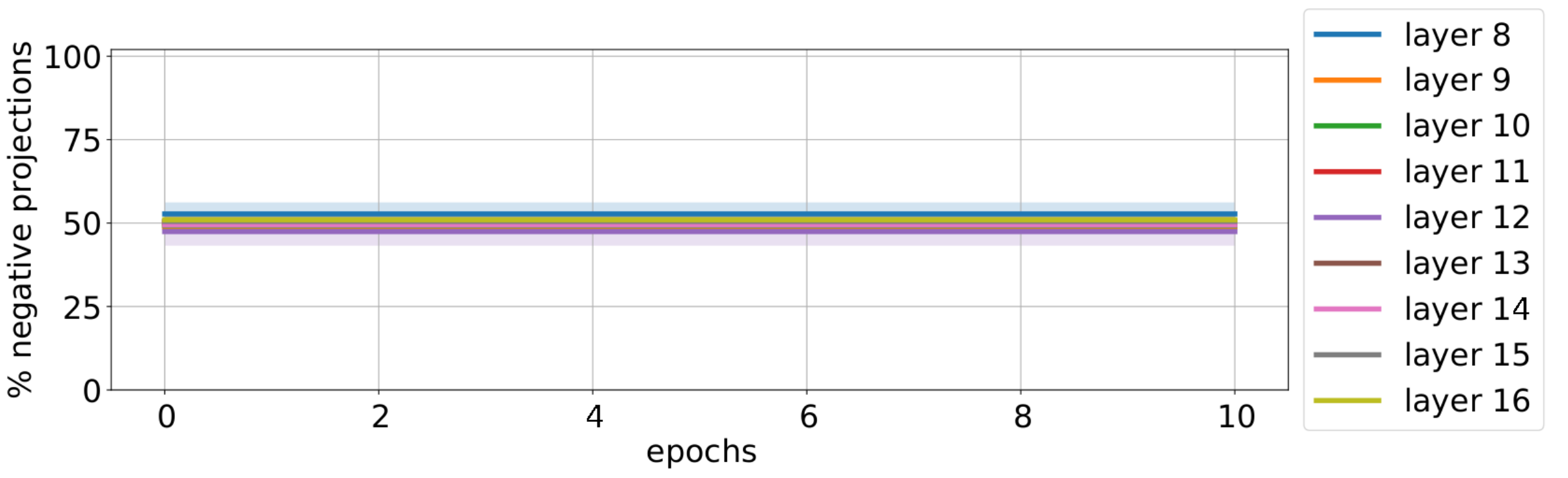}
    \caption{\textbf{Cumulative distribution of negative projection statistics for VGG19 trained on ImageNet} for 10 epochs, with initial weights sampled from $\mathcal{N}(0, 0.01)$. The network fails to learn, with train and validation loss constant throughout training. Correspondingly, the cumulative distribution of our statistic is unchanged by training, indicating that our statistics likely correlate with learning.}
    \label{fig:convergence}
  \end{figure}

Complementing what was described in section~\ref{sec:experiments:learning}, figure~\ref{fig:convergence} shows the development of the cumulative distribution of negative projections for $3$ independent instances of VGG19 that fail to converge on ImageNet. While we observe a change in the deeper layers' weights during training, our statistics remain unbiased, indicating that our measures is likely correlated with \textit{learning}.

\subsection{Biased Layers Are Critical to Performance}
\label{appendix:reinit}

\begin{figure}[ht!]
    \centering
    \subfloat{\includegraphics[width=0.33\textwidth]{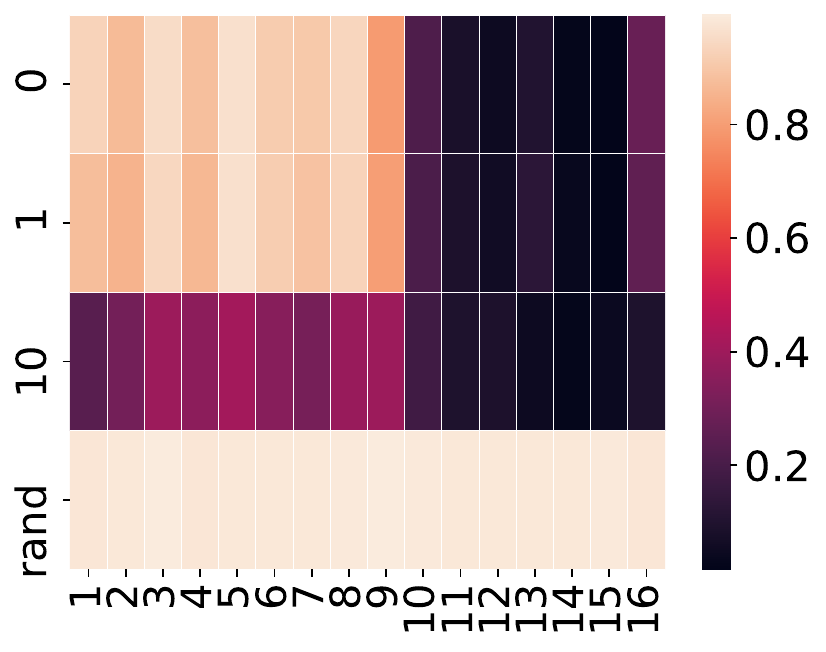}} ~
    \subfloat{\includegraphics[width=0.33\textwidth]{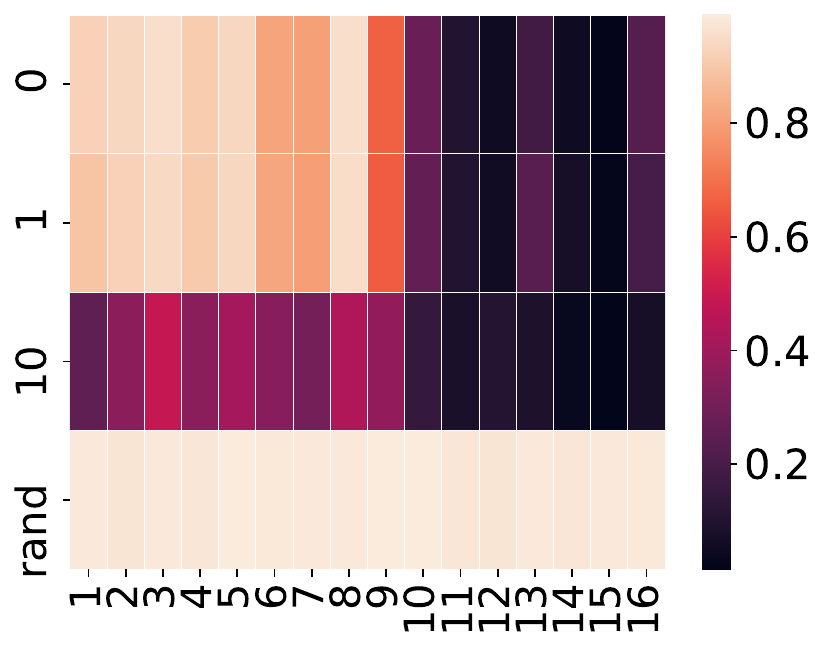}} ~
    \subfloat{\includegraphics[width=0.33\textwidth]{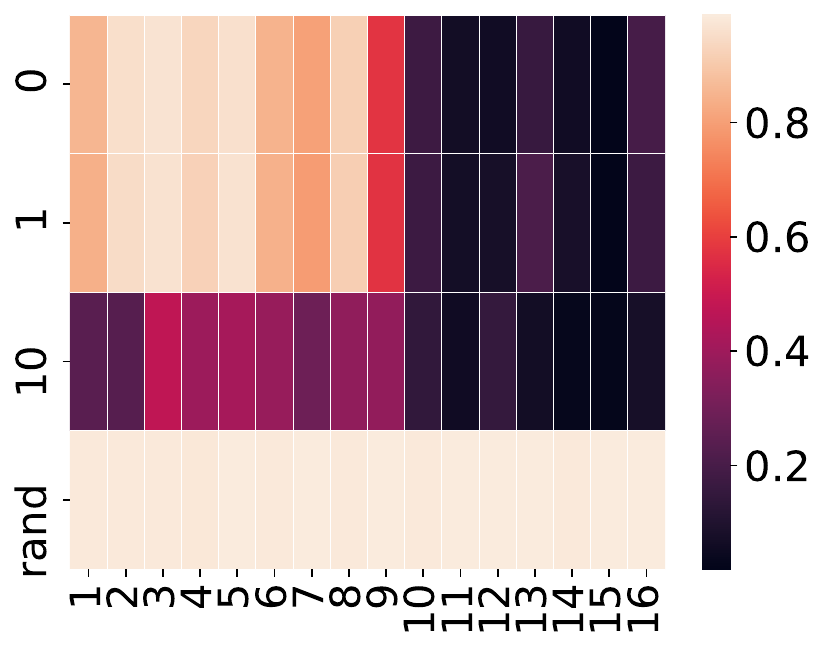}}
    \caption{\textbf{Weight reinitialization experiments for VGG19 trained on CIFAR100.} Each heatmap corresponds to one independent training run.}
    \label{fig:cifar100_vgg19_cont_lr_reinit}
  \end{figure}

\begin{figure}[htbp!]
    \centering
    \includegraphics[width=0.7\textwidth]{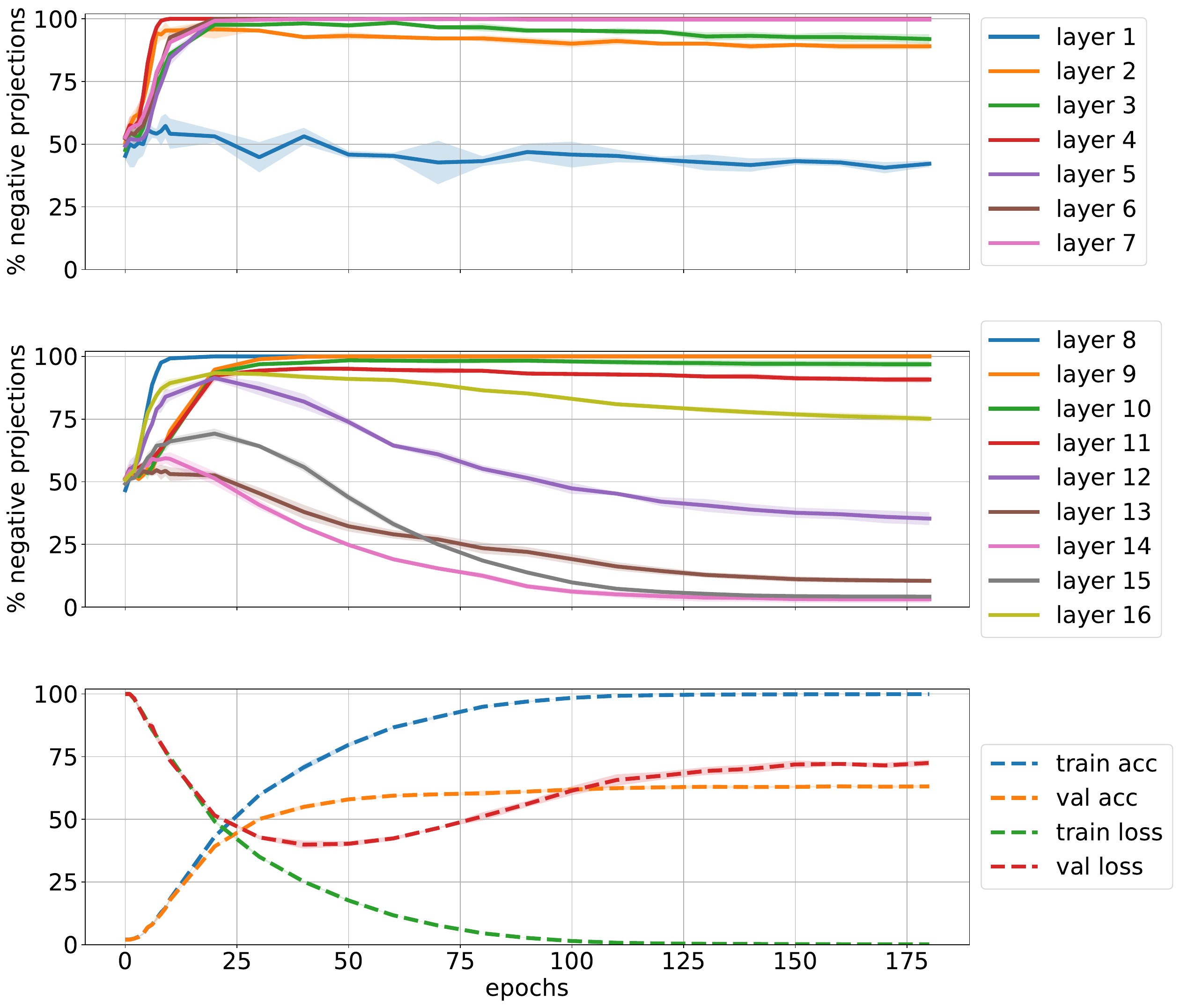}
    \caption{\textbf{Cumulative distribution of negative projection statistics for VGG19 trained on CIFAR100}, with smooth (exponential) learning rate decay.}
    \label{fig:cifar100_vgg19}
  \end{figure}

We extend the main experiments of section~\ref{sec:experiments:critical} by studying critical layers for VGG19 trained with exponential learning rate decay on CIFAR100 (figure~\ref{fig:cifar100_vgg19_cont_lr_reinit}). Also in this case, strongly biased layers are critical to performance. We observed that, compared to VGG19 trained on CIFAR10 (figure~\ref{fig:reinit}), layers $9$ and $10$ are critical to performance and this is reflected by a strong bias in their cumulative distribution of negative projections (figure~\ref{fig:cifar100_vgg19}).

\begin{figure}[h!]
    \centering
    \subfloat{\includegraphics[width=0.33\textwidth]{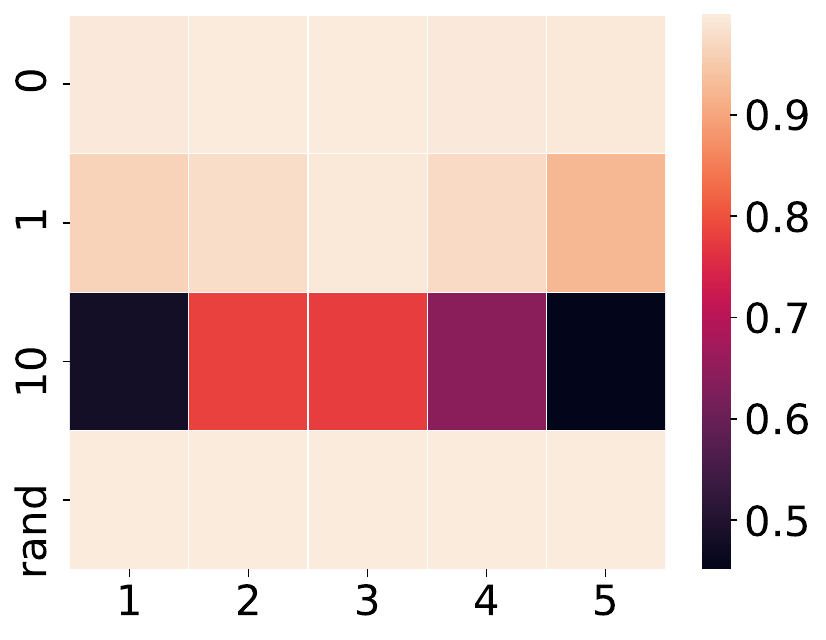}} ~
    \subfloat{\includegraphics[width=0.33\textwidth]{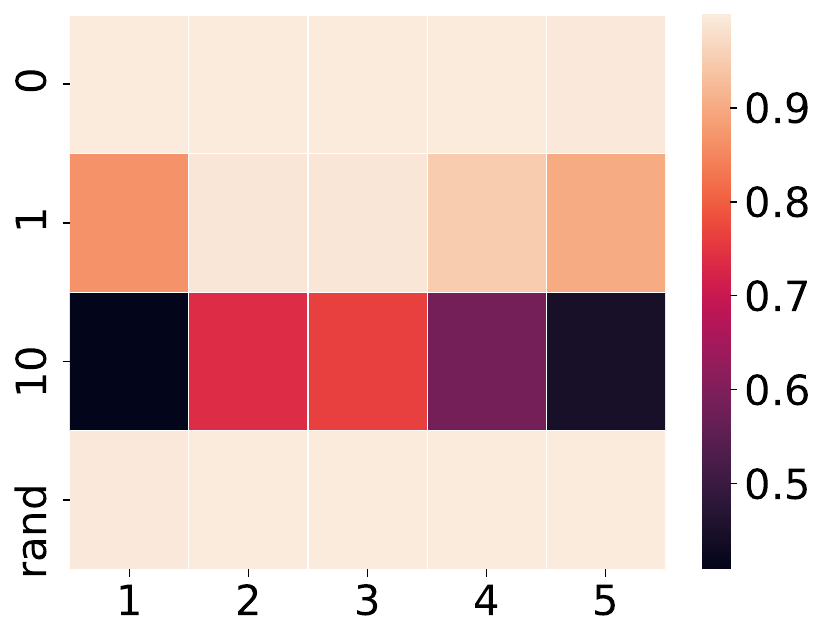}} ~
    \subfloat{\includegraphics[width=0.33\textwidth]{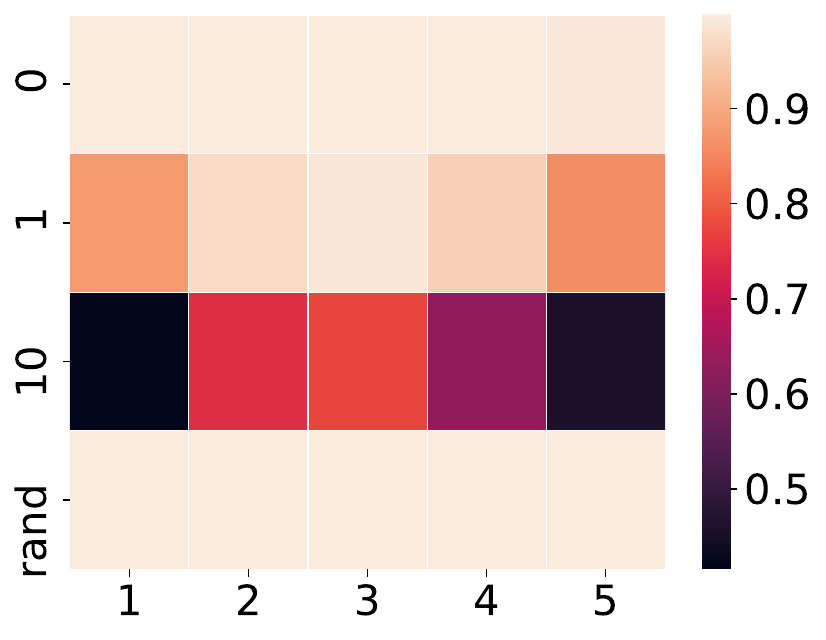}}
    \caption{\textbf{Weight reinitialization experiments for AlexNet trained on ImageNet.} Each heatmap corresponds to one independent training run.}
    \label{fig:imagenet_alexnet_reinit}
  \end{figure}

\begin{figure}[h!]
    \centering
    \subfloat{\includegraphics[width=0.33\textwidth]{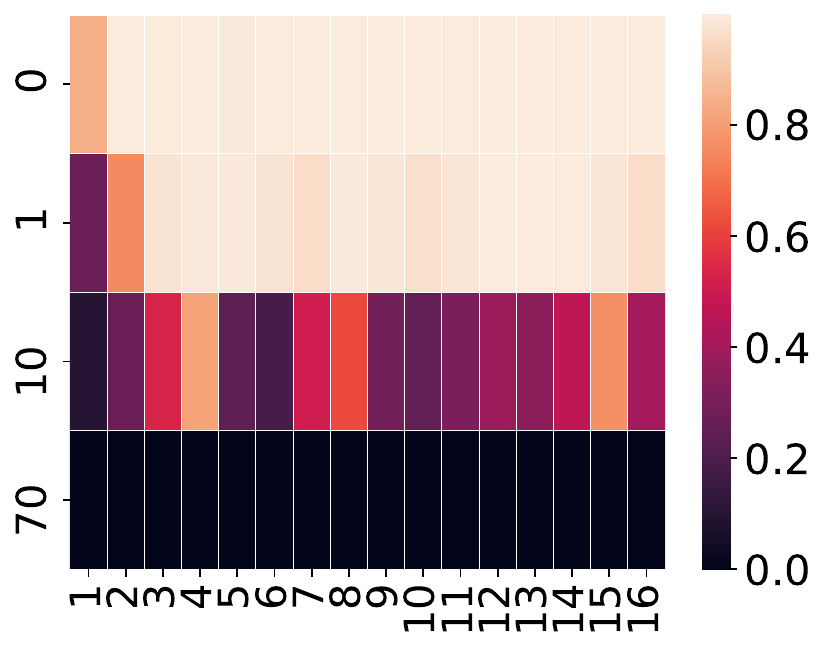}} ~
    \subfloat{\includegraphics[width=0.33\textwidth]{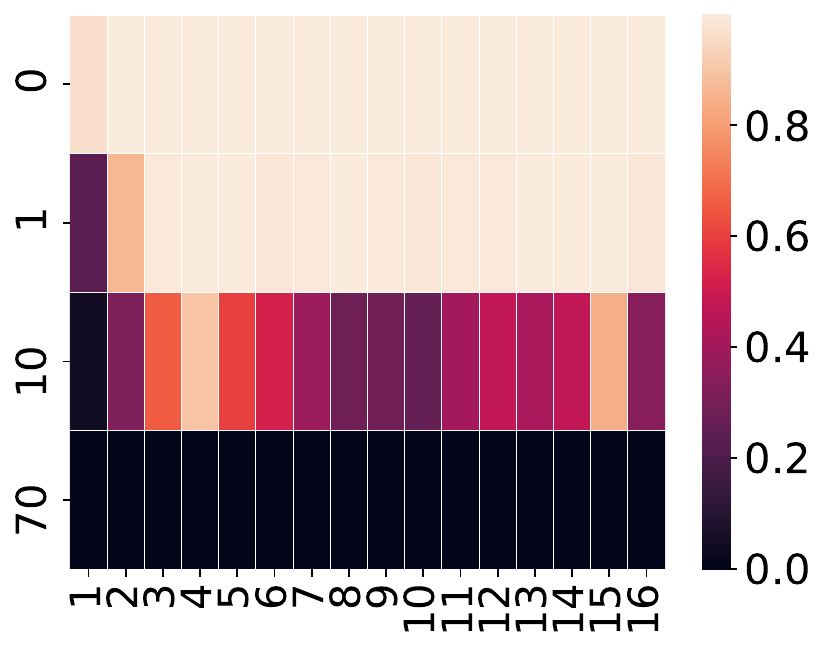}} ~
    \subfloat{\includegraphics[width=0.33\textwidth]{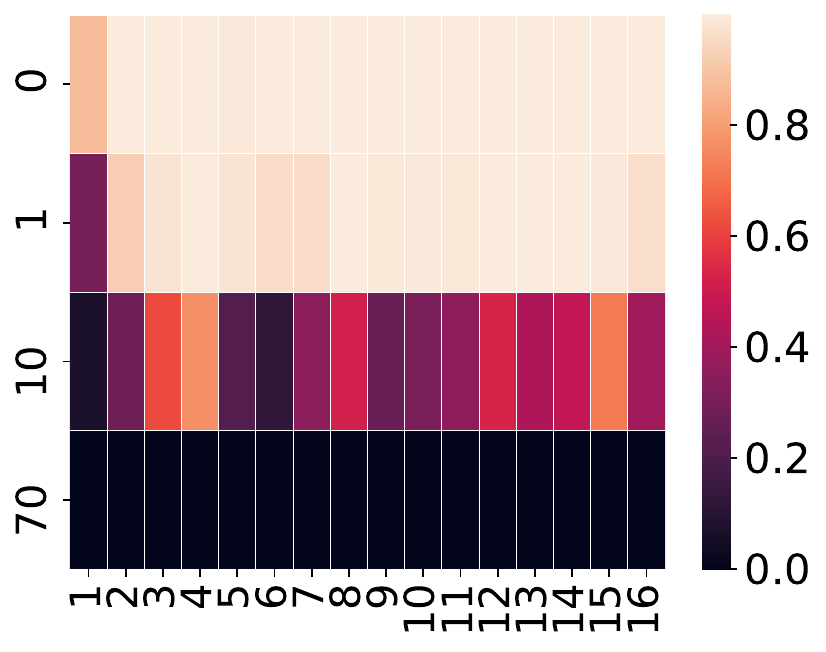}}
    \caption{\textbf{Weight reinitialization experiments for VGG19 trained on ImageNet.} Each heatmap corresponds to one independent training run.}
    \label{fig:imagenet_vgg19_reinit}
  \end{figure}

In contrast, on ImageNet, both VGG19 and AlexNet make use of their full capacity (figures~\ref{fig:alexnet_epochs} and \ref{fig:positive_bias}). This is reflected in the weight reinitialization experiments (figures~\ref{fig:imagenet_alexnet_reinit} and \ref{fig:imagenet_vgg19_reinit} respectively), which show that all layers are critical to performance.

\subsection{Learning Under Noisy Data}
\label{appendix:pixel_shuffle}

\begin{figure}[t!]
    \centering
    \includegraphics[width=0.7\textwidth]{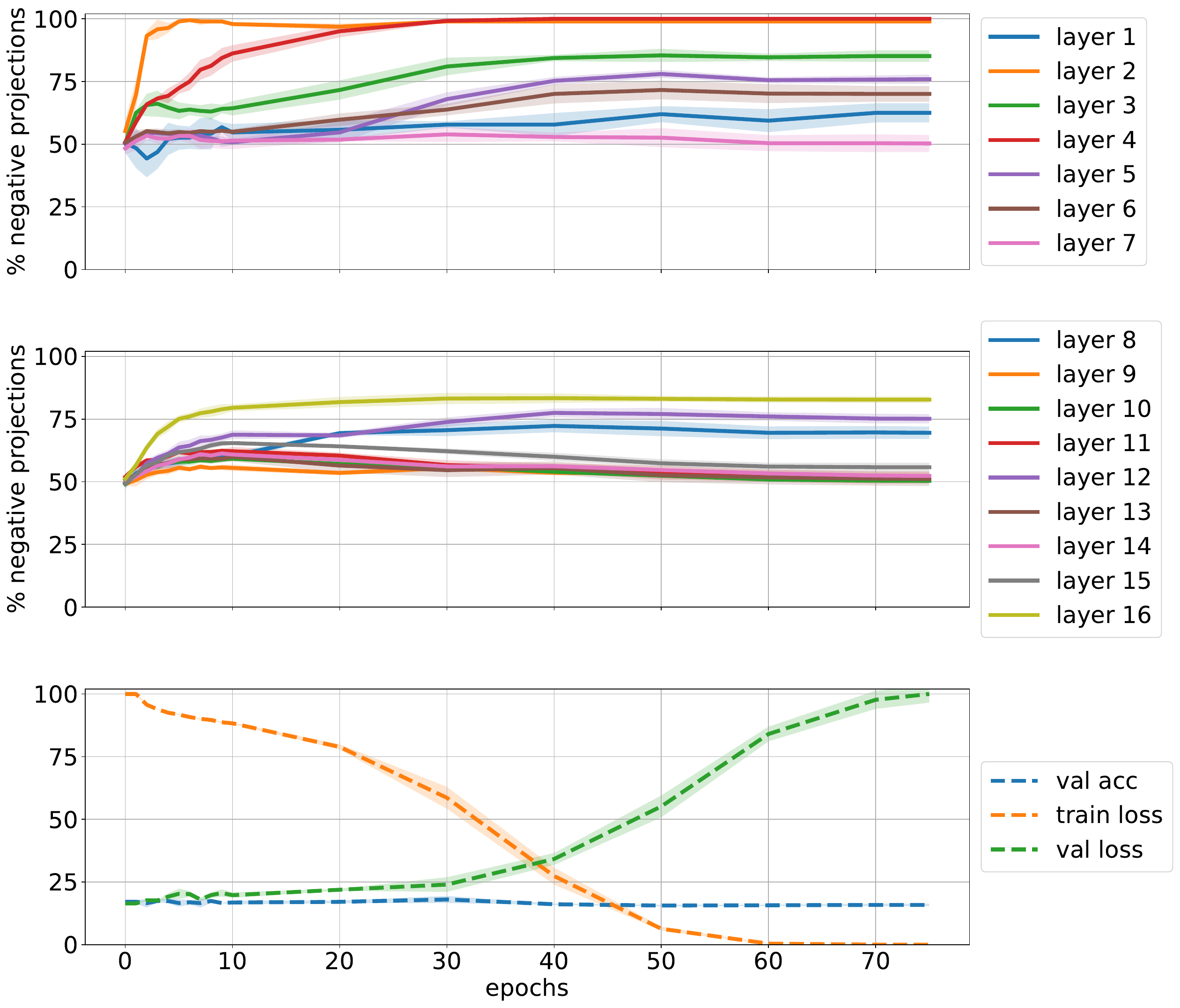}
    \caption{Cumulative distribution of negative projection statistics for \textbf{VGG19 on CIFAR10}. The \textbf{pixel values} of train images have been \textbf{randomly shuffled.}}
    \label{fig:pixel_shuffle}
  \end{figure}

In addition to learning under noisy labels (figure~\ref{fig:noise}), in which image samples are preserved while labels are perturbed, we experiment on training samples with correct labels, but noisy images, in which pixels of all training images are randomly shuffled. This transformation preserves channel-wise statistics, but spatial information is lost. In figure~\ref{fig:pixel_shuffle}, we observe how the statistical bias is reduced, compared to the case of clean data, when the spatial information encoded in each image is disrupted. 

\subsection{Low-level Features}
\label{appendix:first_layer}

  \begin{figure}[b!]
    \centering
    \includegraphics[width=0.55\textwidth]{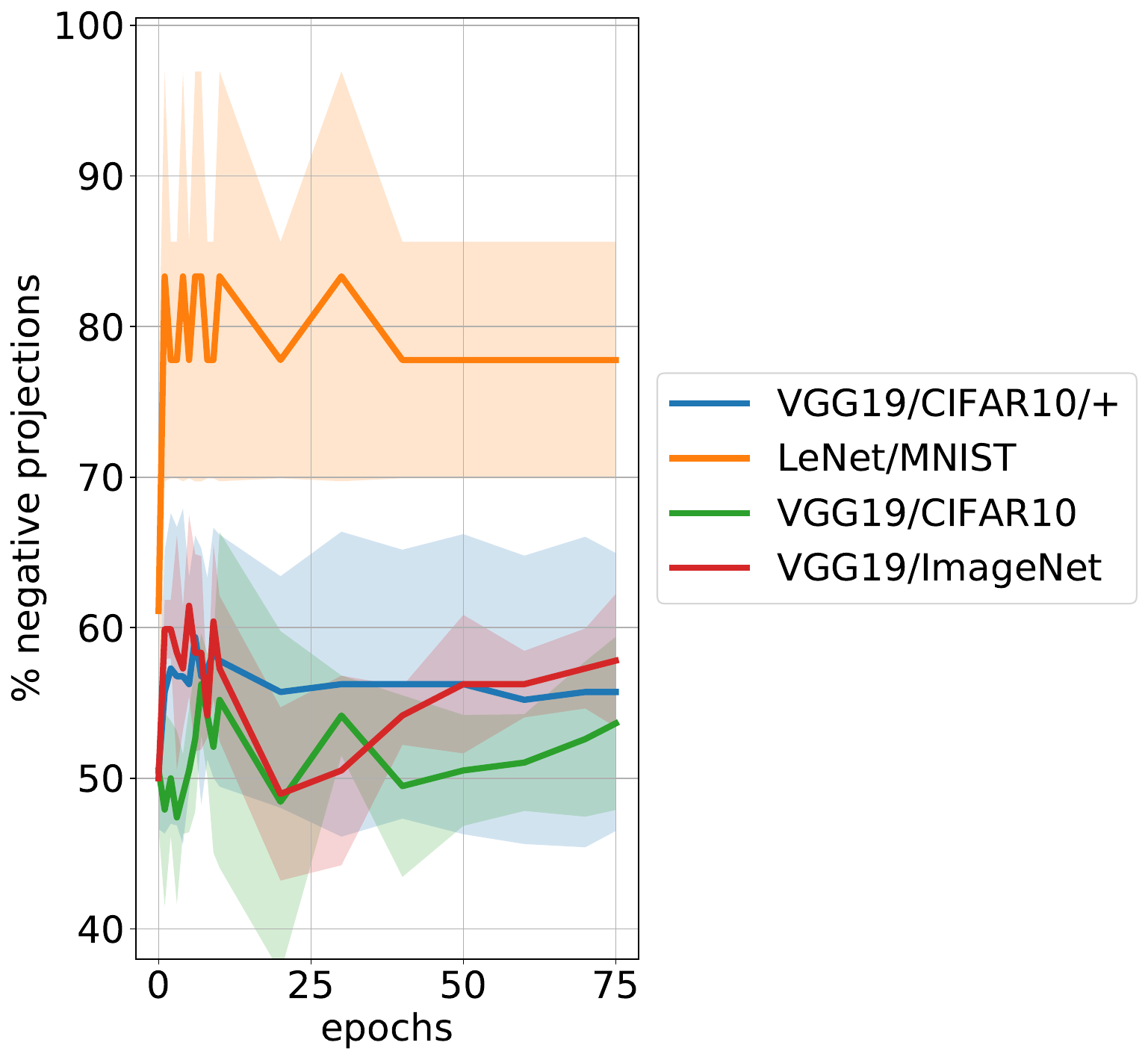}
    \caption{Cumulative distribution of negative projection statistics for the \textbf{first layer of VGG19 trained on CIFAR10, Imagenet, and CIFAR10/+} (pixel values in ${[}0,1{]}$), and \textbf{LeNet on MNIST}. Pixel values of all datasets besides CIFAR10/+ are represented in ${[}-1, 1{]}$. All networks are trained with exponential learning rate decay.}
    \label{fig:first_layer}
  \end{figure}

We take a closer look at the cumulative distribution of projection statistics for the first layer of VGG19 and a variant of LeNet with $6$ convolutional layers, trained on MNIST (figure~\ref{fig:first_layer}). We train VGG19 on RGB datasets (ImageNet and CIFAR10), and represent pixel values in ${[}-1,1{]}$. Additionally, we repeat the experiment for VGG19 trained on CIFAR10 with pixel values in ${[}0,1{]}$, denoted CIFAR10/+. We observe that for all RGB datasets, the distribution of the first layer remains largely unbiased throughout epochs. Furthermore, consistently with our hypothesis (section~\ref{sec:experiments:learning}), the first layer of LeNet on MNIST shows considerable bias towards negative projections. In fact, MNIST digits, as opposed to CIFAR10 and ImageNet images, present only one kind of edge, and thus the first layer might not need to learn the inverse of a edge-detector filter.

\subsection{Training Setup Affects Bias}
\label{appendix:schedules}

\begin{figure}[b!]
    \centering
    \subfloat{\includegraphics[width=0.55\textwidth]{./img/cifar10_vgg19_results.pdf}} ~
    \subfloat{\includegraphics[width=0.45\textwidth]{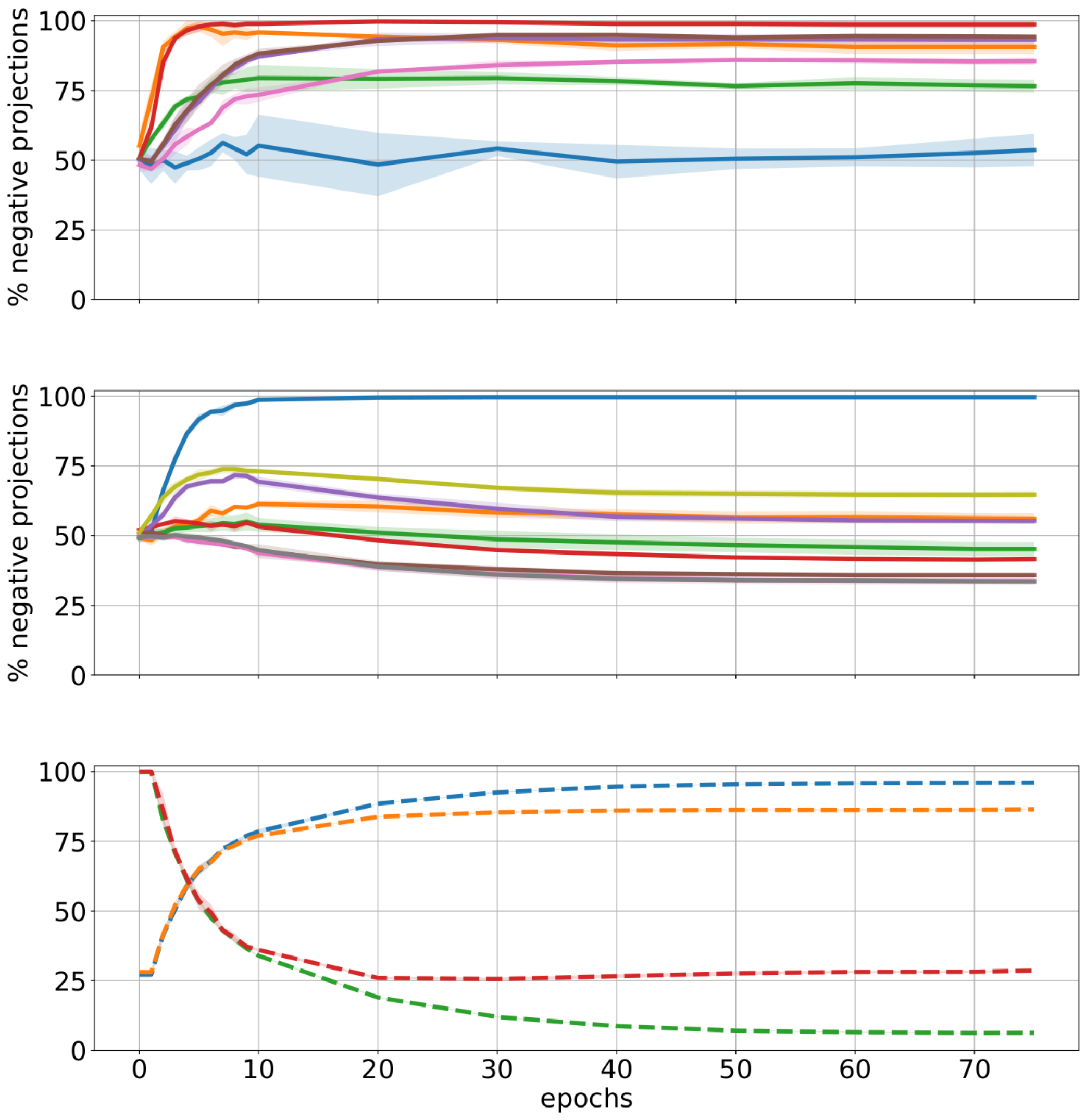}} \\
    \subfloat{\includegraphics[width=0.55\textwidth]{./img/imagenet_vgg19_results.pdf}} ~
    \subfloat{\includegraphics[width=0.45\textwidth]{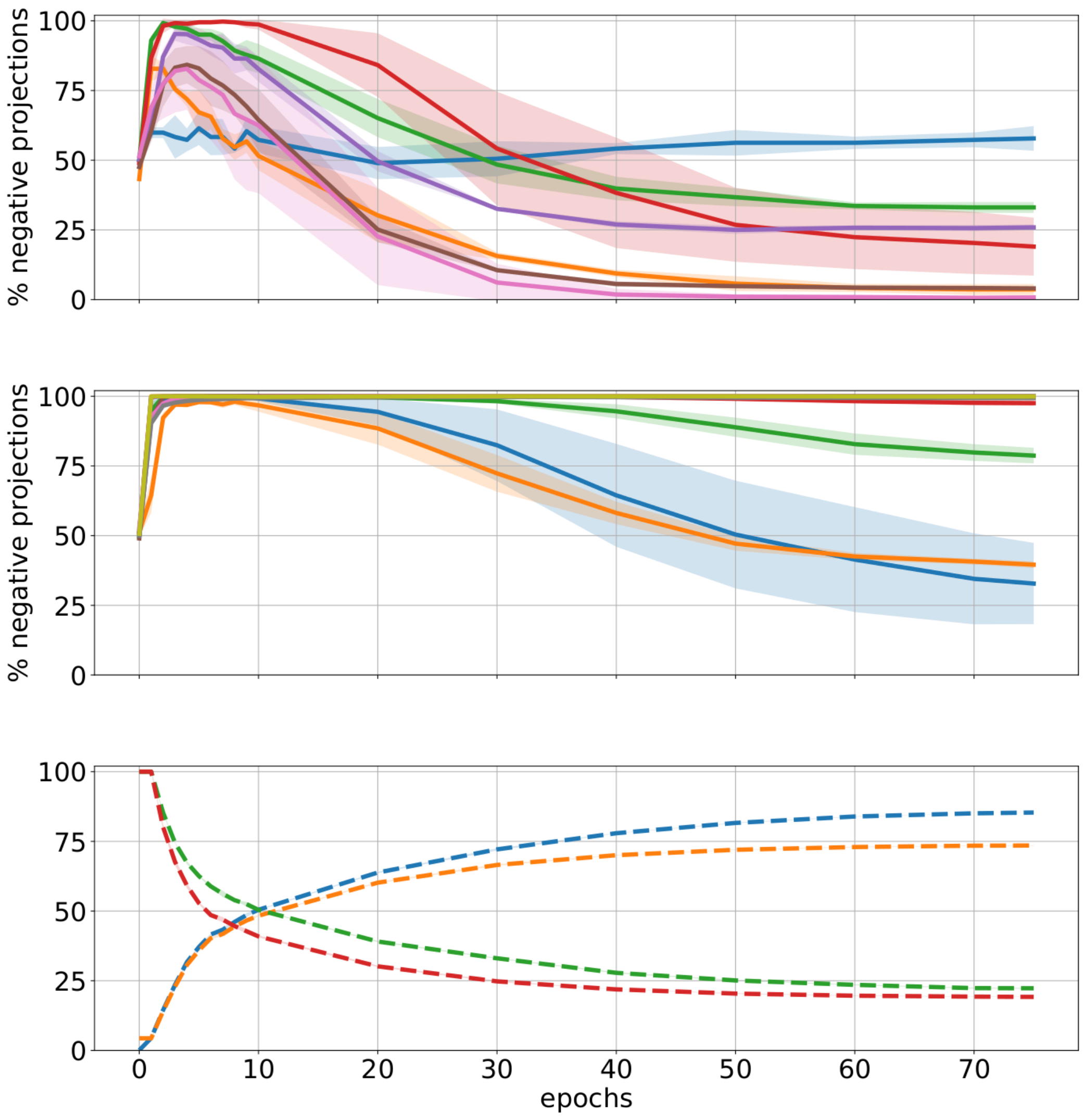}} \\
    \caption{Cumulative density of negative projection statistics for VGG19 with \textbf{step learning rate} (left) \textbf{vs exponential decay schedule} (right). Networks trained on CIFAR10 (top row) and ImageNet (bottom row)}
    \label{fig:schedule_comparison_vgg19}
  \end{figure}
\begin{figure}[b!]
    \centering
    \subfloat{\includegraphics[width=0.33\textwidth]{./img/heatmaps/cifar10_vgg19_reinit_vgg19_74_42.pdf}} ~
    \subfloat{\includegraphics[width=0.33\textwidth]{./img/heatmaps/cifar10_vgg19_reinit_vgg19_74_43.pdf}} ~
    \subfloat{\includegraphics[width=0.33\textwidth]{./img/heatmaps/cifar10_vgg19_reinit_vgg19_74_44.pdf}} \\
    \subfloat{\includegraphics[width=0.33\textwidth]{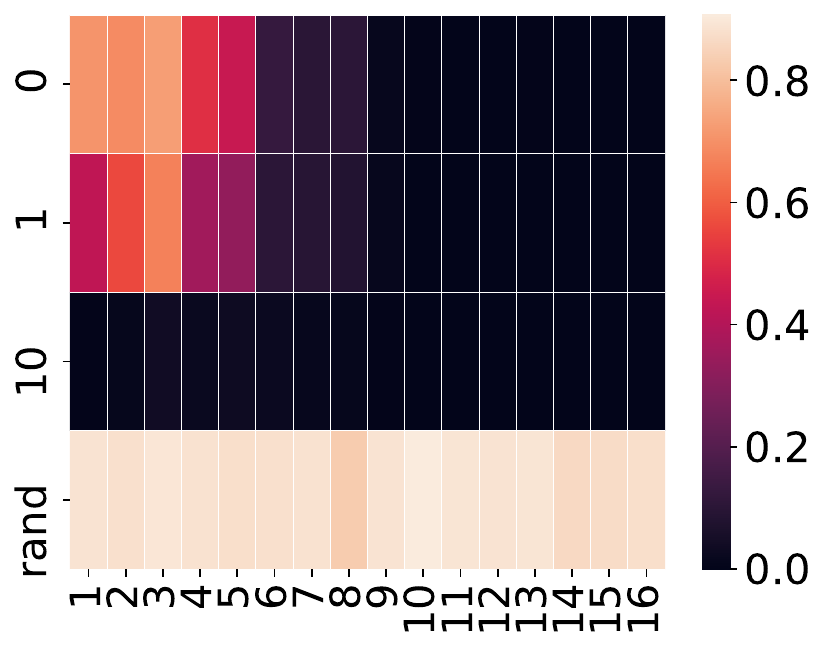}} ~
    \subfloat{\includegraphics[width=0.33\textwidth]{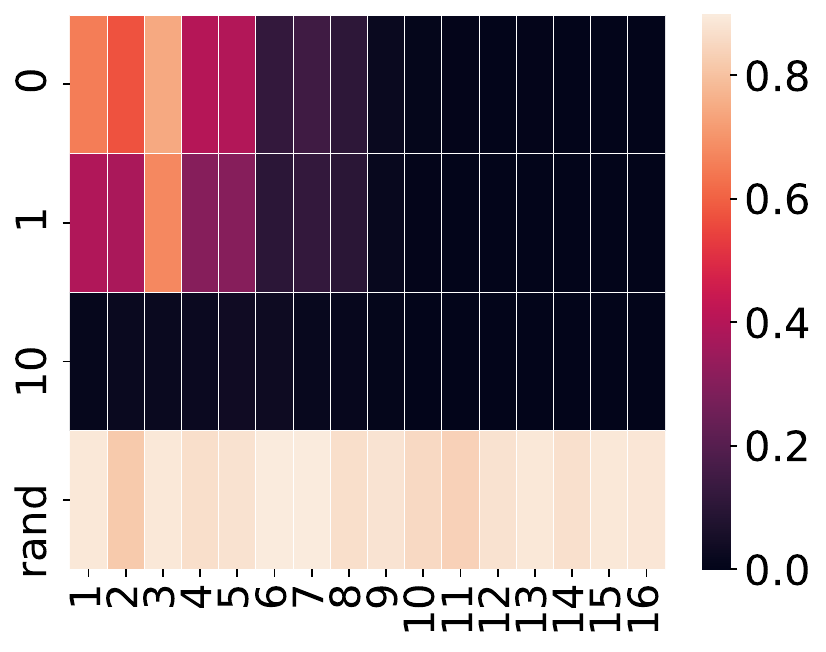}} ~
    \subfloat{\includegraphics[width=0.33\textwidth]{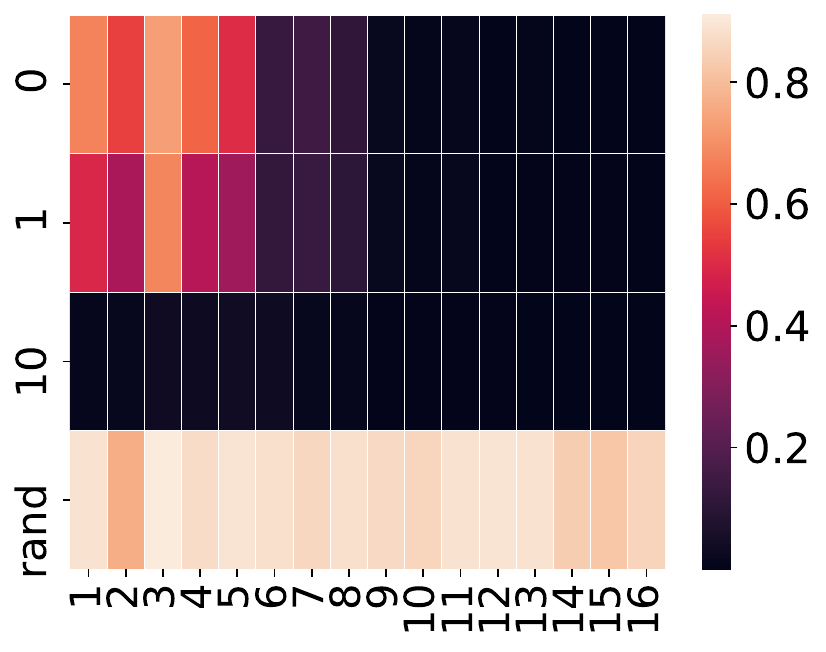}} \\
    \subfloat{\includegraphics[width=0.33\textwidth]{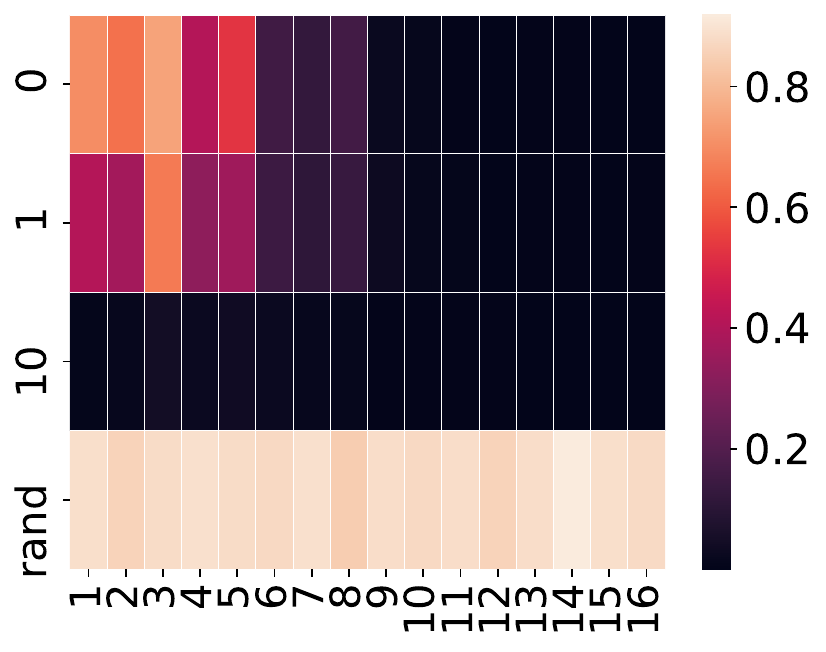}} ~
    \subfloat{\includegraphics[width=0.33\textwidth]{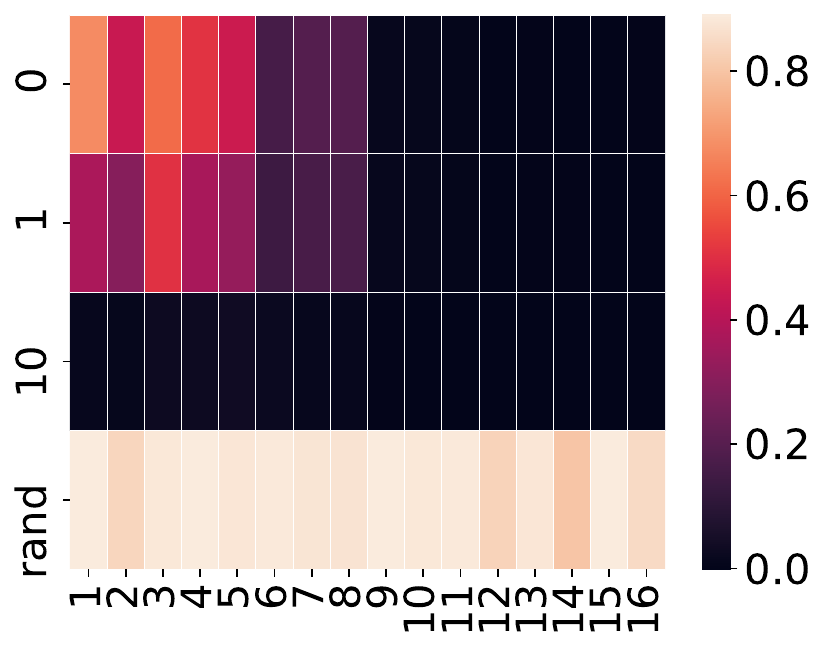}} ~
    \subfloat{\includegraphics[width=0.33\textwidth]{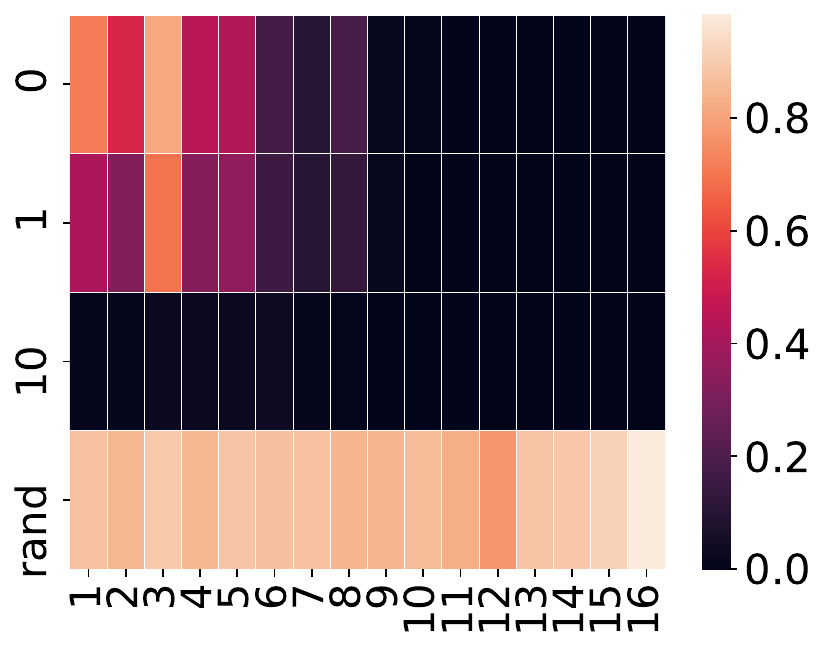}} \\
    \caption{\textbf{Weight reinitialization experiments for VGG19 trained on CIFAR10 with step learning rate schedule.} Each heatmap corresponds to one independent training run. The first three plots from the top-left are also included in the main paper (figure~\ref{fig:reinit}).}
    \label{fig:cifar10_reinit}
  \end{figure}
\begin{figure}[b!]
    \centering
    \subfloat{\includegraphics[width=0.33\textwidth]{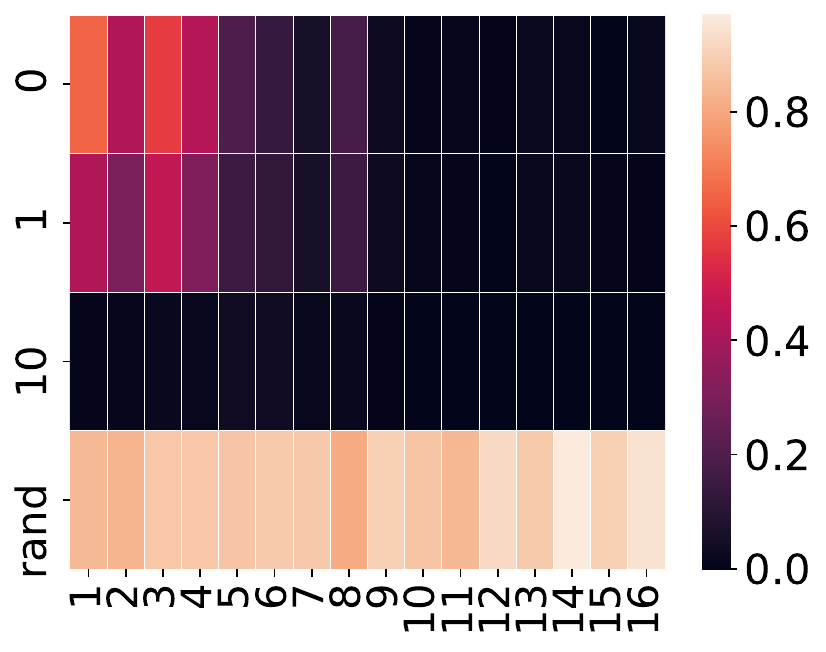}} ~
    \subfloat{\includegraphics[width=0.33\textwidth]{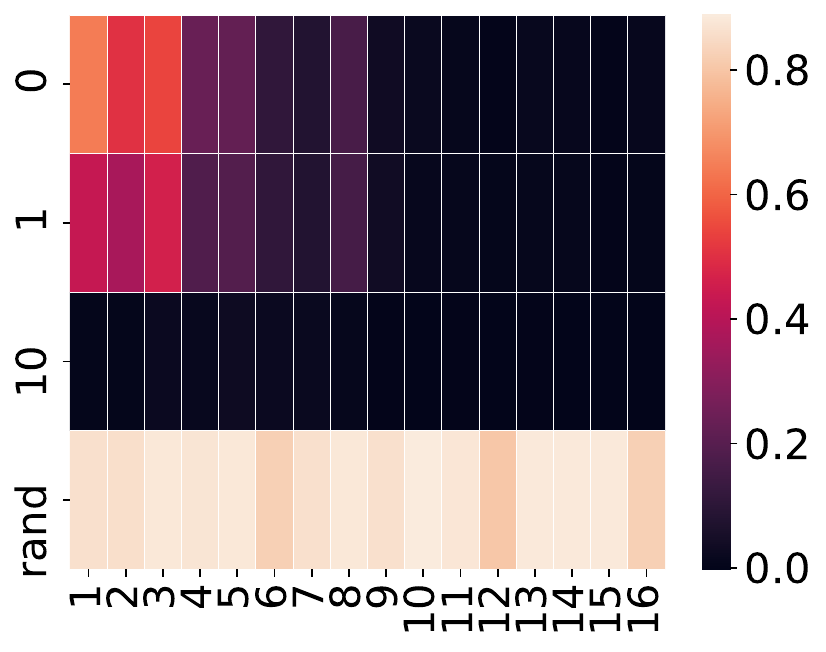}} ~
    \subfloat{\includegraphics[width=0.33\textwidth]{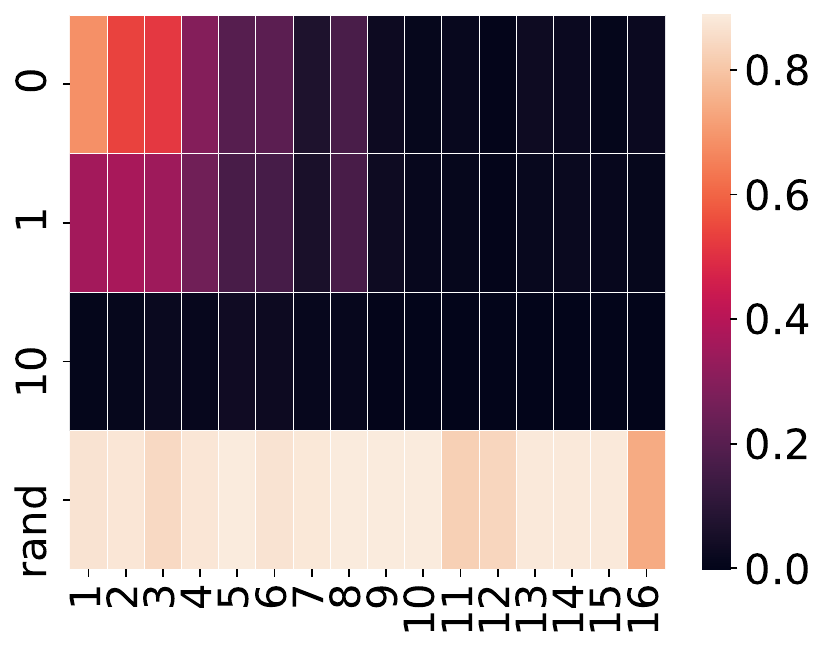}}
    \caption{\textbf{Weight reinitialization experiments for VGG19 trained on CIFAR10 with smooth learning rate decay.} Each heatmap corresponds to one independent training run.}
    \label{fig:cifar10_cont_lr_reinit}
  \end{figure}

We conclude by exploring how learning rate schedules affect our statistics and critical layers. Figure~\ref{fig:schedule_comparison_vgg19}, shows our measure over epochs for step and exponential learning rate schedules for VGG19 trained on CIFAR10 and ImageNet. The step schedule results in reduced variance of our statistic across independent runs and, for CIFAR10, it better separates strongly biased layers ($1$ to $8$). In turn, while the number of critical layers on CIFAR10 is on average the same, the smooth learning rate schedule results in layers with smaller drop in performance under weight reinitialization, corresponding to less statistical bias with respect to the cumulative density of negative projections (figure \ref{fig:cifar10_reinit} and \ref{fig:cifar10_cont_lr_reinit}).

